%% file: main.tex
\documentclass[10pt,twocolumn,letterpaper]{article}

\usepackage{3dv}
\makeatletter
\@namedef{ver@everyshi.sty}{}
\usepackage{times}
\usepackage{epsfig}
\usepackage{graphicx}
\usepackage{amsmath}
\usepackage{amssymb}
\usepackage{float}
\usepackage{tikz}
\usetikzlibrary{spy}

\input{sections/custom_packages}
\usepackage{soul}
\usepackage{rotating}
\usepackage[T1]{fontenc}

\usepackage[pagebackref=true,breaklinks=true,colorlinks,bookmarks=false]{hyperref}
\usepackage{cleveref}

\threedvfinalcopy % *** Uncomment this line for the final submission

\def\threedvPaperID{85} % *** Enter the 3DV Paper ID here

\ifthreedvfinal\fi

\begin{document}

%%%%%%%%% TITLE
\title{Go with the Flows: Mixtures of Normalizing Flows for Point Cloud Generation and Reconstruction}

\newcommand{\printfnsymbol}[1]{%
  \textsuperscript{\@fnsymbol{#1}}%
}

\author{Janis Postels\thanks{equal contribution}\\
ETH Zurich\\
{\tt\small jpostels@vision.ee.ethz.ch}

\and
Mengya Liu\printfnsymbol{1}\\
ETH Zurich\\
{\tt\small mengya.liu@vision.ee.ethz.ch}

\and
Riccardo Spezialetti\\
University of Bologna\\
{\tt\small riccardo.spezialetti@unibo.it}

\and
Luc Van Gool\\
ETH Zurich\\
{\tt\small vangool@vision.ee.ethz.ch}

\and
Federico Tombari\\
Google, TU Munich\\
%Technical University Munich\\
{\tt\small tombari@google.com}
}

\makeatletter
\g@addto@macro\@maketitle{
	\input{figures/teaser/teaser.tex}
}
\makeatother
\maketitle
\thispagestyle{empty}
%%%%%%%%% ABSTRACT
\begin{abstract}
  \input{sections/00_abstract}
\end{abstract}

%%%%%%%%% BODY TEXT
\section{Introduction}

\input{sections/01_introduction}

\section{Related Work}

\input{sections/02_related_work}

\section{Mixtures of Normalizing Flows for Point Clouds}\label{section:theory}

\input{sections/03_method}

\section{Experiments}\label{section:experiments}

\input{sections/04_experiments}

\section{Conclusion}

\input{sections/05_discussion}

{\small
\bibliographystyle{ieee_fullname}
\bibliography{egbib}
}

\clearpage
\twocolumn[
\centering
\section*{Supplementary Material of "Go with the Flows: Mixtures of Normalizing Flows for Point Cloud Generation and Reconstruction"}
]
\vspace{10mm}

\input{sections_supplment/supplement}

% \vfill

%\clearpage
% \clearpage

% {\small
% \bibliographystyle{ieee_fullname}
% \bibliography{egbib}
% }

\end{document}

%% file: sections/custom_packages.tex
\usepackage{array,multirow,graphicx}
\usepackage{float}
\usepackage{subfigure}
\usepackage{acro}

\usepackage[utf8]{inputenc}
\usepackage{pgfplots}
\DeclareUnicodeCharacter{2212}{−}
\usepgfplotslibrary{groupplots,dateplot}
\usetikzlibrary{patterns,shapes.arrows}
\pgfplotsset{compat=newest}

\DeclareAcronym{nf}{
  short = NF,
  long = Normalizing Flow
}

\DeclareAcronym{vae}{
  short = VAE,
  long = Variational Autoencoder
}

\DeclareAcronym{mlp}{
  short = MLP,
  long = multilayer perceptron
}

\DeclareAcronym{elbo}{
  short = ELBO,
  long = Evidence Lower BOund
}

\DeclareAcronym{dpf}{
  short = DPF,
  long = Discrete Point Flow
}

\DeclareAcronym{gan}{
  short = GAN,
  long = Generative Adversarial Network
}

\DeclareAcronym{svr}{
  short = SVR,
  long = single-view reconstruction
}

%% file: figures/teaser/teaser.tex
	\begin{figure}[H]
	\setlength{\linewidth}{\textwidth}
	\setlength{\hsize}{\textwidth}
	\vspace{-9mm}
	\centering
	\renewcommand{\tabcolsep}{0.05pt}
	\renewcommand{\arraystretch}{0.1}
	\begin{tabular}{cccccc}			
		\begin{tikzpicture}[spy using outlines={circle,magenta,magnification=1.5,size=1.2cm, connect spies}]
			\node {\includegraphics[clip,trim=9cm 3cm 9cm 7cm, width=0.15\linewidth]{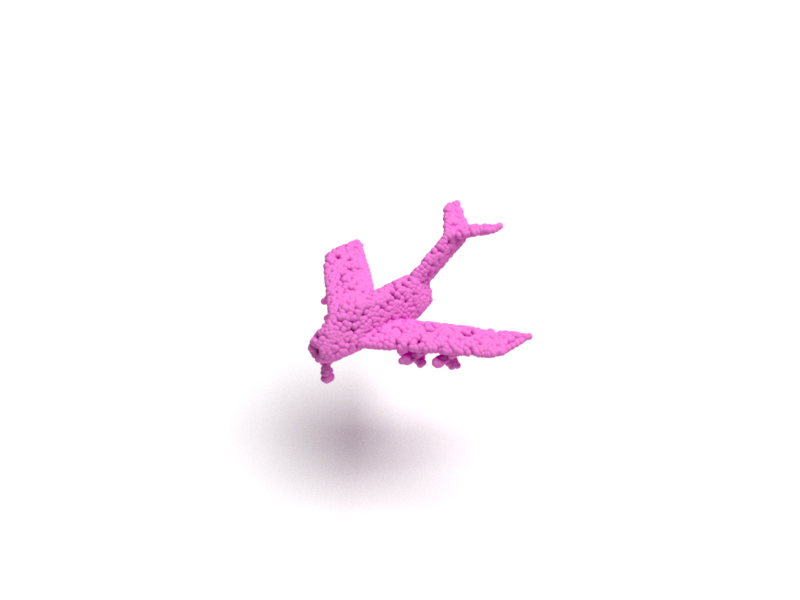}};
			\spy on (0.65, 1.2) in node [right] at (0.0, -0.8);%(1.6, 0.7);
		\end{tikzpicture} &			
		\begin{tikzpicture}[spy using outlines={circle,magenta,magnification=1.5,size=1.2cm, connect spies}]
			\node {\includegraphics[clip,trim=9cm 3cm 9cm 7cm,	width=0.15\linewidth]{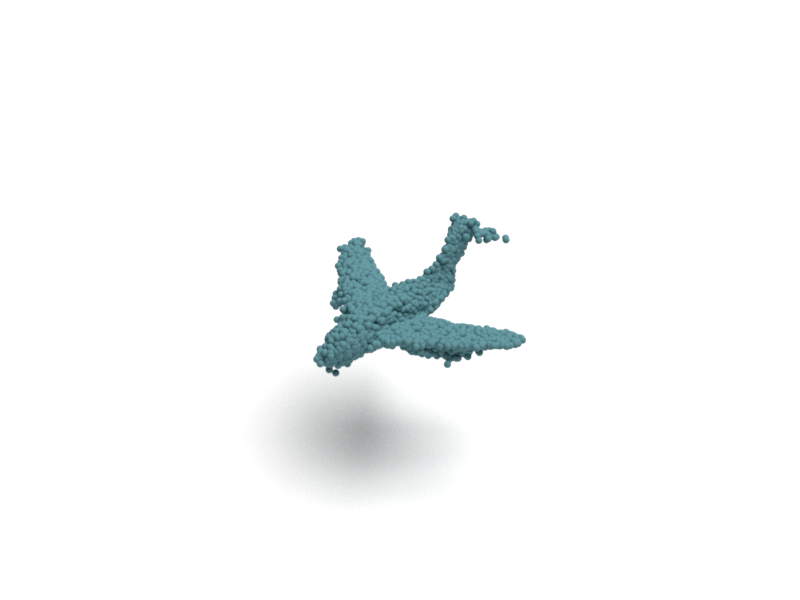}};
			\spy on (0.65, 1.2) in node [right] at (0.0, -0.8);
		\end{tikzpicture} &			
		\begin{tikzpicture}[spy using outlines={circle,magenta,magnification=1.5,size=1.2cm, connect spies}]
			\node {\includegraphics[clip,trim=9cm 3cm 9cm 7cm, width=0.15\linewidth]{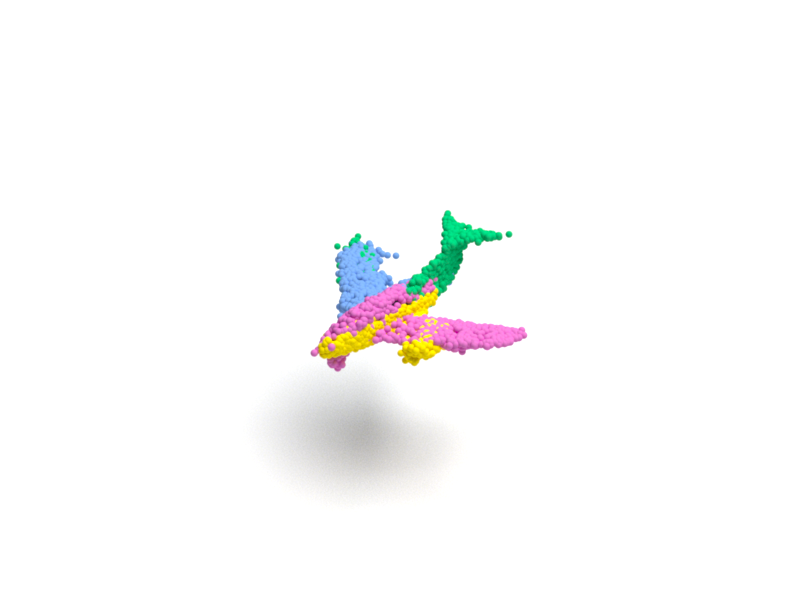}};
			\spy on (0.65, 1.2) in node [right] at (0.0, -0.8);
		\end{tikzpicture} &
		% chair
		\begin{tikzpicture}[spy using outlines={circle,magenta,magnification=1.5,size=1.2cm, connect spies}]
			\node {\includegraphics[clip,trim=10cm 4cm 9cm 6.8cm, width=0.15\linewidth]{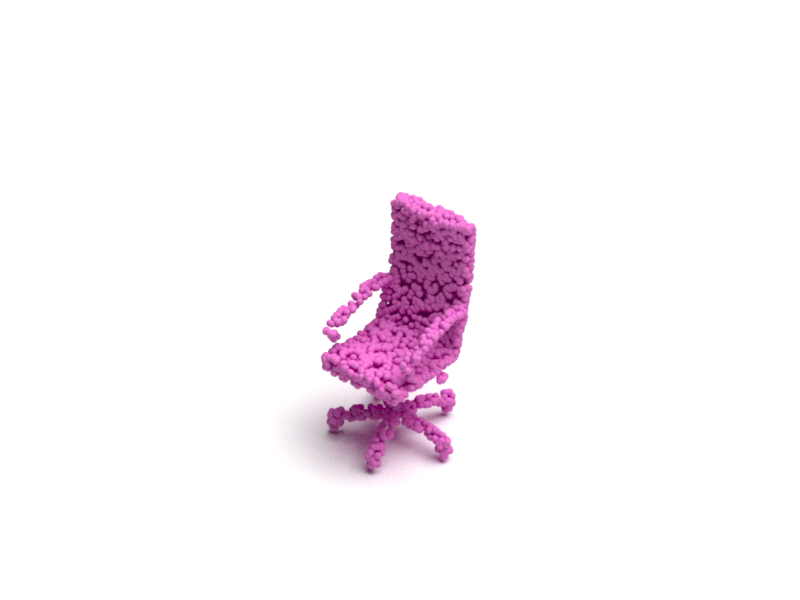}};
			\spy on (-0.6, -1) in node [right] at (0.65, 0.8);
		\end{tikzpicture} &			
		\begin{tikzpicture}[spy using outlines={circle,magenta,magnification=1.5,size=1.2cm, connect spies}]
			\node {\includegraphics[clip,trim=10cm 4cm 9cm 6.8cm, width=0.15\linewidth]{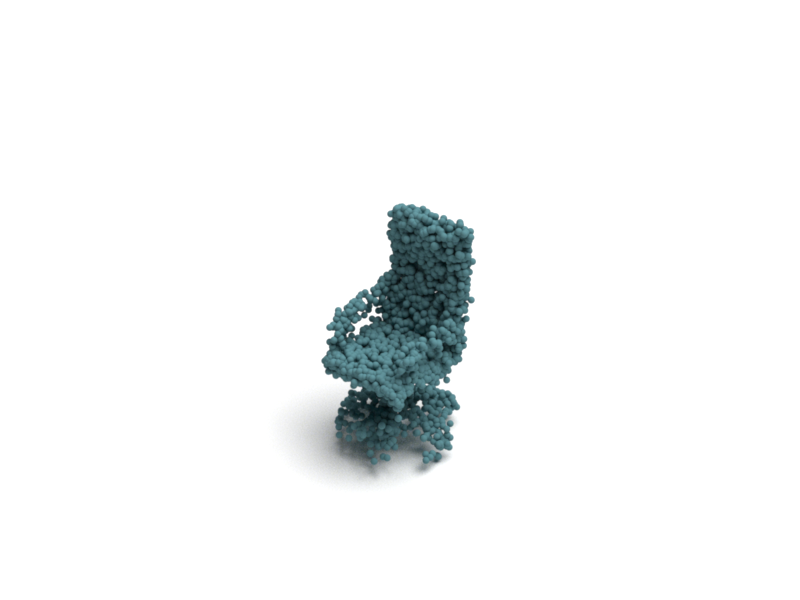}};
			\spy on (-0.5, -1) in node [right] at (0.65, 0.8);
		\end{tikzpicture} &
		\begin{tikzpicture}[spy using outlines={circle,magenta,magnification=1.5,size=1.2cm, connect spies}]
			\node {\includegraphics[clip,trim=10cm 4cm 9cm 6.8cm, width=0.15\linewidth]{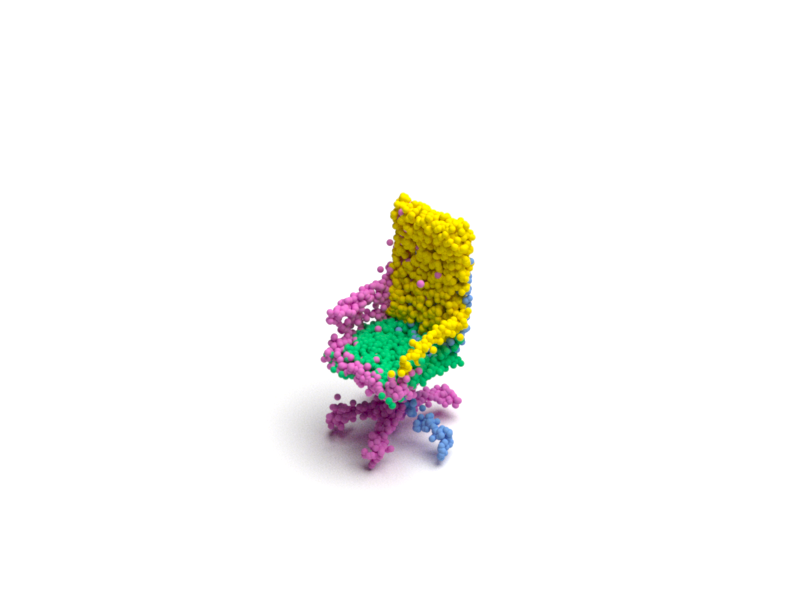}};
			\spy on (-0.5, -1) in node [right] at (0.65, 0.8);
			%clip,trim=3cm 4cm 0cm 5cm,
		\end{tikzpicture} \\			
		\footnotesize Ground truth & 
		\footnotesize Single NF \cite{klokov2020discrete}& 
		\footnotesize NF Mixture (ours) &
		\footnotesize Ground truth & 
		\footnotesize Single NF \cite{klokov2020discrete}& 
		\footnotesize NF Mixture (ours) \\						
	\end{tabular}
	\vspace{1mm}
	\caption{Our method (Ours) uses mixtures of Normalizing Flows (NFs) to generate point clouds overcoming limitations of single normalizing flows \cite{yang2019pointflow, klokov2020discrete}. Each mixture component (indicated by coloring) specializes in a distinct subregion in an unsupervised fashion.%In the right, each color maps to the region generated by a different flow. %Some shape details are magnified inside the circles.
	}
	\label{fig:teaser}
\end{figure}

%% file: sections/00_abstract.tex
Recently \acp{nf} have demonstrated state-of-the-art performance on modeling 3D point clouds while allowing sampling with arbitrary resolution at inference time. However, these flow-based models still have fundamental limitations on complicated geometries. This work generalizes prior work by introducing additional discrete latent variable, \ie mixture model. This circumvents limitations of prior approaches, leads to more parameter efficient models and reduces the inference runtime. Moreover, in this more general framework each component learns to specialize in a particular subregion of an object in a completely unsupervised fashion yielding promising clustering properties. We further demonstrate that by adding data augmentation, individual mixture components can learn to specialize in a semantically meaningful manner. We evaluate mixtures of \acp{nf} on generation, autoencoding and single-view reconstruction based on the ShapeNet dataset.

%% file: sections/01_introduction.tex
Nowadays point clouds, as the output of many modern 3D scanning devices, \eg LiDARs and RGB-D cameras, denote an increasingly popular data format for 3D shapes. Thus, a generative model that can sample shapes represented as point clouds is valuable for several 3D computer vision down-stream tasks such as shape completion, synthesis and up-sampling. 
Although \textit{\ac{vae}} \cite{kingma2013auto}, \textit{\ac{gan}} \cite{goodfellow2014generative} and \textit{Normalizing Flows} \cite{dinh2016density} (NFs) have shown impressive results on various applications \cite{isola2017image, yang2019pointflow, arroyo2021variational}, point clouds remain challenging for these methods due to their lack of a regular underlying grid structure compared to images.
% Old version: Techniques such as \textit{\ac{vae}}, \textit{\ac{gan}} and \textit{Normalizing Flows} (NFs) , have shown impressive results on image generation tasks \cite{kingma2013auto, goodfellow2014generative, dinh2016density}, however, point clouds due to its speciality that the lack of a regular underlying grid structure like image has, is still challenging for these methods to model the complex distribution of 3D shapes.
% Point clouds denote an increasingly popular data format for 3D shapes. While they are a natural intermediate data format between sensor data and more elaborate representations (\eg meshes), as an intrinsic shape representation they also scale much more gracefully than extrinsic representations (\eg voxels) with increasing resolution. 
%are particularly promising.
% generative models of 3D point clouds are gaining attention \cite{achlioptas2018learning, yang2019pointflow} since they constitute a natural representation when considering point clouds as 3D distributions and can generate shapes with arbitrary number of points. 
% Recently, among generative models on point clouds, approaches

Prior work handled the irregular structure of point clouds by generating shapes with a fixed number of points  using GANs or auto-regressive models \cite{achlioptas2018learning,gadelha2018multiresolution,zamorski2020adversarial,shu20193d,fan2017point}. Recently, another family of generative models, \acp{nf}, has gained attention due to its appealing properties \cite{yang2019pointflow, pumarola2020c, klokov2020discrete}. While they naturally allow trading off runtime with resolution by adapting the number of generated points at inference time, their invertibility also allows training them by directly minimizing the negative log-likelihood of the data leading to improved training stability over GANs \cite{achlioptas2018learning}. 
%Typically, the point cloud distribution is modeled as a distribution of distributions \cite{yang2019pointflow, klokov2020discrete} in the framework of VAEs, where the decoder parameterizes an individual point cloud, \ie distribution of points, conditioned on a sample from the (latent) distribution of shapes.
% At inference time they naturally allow trading off runtime with resolution by adapting the number of generated points. Furthermore, their invertibility allows training them by directly minimizing the negative log-likelihood of the data. This leads to improved training stability over \acp{gan}. Typically, the distribution of point clouds is modeled as a distribution of distributions \cite{yang2019pointflow, klokov2020discrete} in the framework of \acp{vae}. Here, the decoder parameterizes an individual point cloud, \ie distribution of points, conditioned on a sample from the (latent) distribution of shapes.

%While the application of continuous \cite{yang2019pointflow} and discrete \cite{klokov2020discrete} \acp{nf} yielded state-of-the-art performance on shape generation and reconstruction benchmarks, %performance regarding generation and reconstruction metrics, they still come with the caveat of high computational costs, \eg training \cite{klokov2020discrete} on ShapeNet \cite{chang2015shapenet} takes longer than a week. 
While the application of continuous \cite{yang2019pointflow} and discrete \cite{klokov2020discrete} \acp{nf} yielded state-of-the-art performance on shape generation and reconstruction benchmarks, it also has innate limitations.
%One root cause for this is related to the strength of \acp{nf}, \ie their invertibility. 
Transforming a standard Gaussian using an invertible map into a complex geometry accurately (\eg with holes or multiple modes) requires squeezing/expanding space infinitely strong \cite{dinh2019rad, cornish2020relaxing}. Therefore, one must resort to very deep \acp{nf} \cite{cornish2020relaxing}. Our work generalizes \acp{nf} by introducing additional discrete latent variables. Thus, shapes are composed as a product of experts using multiple \acp{nf}.
% Similarly to how our visual system decomposes shapes into parts %To bypass this barrier for the applicability of \acp{nf} to practical problems, this work introduces mixtures of \acp{nf} for modeling point clouds. 
%In these mixtures multiple \acp{nf} learn to compose a final shape as the product experts, where each \ac{nf} learns to specialize in a subregion of the shape in an unsupervised fashion similar to AtlasNet \cite{groueix2018papier} - but with the ability of generating an arbitrary number of points. This alleviates the problem of mismatching abstract geometric properties of the source distribution and the target distribution by learning to sew together the final object using several invertible maps. Moreover, the resulting generative model exhibits interesting clustering properties that potentially enable broad application.
%Therefore, this work introduces additional discrete latent variables, effectively yielding a mixture of \acp{nf}. Similar to how our visual system decomposes shapes into parts \cite{hebb2005organization,hoffman1984parts}, 
% Similarly to how our visual system decomposes shapes into parts \cite{hebb2005organization,hoffman1984parts}, to bypass this barrier for the applicability of \acp{nf} to practical problems, this work introduces mixtures of \acp{nf} for modeling point clouds. 
%the proposed framework learns to compose a shape as the product experts using multiple \acp{nf}. 
Each \ac{nf} specializes in a subregion in an unsupervised fashion - see \cref{fig:teaser} where the point color indicates the \ac{nf}. Importantly, this alleviates the problem of mismatching abstract geometric properties of the source distribution and the target distribution by learning to sew together the final object using several invertible maps. Besides performance gains, this further yields interesting clustering that potentially enables broad applications such as unsupervised part segmentation, semantic correspondence, etc.\ %Note, our approach naturally inherits the ability of generating an arbitrary number of points, which clearly differentiates it from \cite{groueix2018papier}.

We demonstrate that mixtures of \acp{nf} introduced in \cref{section:theory} generalize and exceed single-flow-based models at similar size, while reducing the inference runtime. This increased representational strength manifests itself in superior generation and reconstruction and in improved details on the generated/reconstructed point clouds resulting from individual \acp{nf} specializing in subregions of the 3D shapes.

%% file: sections/02_related_work.tex
\subsection{Mixtures of Normalizing Flows}
\acp{nf} \cite{rezende2015variational, dinh2014nice, dinh2016density} are a class of generative models that allow efficient likelihood evaluation using invertible transformations. Recently, they have fueled a variety of applications \cite{kingma2016improved, chen2016variational, lugmayr2020srflow, wolf2021deflow}. Despite their popularity, there have been surprisingly few works on mixtures of \acp{nf} \cite{dinh2019rad, cornish2020relaxing, giaquinto2020gradient, izmailov2020semi}, all focused on toy data problems. 
\cite{dinh2019rad} separates the space into disjoint subsets using piece-wise linear activation functions to let each flow specialize on one subset. However, the discontinuity arising from partitioning leads to training difficulties \cite{cornish2020relaxing}. In turn, \cite{cornish2020relaxing} relaxes the invertibility constraint of \acp{nf} by introducing additional continuous latent variables yielding improved performance on MNIST and CIFAR10. \cite{giaquinto2020gradient} trains a mixture of \acp{nf} using bosting, where each flow learns the residual likelihood. Our work refrains from applying \cite{cornish2020relaxing} or \cite{giaquinto2020gradient} since the continuous nature of the latent variables \cite{cornish2020relaxing} and the iterative training procedure \cite{giaquinto2020gradient} do not allow obtaining well separated clusters. 
The mixture of \acp{nf} trained in \cite{pires2020variational} is closest to ours. However, they operate on toy data and the latent variables of their \ac{vae} only encompass the mixture weights whereas our \ac{vae}'s continuous latent variables encode 3D shapes on which we condition the mixture of \acp{nf}. Lastly, \cite{izmailov2020semi} uses a Gaussian mixture model as the base distribution for a \ac{nf} and applies this scheme to semi-supervised learning.

\subsection{Generative Models for Point Clouds}
%The point cloud representation is the most widely adopted format to store 3D data for many scene understanding tasks due to their simplicity and direct relation to common data acquisition technique such as LiDARs, depth cameras, laser scanner, etc. 
Due to the unorganized structure of point clouds, pioneering generative models treat point clouds as a set of 3D points organized into a $N \times 3$ matrix, where $N$ is fixed \cite{achlioptas2018learning,gadelha2018multiresolution,zamorski2020adversarial,shu20193d,fan2017point,sun2020pointgrow,valsesia2018learning}. For example, Gadelha \etal{} \cite{gadelha2018multiresolution} combine a multi-resolution encoder-decoder to form a VAE \cite{kingma2013auto} for point cloud generation. Achlioptas 
\etal{} \cite{achlioptas2018learning} explore the use of GANs \cite{goodfellow2014generative,gulrajani2017improved} to generate point clouds. However, generating a point cloud with a fixed number of points limits its flexibility. This issue has been partially mitigated with the introduction of \textit{plane-folding} decoders \cite{groueix2018papier,yang2018foldingnet}, which learn to deform 2D points sampled from a grid into a set of 3D points allowing to generate shapes with an arbitrary number of points. However, the above methods \cite{ben2018multi,valsesia2018learning,sun2020pointgrow} rely on heuristic set distances such as the Chamfer distance (CD) and the Earth Mover’s distance (EMD), which both lead to several drawbacks \cite{yang2019pointflow,cai2020learning}. While the CD favors point clouds that are concentrated in the mode of the marginal point distribution, the EMD is often computed by approximations and thus can lead to biased gradients. 

Alternatively, a point cloud can be viewed as a point/3D distribution. PointGrow \cite{sun2020pointgrow} models this distribution auto-regressively. ShapeGF \cite{cai2020learning} applies an energy-based framework to model a shape by learning the gradient field of its log-density.
%\cite{stypulkowski2019conditional} models point clouds using conditional \acp{nf}.
However, they do not learn the low-dimensional shape embeddings leading to poor performance compared to related work. PointFlow \cite{yang2019pointflow} employs two continuous \acp{nf} \cite{grathwohl2018ffjord} to model both shape distribution and point distributions. \cite{spurek2020hyperflow} generates the weights of the continuous \ac{nf} using a hypernetwork paired with a spherical log-normal base distribution achieving similar results as PointFlow \cite{yang2019pointflow}. But these works are computationally expensive due to differential equations \cite{chen2018neural}. \ac{dpf} \cite{klokov2020discrete} uses \textit{discrete} affine coupling layers %%to avoid the computationally expensive solvers for differential equations,
resulting in a significant speed-up of the method. Moreover, other works develop conditional \acp{nf} to improve the representation performance. Pumarola \etal{} \cite{pumarola2020c} proposes a novel conditioning scheme for \acp{nf} to address 3D reconstruction and rendering from point clouds. SoftFlow \cite{kim2020softflow} conditions a single \ac{nf} on the noise magnitude used during training. More recently, \cite{luo2021diffusion} uses a diffusion probabilistic process to model 3D point clouds.
%Our method generalizes other flow-based methods overcoming limitations when modelling complex geometries by representing a shape as a mixture of \acp{nf}.

Concurrently, \cite{kimura2020chartpointflow} uses multiple continuous \acp{nf} for point clouds inheriting long training times from PointFlow \cite{yang2019pointflow}. Note that they do not train a mixture of \acp{nf}, but rather an unrelaxed version using hard assignments. Finally, \cite{paschalidou2021neural} also proposes multiple invertible maps for point clouds. However, they focus on reconstruction - without learning a generative model - and rely on a set of handcrafted optimization objectives instead of maximizing log-likelihood.

%% file: sections/03_method.tex
This section, initially, revisit VAEs \cite{kingma2013auto}, which are used to approximate the distribution of point clouds, and \acp{nf}. Then, we introduce our main contribution - mixtures of \acp{nf} - for modeling and learning from point clouds.
% \cite{yang2019pointflow, klokov2020discrete

\subsection{Background}\label{section:background}

\noindent{\textbf{Normalizing Flows}} \cite{rezende2015variational, dinh2014nice, dinh2016density} are explicit generative models that transform a simple base distribution $p(y)$ (\eg Gaussian $\mathcal{N}(0, 1)$) into a complex data distribution $p(x)$ (\eg point cloud) using a series of $n$ invertible transformations $f = f_{n-1} \circ  ... \circ f_0$ with $f: y \mapsto x$. Hereby, the $f_i$ are designed to allow efficient evaluation of the log-determinant of their Jacobian. Thus, using a change of variable, we can train \acp{nf} by directly minimizing the negative log-likelihood of the data $-\log(p_f(x))$ under the model as follows
\begin{align} \label{eq:basic_nf}
    \mathcal{L}(\theta) &= \underset{x\sim p(x)}{E}\left[ -\log(p_f(x)) \right] \\
    &= - \underset{x\sim p(x)}{E}\left[ \log(p(y)) - \sum_{i=0}^{n-1} \log\left( \det|J_{f^{-1}_i} (x)| \right) \right]  \nonumber
\end{align}
where $\theta$ denotes the parameters of the invertible maps and $y=f^{-1}(x)$. A common choice for the invertible transformation is the coupling layer \cite{dinh2016density}. Given an input $z \in \mathrm{R}^d$ a coupling layer c splits the dimensions of $y$ into two sets $L \subset \{1, \dots, d\}$ and  $K \subset \{1, \dots, d\}$ with $L \cup K = \emptyset$. It then applies the identity mapping to one set $c(y^L) = y^L$. The other set of features is scaled by $s = s(y^L)$ and translated by $t = t(y^L)$ such that $c(y^K) = s(y^L) \odot y^K + t(y^L)$ where $s(y^L)$ and $t(y^L)$ are typically (non-invertible) \acp{mlp}. The log-determinant of the Jacobian of this transformation is equivalent to summing the scaling factors $\log\left(\det|J_{c} (y)|\right) = \sum_{k\in K} s_k(y^L)$. Further, conditioning a \ac{nf} on a variable $z$ is commonly achieved \cite{ardizzone2019guided} by introducing conditional scaling and translation in each coupling layer - $t(y;z)$ and $s(y;z)$ - and/or parameterizing the mean and the diagonal covariance matrix of the base distribution as functions of $z$, \ie $p(y|z) = \mathcal{N}(y; \mu(z), \Sigma(z))$. Here, we follow \cite{klokov2020discrete} and apply both mechanisms.

\noindent{\textbf{Variational Autoencoders}} \cite{kingma2013auto} are latent variable models that approximate a data distribution $p(x)$ by minimizing the negative \ac{elbo} 
\begin{align} \label{eq:basic_elbo}
    &-ELBO(\theta, \psi, \phi) = \underset{z\sim p(z|x)}{E}\left[ -\log(p_{\theta}(x|z)) \right] \nonumber\\
    &+ D_{KL}\left( p_{\phi}(z|x) || p_{\psi}(z) \right) = \mathcal{L}_D + \mathcal{L}_{Prior}
\end{align}

Here, $D_{KL}$ is the Kullback-Leibler divergence, $p_{\phi}(z|x)$ denotes an encoder parameterizing the approximate posterior distribution, $p_{\psi}(z)$ is a prior distribution of z and $p_{\theta}(x|z)$ is a decoder model parameterizing the distribution of x conditioned on z. While the prior distribution is often fixed (\eg standard Gaussian $\mathcal{N}(0, 1)$), this work chooses the more flexible approach of learning its parameters.

\begin{figure*}[t]
\centering
\includegraphics[width=0.85\linewidth]{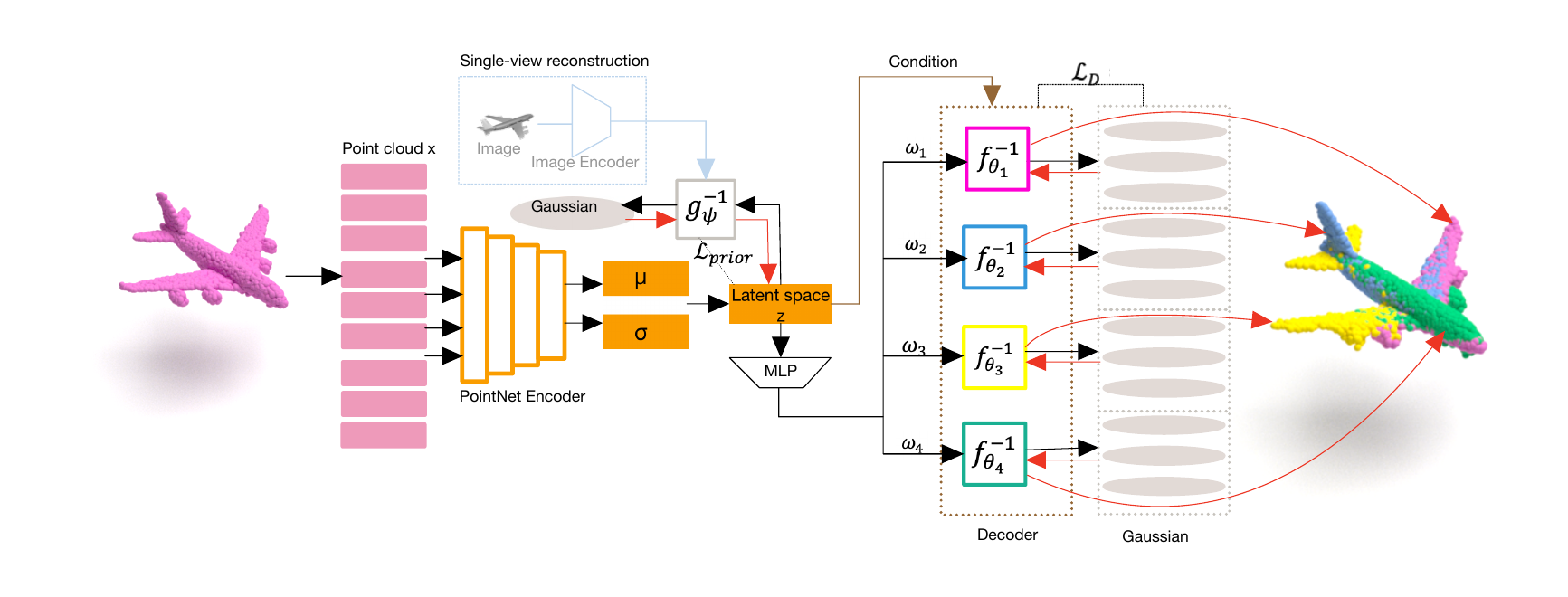}
\caption{Model architecture. Black arrows indicate the training process, while red arrows represent the sampling of a point cloud from our model. During training, PointNet encodes a point cloud $X$ to infer a posterior distribution $p_{\phi}(z|X)$ which can be seen as a distribution over shapes. A prior distribution parameterized by a \ac{nf} $g_{\psi}^{-1}$ is trained by maximizing the log-likelihood of $z$. When training \ac{svr}, we condition $g_{\psi}^{-1}$ on an image encoding. The decoder is parameterized as a mixture of \acp{nf} conditioned on $z$ where each flow $f_{\theta_i}^{-1}$ learns to specialize in a subregion of the shape. Our model is optimized end-to-end by minimizing $\mathcal{L} = \mathcal{L}_{prior} + \mathcal{L}_D$.}
\vspace{-3mm}
\label{fig:architecture}
\end{figure*}

\subsection{Method}\label{seciotn:method}
We model a distribution $p(X)$ of 3D point clouds $X\in \mathrm{R}^{N\times 3}$ where each point cloud itself represents a distribution $p(x)$ over points $x\in \mathrm{R}^3$ in 3D space. We model $p(X)$ using a \ac{vae} wherein each point distribution $p(x)$ is represented by a mixture of \acp{nf}. Subsequently, in this context $z\in\mathrm{R}^d$ refers to a d-dimensional latent representation of an entire point cloud. The overall architecture is depicted in \cref{fig:architecture}. 
%Our main contribution is the decoder model which we implement as a mixture of conditional \acp{nf} that learn to specialize in distinct subregions of the point cloud in an unsupervised fashion. Subsequently, we discuss each component of the \ac{vae} in more details. \\

\input{figures/2d_toy_example/toy_example}

\noindent{\textbf{Mixtures of \acp{nf} for point distributions.}} Prior work estimated point distributions using either a single continuous \cite{yang2018foldingnet} or discrete \cite{klokov2020discrete} \ac{nf} conditioned on the latent shape representation $z$. Despite strong performance, using a single (conditional) \ac{nf} to transform a standard Gaussian $\mathcal{N}(0, 1)$ into complicated geometries has fundamental limitations \cite{dinh2019rad, cornish2020relaxing}, since it requires infinite bi-Lipschitz constants in the limit of arbitrary precision \cite{cornish2020relaxing}. Further explanations and a toy example are in the supplement. To achieve such bi-Lipschitz constants, it is necessary to use large/deep \acp{nf} which limit their practical relevance. Fig. \ref{fig:toy_example} provides further intuition regarding the advantages of mixtures of \acp{nf} in form of a 2D toy example. To bypass this shortcoming, one could either use a prior distribution with a similar geometry as the target distribution or a mixture of \acp{nf}. The former requires dynamically adapting the prior distribution to the geometry of the target distribution, leading to a chicken-and-egg problem. Therefore, we choose to model point clouds as a mixture of \acp{nf}, \ie using several independent invertible maps. Thus, complicated geometries are composed of separate simpler geometries. Formally, we model the conditional point distribution $p_{\theta}(x|z)$ in \cref{eq:basic_elbo} using a mixture of $m$ conditional \acp{nf}
\begin{equation} \label{eq:mixture_nf}
    p(x|z) = \sum_{i=0}^{m}w_i(z) \mathcal{N}(f^{-1}_{\theta_i}(x); \mu(z), \Sigma(z)) \det|J_{f^{-1}_{\theta_i}}(x;z)|
\end{equation}
where $f_{\theta_i}$ denotes the i-th \ac{nf} with its parameters $\theta_i$, $\mathcal{N}(f^{-1}_{\theta_i}(x); \mu(z), \Sigma(z))$ is the likelihood under the shared prior distribution and $w_i(z) \geq 0$ are the mixture weights denoting the probability that a point of a point cloud is generated by i-th \ac{nf}. They are a function of the latent variable $z$ and normalized to $\sum_{i=0}^{m}w_i(z)=1$ $\forall z$. It is important to condition the mixture weights on the latent shape representation $z$, as using identical mixture weights for modelling all shapes is too restrictive. Consequently, the first part of \cref{eq:basic_elbo} becomes
\begin{align} \label{eq:mixture_elbo_reconstruction}
    \mathcal{L}_D =& \underset{z\sim p_{\phi}}{E} \Big[-
    \log \Big(\sum_{i=0}^{m}w_i(z) \mathcal{N}(f^{-1}_{\theta_i}(x); \mu(z), \Sigma(z)) \nonumber \\&
    * \det|J_{f^{-1}_{\theta_i}}(x;z)| \Big) \Big]
\end{align}
This objective leads to specialization of individual \acp{nf}. This is proved in the supplement. \\
\noindent{\textbf{Latent shape representation $z$.}} We model the conditional distribution $p_{\phi}(z|X)$ given a point cloud $X$ as a normal distribution $\mathcal{N}(z;\mu(X), \Sigma(X))$. The mean $\mu(X)$ and the diagonal covariance matrix $\Sigma(X)$ are parameterized by a permutation invariant version of PointNet \cite{qi2017pointnet}. \\
\noindent{\textbf{Learned prior distribution $p_{\psi}(z)$.}} Ideally, the prior over latent representations of point clouds matches the marginal distribution of latent representations of real point clouds after the training, as this enables generating realistic point clouds at inference time. Therefore, we use a learned prior distribution \cite{chen2016variational} parameterized by a discrete \ac{nf} $g_{\psi}$ based on coupling layers with parameters $\psi$. During generative modelling of point clouds we use an unconditional prior distribution. However, on \ac{svr} we condition the prior distribution on a latent representation of the image/view which is produced by a ResNet18 \cite{He_2016_CVPR}. This is achieved using a conditional \ac{nf} (\cref{section:background}). Combining this with our parameterization of the approximate posterior distribution yields the following prior loss:
\begin{align}\label{eq:mixture_elbo_prior}
    \mathcal{L}_{Prior} = &-\mathcal{H}\left( p_{\phi} \right) - \underset{z\sim p_{\phi}(z|X)}{E} \left[ \log \left( p_{\psi}(z) \right) \right] \nonumber \\
    = &-\frac{d}{2} \log \left( 2\pi \right) - \frac{1}{2} \sum_{i=1}^d \log \left( \Sigma(X) \right) \\ &- \underset{z\sim p_{\phi}}{E} \left[ \log(\mathcal{N}(g_{\psi}^{-1}(z)))+ \log( \det|J_{g_{\psi}^{-1}} (z)|) \right] \nonumber
\end{align}
\noindent{\textbf{Optimizing mixtures of \acp{nf}.}} Point clouds typically only cover the surface of a 3D shape, thus are two-dimensional. We follow the common practice of adding Gaussian noise ($\mu = 0$, $\sigma = 0.02$) to point clouds during training \cite{yang2019pointflow, klokov2020discrete, kim2020softflow}. This stabilizes the training as transforming a 3D Gaussian into a 2D distribution yields a pathological training objective \cite{NEURIPS2020_05192834}. Moreover, when training a mixture of \acp{nf} it is necessary to initially enforce a uniform prior on the mixture weights $w_i(z)$ which encourages each \ac{nf} to spread its probability mass over the entire shape. Otherwise, the mixture of \acp{nf} can get stuck in suboptimal solutions where the regions that a \ac{nf} is responsible for are highly disjoint or the model learns to only make use of one \ac{nf}. We implement this as a hard prior where we fix $w_i(z) = w_0$ during a warm-up period. Empirically, we found 5 epochs to be sufficient.

%% file: figures/2d_toy_example/toy_example.tex
\begin{figure}
\raggedleft
\includegraphics[width =.09\textwidth]{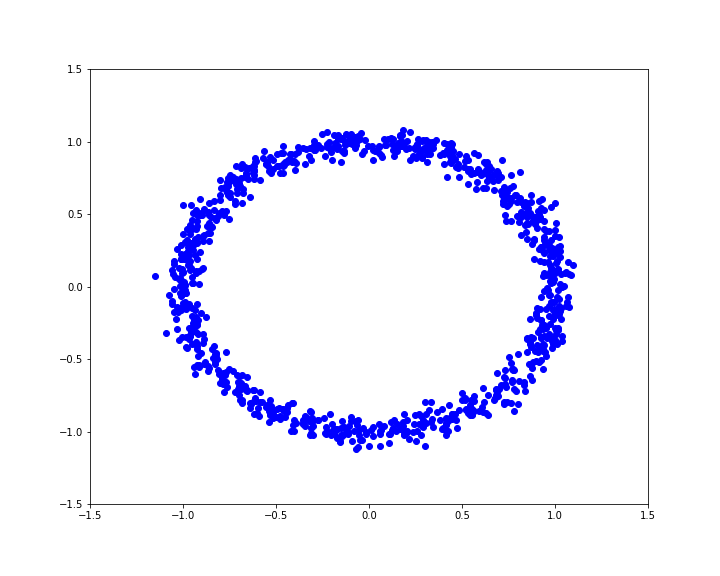} 
\includegraphics[width =.09\textwidth]{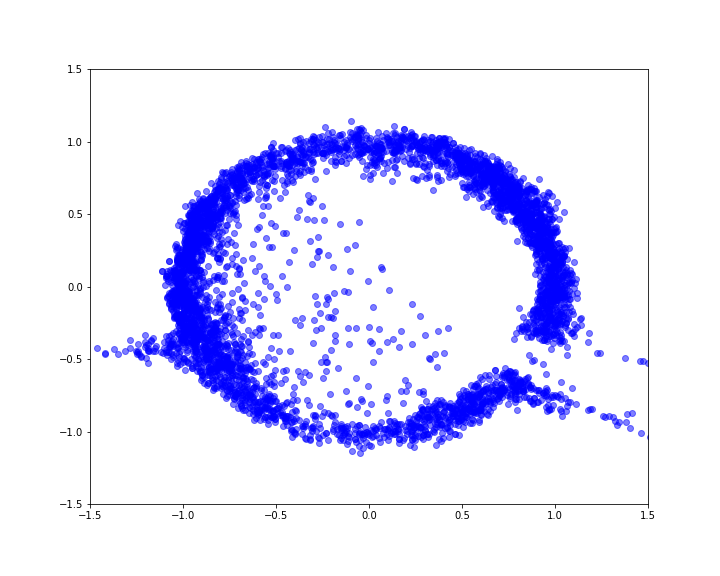} 
\includegraphics[width =.09\textwidth]{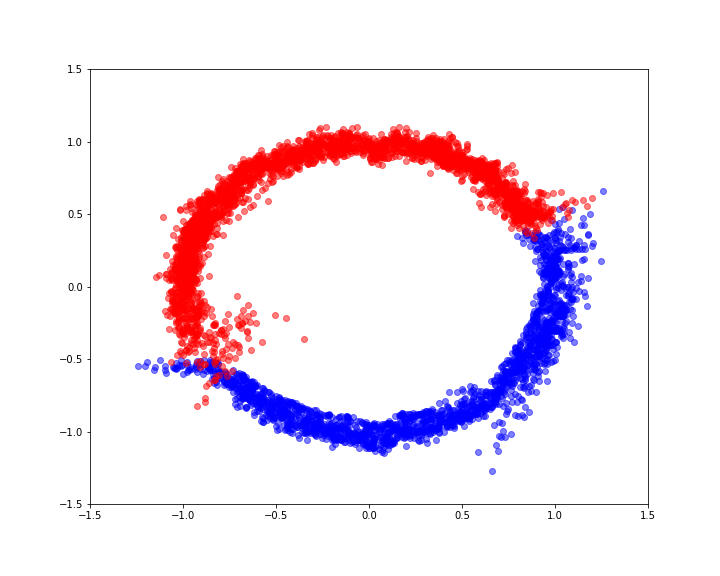} 
\includegraphics[width =.09\textwidth]{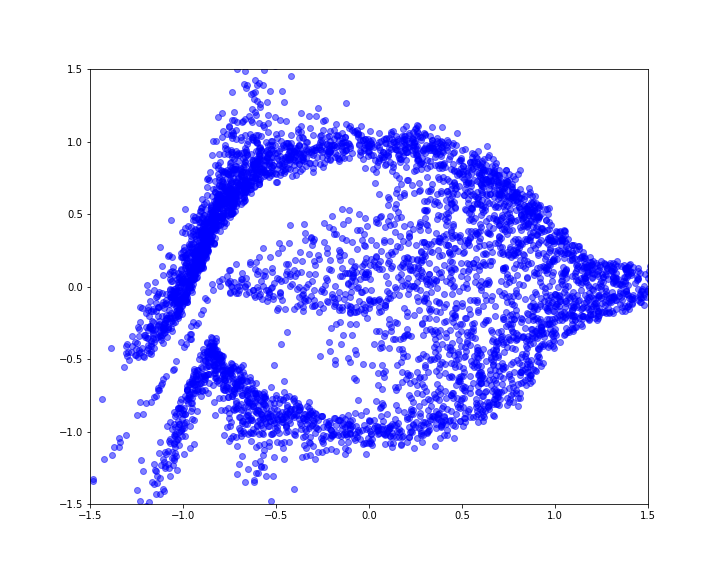} 
\includegraphics[width =.09\textwidth]{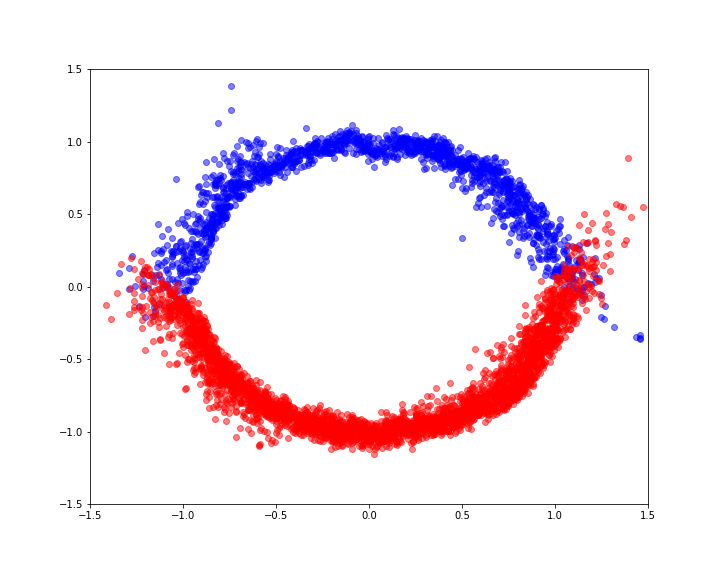} 
\caption{Toy example. Fitting ground truth distribution (1.) using a NF with 4/2 (2. \& 3. / 4. \& 5.) coupling layers. We use Mixtures of \acp{nf} (2 NFs) (3. \& 5.) - color indicates the NF - and a single NF (2. \& 4.). Hyperparameters of the coupling layers chosen such that the single \ac{nf} has more parameters.}
\label{fig:toy_example}
\vspace{-4mm}
\end{figure}
% \end{document}

%% file: sections/04_experiments.tex
\label{sex:experiments}
We evaluate the performance of our model on three  tasks: point cloud generation (\cref{subsec:exp_generative}), autoencoding (\cref{subsec:exp_autoencoding}), and \ac{svr} (\cref{subsec:sv_rec}). Further, we show that mixtures of \acp{nf} learn to specialize in a semantically meaningful and consistent manner (\cref{section:rotation_invariance}) and yield increasing benefit when decreasing the model size (\cref{section:small_models}).

\noindent{\textbf{Dataset.}} We conduct all the experiments using the normalized version of ShapeNet dataset \cite{chang2015shapenet} provided by \cite{klokov2020discrete}.  For more details, please refer to the supplementary material.
%In our autoencoding experiments we use the ShapeNetCore.v2, which contains $55$k point clouds subdivided into $55$ classes. As for point cloud generation, we follow \cite{yang2019pointflow} and focus on three categories of the ShapeNet \cite{chang2015shapenet} dataset: \textit{airplanes}, \textit{cars}, and \textit{chairs}. Finally, for single-view reconstruction we adopt the dataset from \cite{choy20163d}, which contains renders of shapes from the $13$ classes of ShapeNetCore.v1. For each shape 24 images at a resolution of $137 \times 137$ are rendered from random viewpoints. The ground truth point clouds are obtained sampling from the original meshes. We randomly split per class in dataset into 70/10/20 proportion distributing to train/validation/test set for generation and autoencoding, for single-view reconstruction, we use the same train/test split from \cite{choy20163d}. All experiments regarding generation and autoencoding are conducted on the normalized dataset provided by \cite{klokov2020discrete}. Similarly, for single-view reconstruction the models are trained on normalized data. However, we scale the data into a unit radius sphere during evaluation to ensure comparability with related work.

\noindent{\textbf{Evaluation Metrics.}}
Following the evaluation proposed by \cite{achlioptas2018learning,klokov2020discrete,yang2019pointflow,fan2017point}, we measure the quality of reconstructed shapes, in \cref{subsec:exp_autoencoding} and \cref{subsec:sv_rec}, in terms of Chamfer Distance (CD) and Earth Mover’s Distance (EMD). However, these metrics were demonstrated to have severe limitations due to their sensitivity to outliers \cite{tatarchenko2019single,achlioptas2018learning}. For this reason, we include the more robust F1-score \cite{knapitsch2017tanks} (F1) that measures the percentage of points that are correctly reconstructed \cite{tatarchenko2019single}, \ie{} the euclidean distance between each predicted point and the ground truth under a certain threshold $\tau$. Following prior work \cite{achlioptas2018learning,yang2019pointflow,klokov2020discrete}, we evaluate generative modelling performance in \cref{subsec:exp_generative} by comparing generated and reference point clouds with the following metrics: \\
%\begin{itemize}
\indent i) The \textit{Jensen-Shannon divergence} (JSD) measures the similarity between two marginal point distributions obtained by taking the union of all generated (or reference) point clouds, and discretizing them to a voxel grid. \\
\indent ii) \textit{Coverage} (COV) measures the fraction of test point clouds that are matched to at least one generated point cloud. Matching pairs are based on either CD, EMD or F1.\\
\indent iii) The \textit{Minimum matching distance} (MMD) is the average distance of test point clouds to their nearest neighbor in the generated set according to CD, EMD or F1. Note that, unlike CD/EMD, F1 is larger for more similar point clouds.\\
\indent iv) \textit{1-nearest neighbour accuracy} (1-NNA) is the leave-one-out accuracy of the 1-NN classifier in identifying generated and reference point clouds. The nearest neighbor is computed using either CD, EMD or F1.\\
%\end{itemize}
For more details we refer the reader to \cite{achlioptas2018learning,yang2019pointflow}. While JSD and COV measure the diversity of the generated samples, MMD and 1-NNA aim at quantifying their perceptual quality. Prior work only uses CD and EMD for computing MMD. However, we observe that matches of MMD based on CD and EMD are highly dissimilar (see \cref{subsec:exp_generative}). Consequently, MMD is unlikely to reflect the quality of high frequency components of generated point clouds. Therefore, we also include F1 into our evaluation and, further, evaluate generative modeling using the FID score \cite{heusel2017gans} which quantifies both, perceptual quality and diversity. While more advanced measures that disentangle perceptual quality and diversity \cite{naeem2020reliable} exist, here we rely on the more well-known FID score which has been shown to correlate well with human perceptual scores on images. We compute the FID score on the features extracted by a DGCNN \cite{dgcnn} pretrained on ShapeNet provided by \cite{tao2020}.

\noindent{\textbf{Experimental setup.}} We use the same configuration of the encoder and prior as in DPF \cite{klokov2020discrete}. As in \cite{klokov2020discrete} we set the size $D$ of the latent shape representation $z$ to 128 for generation and 512 for both autoencoding and \ac{svr}. The decoder is a mixture of $M$ \acp{nf} with each flow containing $N$ conditional affine coupling layers. These coupling layers compute scaling and translation, as discussed in \cref{section:background}, using  two fully connected layers with $H$ hidden neurons, batch normalization and Swish activation functions. We modulate scaling and translation using an encoding of z according to FiLM \cite{perez2018film, klokov2020discrete}. To ensure comparability with DPF \cite{klokov2020discrete}  we enforce our method to have similar parameter count. Therefore, we reduce $N$ and $H$ in each of the $M$ flows. In all experiments we follow \cite{klokov2020discrete} and set $N=63$ and $H=64$ for the single-flow model. A detailed description of the selection of $H$ and $N$ for mixtures of \acp{nf} can be found in the supplementary materials. We further refer to the supplement for an ablation study on the number $M$ of mixture components. %While we find that the size of the mixture does not have a strong impact on autoencoding performance (any $M>1$ yields a similar increase), $M=4$ performs well across all categories. Consequently, use $M=4$ in our experiments.
This ablation study found that $M=4$ performs well across all categories. Consequently, use $M=4$ in our experiments.

%Analogously to prior work, we use point clouds with 2048 points during training and evaluation of generative modelling and autoencoding, whilst our  \acp{svr} model is evaluated on 2500 points. We follow prior work and evaluate generative modeling 10 times and report the average results.
\noindent{\textbf{Baselines.}} On autoencoding and generation, we compare our method with existing models for point clouds including recent flow-based methods such as \ac{dpf} \cite{klokov2020discrete} and PointFlow \cite{yang2019pointflow} as well as other popular works such as AtlasNet \cite{groueix2018papier} and latent-GAN \cite{achlioptas2018learning}. We retrained DPF \cite{klokov2020discrete} using the official implementation published by the authors due to the lack of pre-trained models. Results of other works are either obtained from \cite{klokov2020discrete} or using a pretrained model provided by the corresponding authors. On \ac{svr}, we compare our results against the most similar work in the literature, \ie methods that reconstruct a shape from an image in form of a point cloud. This includes: AtlasNet \cite{groueix2018papier}, DCG \cite{wang2019deep}, Pixel2Mesh \cite{wang2018pixel2mesh} and \ac{dpf} \cite{klokov2020discrete}.

\noindent{\textbf{Oracle.}} Similar to DPF \cite{klokov2020discrete}, we provide an "oracle" to quantify an upper bound on the performance of our model. In the evaluation of generative modeling the oracle compares a set point clouds obtained from the test set with one obtained from the training set. During the evaluation of autoencoding and \ac{svr}, the oracle provides a point cloud obtained by sampling from the ground truth.

\noindent{\textbf{Optimization.}} Details can be found in the supplementary.
%We train all our models on Nvidia Titan RTX using ADAM \cite{kingma2015adam} for 1450 (generation), 1050 (autoencoding) or 36 (\ac{svr}) epochs using a batch size of 36. We start each training with a learning rate of $2.56 \cdot 10^{-4}$ and divide it by four at certain epochs (generation: 800, 1200, 1400; autoencoding: 400, 800, 1000; \ac{svr}: 20, 30, 35).

\subsection{Generative Modeling}
\label{subsec:exp_generative}

\noindent{\textbf{Experimental setup.}} We evaluate how well mixtures of \acp{nf} fit the distribution of point clouds. In line with prior work, we train a mixture of \acp{nf} on the three categories car, chair and airplane. We compute the evaluation metrics mentioned in \cref{section:experiments} between the test set and a set of generated point clouds of equal size. Each point cloud comprises 2048 points. We repeat the evaluation 10 times and report the average results. Results including standard deviation can be found in the supplementary.

\noindent{\textbf{Results.}}  Table \ref{table:comparison_related_work_generation} shows the results of this experiment. Mixtures of \acp{nf} obtain the best results regarding JSD, COV-CD/EMD and 1-NNA-EMD. Latent-GAN-CD/EMD \cite{achlioptas2018learning} shows strong performance on MMD-CD/EMD while performing poorly on EMD/CD as expected since it is optimized using CD/EMD. Compared with other flow-based models, for which we also evaluate the FID score, mixtures of \acp{nf} yield better performance than \ac{dpf} \cite{klokov2020discrete} across most metrics and clearly outperform PointFlow \cite{yang2019pointflow}. We also compare our approach with DPF based on the F1-score ($\tau = 10^{-4}$) where we outperform DPF 7 out of 9 times. Fig. \ref{fig: qualitative_results_of_generation} shows qualitative examples. Most importantly, we observe that each component of our mixture of \acp{nf} specializes in a distinct subregion of the shape. Interestingly, this specialization generalizes across different shapes.

%\noindent{\textbf{Discuss need metrics analysis or not. annotated temporarily}}
\noindent{\textbf{Analysis of MMD.}} The role of the metrics MMD-CD/EMD is to quantify the perceptual quality of the generated point clouds. However, a qualitative examination reveals that matches between generated point clouds and test samples are dissimilar (see supplement). More, an ablation study in \cref{section:small_models} regarding the performance of models with fewer parameters reveals that the quality of the reconstructed point clouds clearly degrades with decreasing model size. However, in the supplement we further demonstrate that MMD-CD/EMD remain largely unchanged for smaller models despite an obvious degradation in the perceptual quality.
 
\def\gefiga#1{{\includegraphics[clip,trim=8cm 4.5cm 8cm 7.5cm,width=0.30\linewidth]{figures/generation/#1.png}}}
\def\gefigb#1{{\includegraphics[clip,trim=8cm 4cm 8cm 7cm,width=0.30\linewidth]{figures/generation/#1.png}}}
\def\gefigc#1{{\includegraphics[clip,trim=8cm 3cm 8cm 6.5cm,width=0.30\linewidth]{figures/generation/#1.png}}}
\begin{figure}
	\centering	
	\setlength{\arrayrulewidth}{.5pt}%
	\setlength{\tabcolsep}{1pt}
	\renewcommand{\arraystretch}{0.5}
	\begin{tabular}{ccc}
		% airplane
		\gefiga{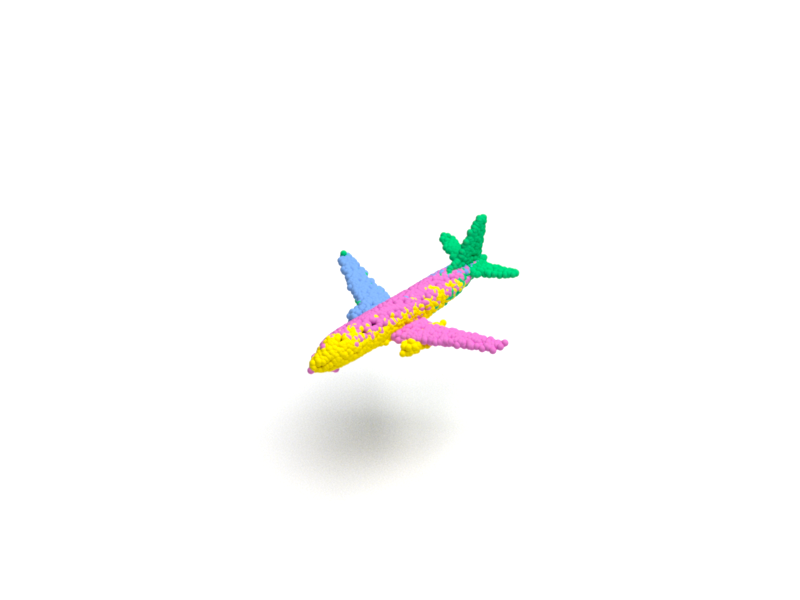} &
		\gefiga{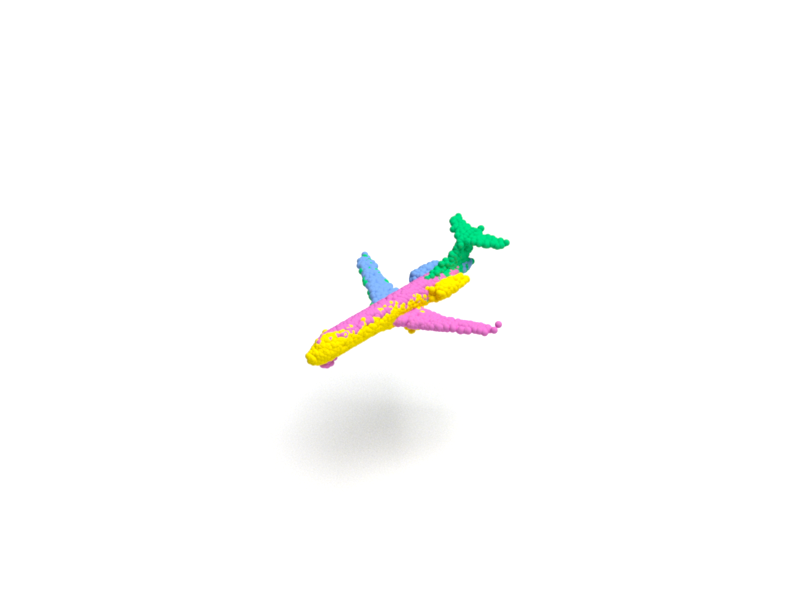} &
		\gefiga{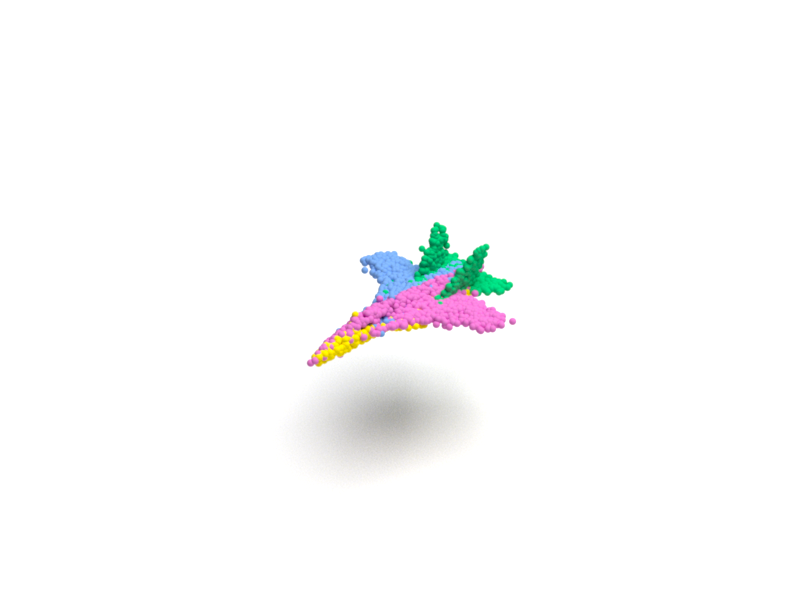} \\
		% chair
		\gefigb{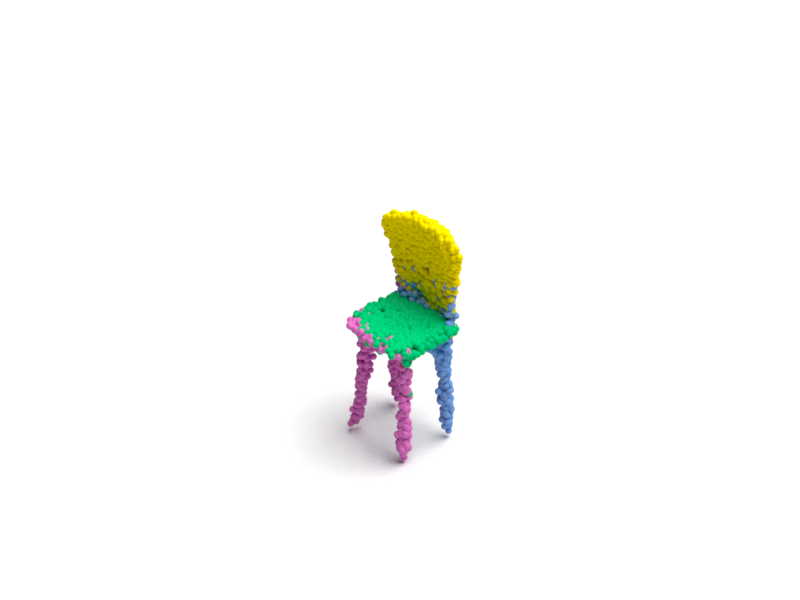} &
		\gefigb{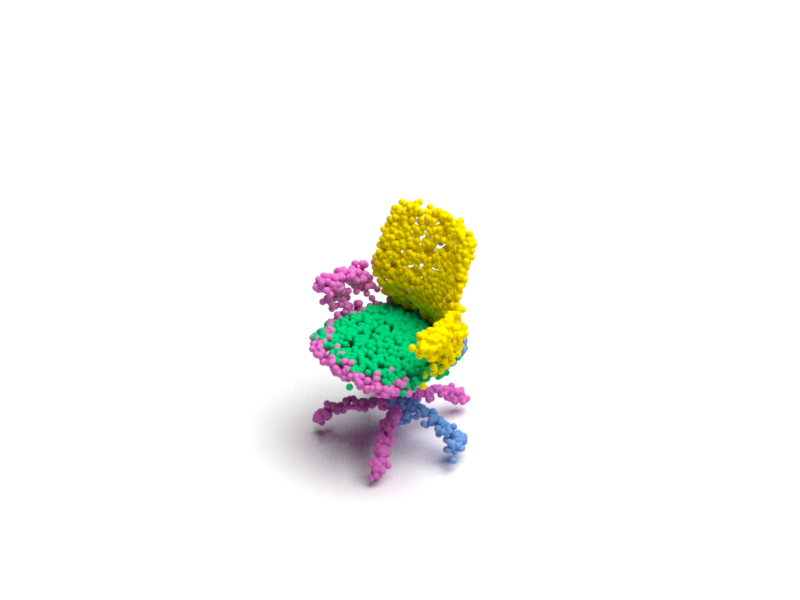} &
		\gefigb{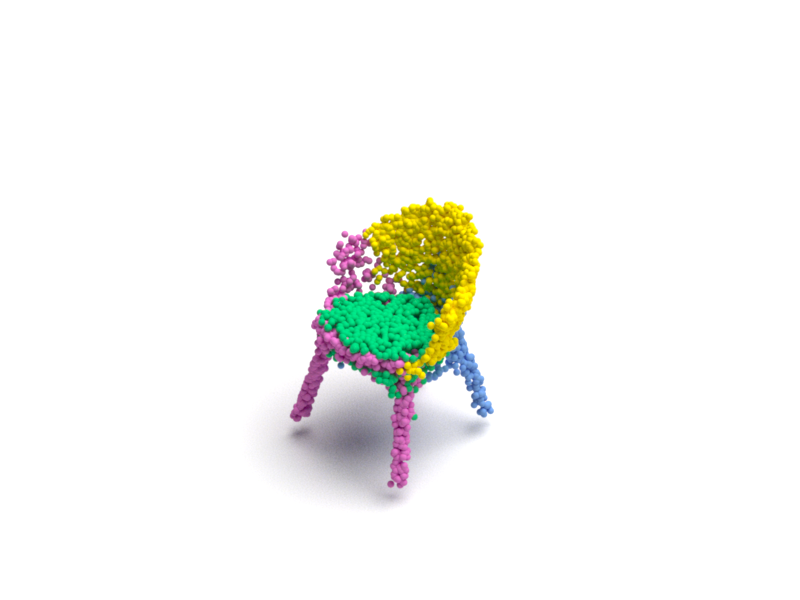} \\
	    % car
		\gefigc{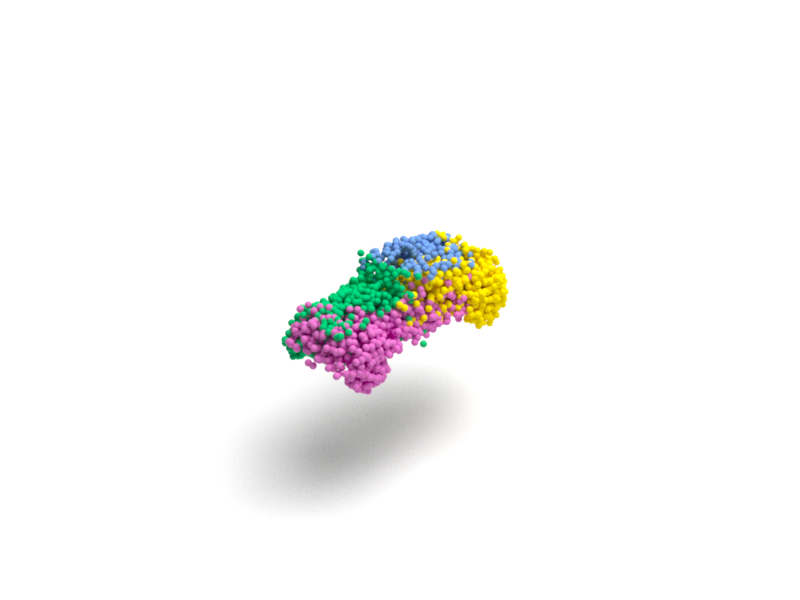} &
		\gefigc{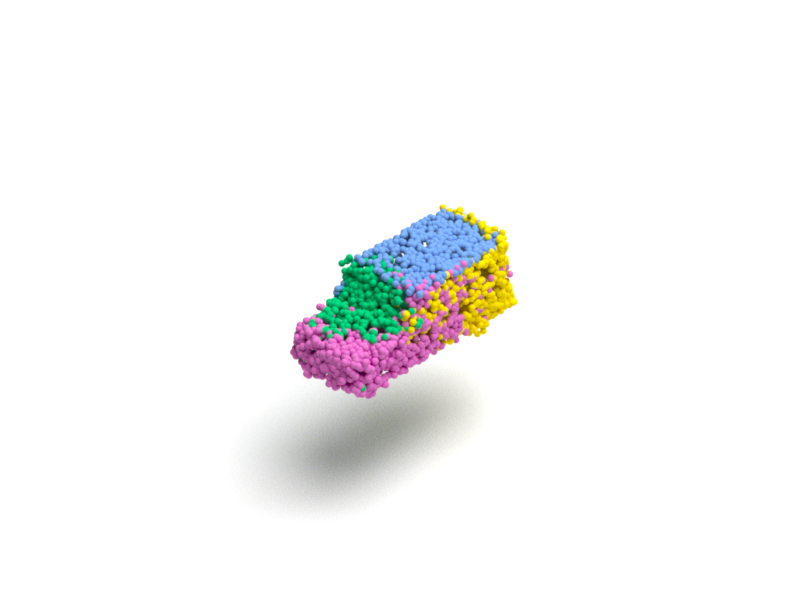} &
		\gefigc{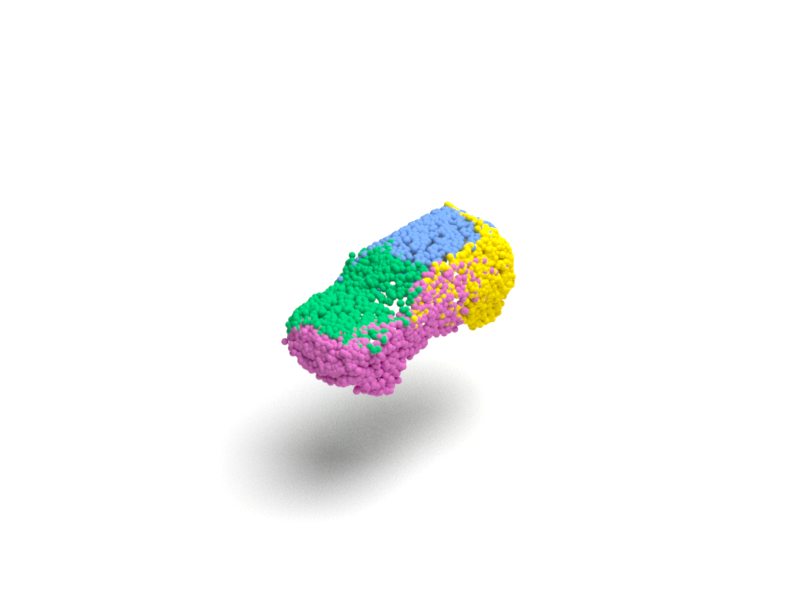} \\

	\end{tabular}
	\caption{Qualitative results of generation. From above to below, 
	we show samples generated with mixtures of \acp{nf} using models trained on the airplane, chair and car categories of ShapeNet \cite{chang2015shapenet}.}
	\label{fig: qualitative_results_of_generation}
\end{figure}

\begin{table*}[h]
\begin{center}
 \begin{tabular}{| c | c | c | c | c | c | c | c | c | c | c | c | c |} 
 \hline
 & & JSD $\downarrow$ & \multicolumn{3}{|c|}{MMD} & \multicolumn{3}{|c|}{COV $\uparrow$} & \multicolumn{3}{|c|}{1-NNA $\downarrow$}& FID $\downarrow$\\
 Category & Method & & CD $\downarrow$ & EMD $\downarrow$ & F1 $\uparrow$ & CD & EMD & F1 & CD & EMD& F1 &\\ [0.5ex] 
 \hline\hline
 & l-GAN-CD \cite{achlioptas2018learning} & 2.76 & \textbf{5.69} & 5.16 & - & 39.5 & 17.1 & - & 72.9 & 92.1 & - & -\\ 
 
 & l-GAN-EMD \cite{achlioptas2018learning} & 1.77 & 6.05 & \textbf{4.15} & -& 39.7 & 40.4 & - & 75.7 & 73.0 & - & -\\

 Airplane & PointFlow \cite{yang2019pointflow} & 1.42 & 6.05 & 4.32 & - & 44.7 & \textbf{48.4} & - & 70.9 & 68.4 & - & 0.68\\
 
 & DPF \cite{klokov2020discrete} & \underline{1.14} & \underline{6.03} & 4.27 & \textbf{50.84} & \underline{46.4} & 48.2 & \textbf{42.7} & \underline{70.3} & \underline{67.5} & 72.7 & \underline{0.16}\\
 
 & Ours 4 Flows & \textbf{1.03} & 6.06 & \underline{4.26} & 50.11 & \textbf{46.5} & \textbf{48.4} & 42.3 & \textbf{70.1} & \textbf{66.9}& \textbf{71.7} & \textbf{0.15}\\ 
 \hline
 & Oracle & 0.50 & 5.97 & 3.98 & 75.48 & 51.4 & 52.7 & 94.3 & 49.8 & 48.2 & 50.2 & 0.07\\ [1ex] 
 
 \hline\hline
 & l-GAN-CD \cite{achlioptas2018learning} & 3.65 & \textbf{16.66} & 7.91 & - & 42.3 & 17.1 & - & 68.5 & 96.5 & - & -\\ 
 
 & l-GAN-EMD \cite{achlioptas2018learning} & \textbf{1.27} & \underline{16.78} & \textbf{5.75} & - & 44.3 & 43.8 & - & 66.6 & \underline{67.8} & - & -\\

 Chair & PointFlow \cite{yang2019pointflow} & 1.51 & 17.15 & 6.20 & - & 43.3 & \textbf{46.5} & - & 67.0 & 70.4 & - & 0.29\\
 
 & DPF \cite{klokov2020discrete} & \underline{1.37} & 17.24 & 6.13 & 19.63 & \underline{45.1} & 46.0 & 34.7 & \textbf{64.8} & 68.2 & 67.7 & \textbf{0.26}\\
 
 & Ours 4 Flows & 1.45 & 17.30 & \underline{6.11} & \textbf{21.08} & \textbf{45.2} & \textbf{46.5} & \textbf{39.2} &  \underline{65.3} & \textbf{65.6}  & \textbf{62.2} & \textbf{0.26}\\ 
 \hline
 & Oracle & 0.49 & 16.39 & 5.71 & 49.14 & 52.8 & 53.4 & 99.8 & 49.7 & 49.6 & 49.5 & 0.08\\ [1ex] 
 
 \hline\hline
 & l-GAN-CD \cite{achlioptas2018learning} & 2.65 & \textbf{8.83} & 5.36 & - & 41.3 & 15.9 & - & \textbf{62.6} & 92.7 & - &  -\\ 
 
 & l-GAN-EMD \cite{achlioptas2018learning} & 1.31 & \underline{9.00} & \textbf{4.40} & - & 38.3 & 32.9 & - & \underline{65.2} & \textbf{63.2} & - & -\\

 Car & PointFlow \cite{yang2019pointflow} & 0.59 & 9.53 & 4.71  & - & \textbf{42.3} & 35.8 & - & 70.1 & 74.2 & - & 0.20\\
 
 & DPF \cite{klokov2020discrete} & \underline{0.57} & 9.67 & \underline{4.60} & 18.11 & 40.8 & \underline{43.7} & 37.7 & 71.3 & 66.0 & 69.1 & \underline{0.11}\\
 
 & Ours 4 Flows & \textbf{0.55} & 9.50 & 4.62 & \textbf{18.20} & \underline{41.4} & \textbf{43.8} & \textbf{37.8} & 69.0 & \underline{64.8} & \textbf{66.1} & \textbf{0.10}\\ 
 \hline
 & Oracle & 0.37 & 9.24 & 4.56 & 35.03 & 52.8 & 52.7 & 99.5 & 50.9 & 50.5 & 49.1 & 0.05\\ [1ex] 
 \hline\hline
 & l-GAN-CD \cite{achlioptas2018learning} & 3.02& \textbf{10.39} & 6.14&- & 32.6&32.3 &- &70.7 &94.3 & - & -\\
 & l-GAN-EMD \cite{achlioptas2018learning} &1.45 &\underline{10.61} & \textbf{4.77} &- &40.8 & 39.0&- &69.2 &68 &- &-\\
 Average & PointFlow \cite{yang2019pointflow} &1.17 &10.91 &5.08 & -&43.4 &43.6 &- &69.3 &71 &- &0.39\\
 & DPF \cite{klokov2020discrete} &\underline{1.03} & 10.98& \underline{5.00} & 29.5&\underline{43.9} & \underline{46.0} & 37.7 & \underline{68.8} & \underline{67.2} & 69.1 & \underline{0.18}\\
  & Ours 4 Flows & \textbf{1.01} & 10.95 & \underline{5.00} & \textbf{29.80} & \textbf{44.4} & \textbf{46.2} & \textbf{39.8} & \textbf{68.1} & \textbf{65.8} & \textbf{66.7} & \textbf{0.17} \\
  \hline
  & Oracle & 0.45 & 10.53 & 4.75 & 53.22 & 52.3 & 52.9 & 97.9 & 50.1 & 49.4 & 49.6 & 0.07 \\ [0.3ex] 
 \hline
\end{tabular}
\caption{Generative modeling. JSD, MMD-EMD and MMD-F1 ($\tau=10^{-4}$) are multiplied by $10^2$, MMD-CD is multiplied by $10^4$}
\label{table:comparison_related_work_generation}
\end{center}
\end{table*}

\subsection{Autoencoding}
\label{subsec:exp_autoencoding}

\noindent{\textbf{Experimental setup.}} We evaluate the autoencoding using mixtures of \acp{nf} (4 components) jointly on all categories of ShapeNet \cite{chang2015shapenet}. We report CD, EMD and F1-score ($\tau = 10^{-4}$) by following prior work in comparing test samples with 2048 points with their reconstructions of equal size.

\noindent{\textbf{Results.}} In \cref{table:comparison_realted_work_ae} mixtures of \acp{nf} obtain the highest F1-score and second best CD/EMD. Moreover, we outperform latent-GAN \cite{achlioptas2018learning} and PointFlow \cite{yang2019pointflow} across all metrics. AtlasNet \cite{groueix2018papier} trained with CD as criteria performs the best on CD. As for EMD, we are slightly worse than DPF \cite{klokov2020discrete}. We argue that this is expected since EMD favors evenly distributed point clouds \cite{klokov2020discrete} which is simpler to achieve using a single \ac{nf}. Conversely, we report a lower CD than DPF \cite{klokov2020discrete}. CD prioritizes regions \cite{klokov2020discrete}, this reflects the ability of our model to better capture the local geometry of the shape, as qualitatively shown in \cref{fig:teaser}. By zooming into a specific part of the shape, we can see how our model precisely reconstructs fine-grained geometric details, conversely DPF \cite{klokov2020discrete} tends to get a smoother and noisier shape, this confirming our expectations as explained in the methodology section. For example, looking at the tail of the airplane and the legs of the chair, we can see how our method is able to reconstruct them completely and clearly, while DPF \cite{klokov2020discrete} fails at reconstructing high-frequency regions. %This performance matches our expectation according to the theory section.

% %  Without EMD
% \begin{table*}[h]
% \begin{center}
%  \begin{tabular}{| c | c | c | c | c |} 
%  \hline
%  Method & CD & F1, $\tau=10^{-3}$ & F1, $\tau=10^{-4}$ \\ [0.5ex] 
%  \hline\hline
%  l-GAN-CD \cite{achlioptas2018learning} & 7.07 & &  \\ 
 
%  l-GAN-EMD \cite{achlioptas2018learning} & 9.18 & &  \\

%  AtlasNet \cite{groueix2018papier} & 5.66 & &  \\
 
%  PointFlow \cite{yang2019pointflow} & 7.54 & 93.6& 32.3  \\
 
%  DPF \cite{klokov2020discrete} & 6.17 & &  \\
 
%  Retrained DPF & 6.92 & 93.7 & 34.51 \\
%   4flows & 6.9 & 93.8 & 34.69 \\
%  Best Flow Mixture.. &  &  &  \\ 
%  \hline
%  Oracle & 3.10 &  &  \\ [1ex] 
 
%  \hline
% \end{tabular}
% \caption{Autoencoding. Comparison with related work on full dataset.}
% \label{table:comparison_realted_work_ae}
% \end{center}
% \end{table*}

% With EMD
\begin{table}[h]
\begin{center}
 \begin{tabular}{| c | c | c | c | c |} 
 \hline
 Method & CD $\downarrow$& EMD $\downarrow$& F1 $\uparrow$, $\tau=10^{-4}$\\ [0.5ex] 
 \hline\hline
 l-GAN-CD \cite{achlioptas2018learning} & 7.07 & 7.70 & -\\ 
 
 l-GAN-EMD \cite{achlioptas2018learning} & 9.18 & 5.30 & -\\

 AtlasNet \cite{groueix2018papier} & \textbf{5.66} & 5.81 & - \\
 
 PointFlow \cite{yang2019pointflow} & 7.54 & 5.18 & 32.3 \\

 DPF \cite{klokov2020discrete} & 6.92 & \textbf{4.66} & \underline{34.5}\\
 
 Ours 4 Flows & \underline{6.88} & \underline{4.80} & \textbf{34.8} \\
 \hline
 Oracle & 3.10 & 3.13 & 76.0  \\ [1ex] 
 
 \hline
\end{tabular}
\vspace{0.5em}
\caption{Autoencoding. Comparison with related work on the full ShapeNet dataset \cite{chang2015shapenet}. CD is multiplied by $10^4$, EMD by $10^2$.}
\label{table:comparison_realted_work_ae}
\end{center}
\end{table}

\subsection{Single-view Reconstruction}
\label{subsec:sv_rec}

\noindent{\textbf{Experimental setup.}} We evaluate the ability of mixtures of \acp{nf} to reconstruct point clouds from a single RGB image. At test time we sample from the prior conditioned on the encoding of RGB test images and subsequently sample from the decoder conditioned on this sample. We report CD, EMD and F1-score. Unlike in our experiments on autoencoding \cref{subsec:exp_autoencoding}, here we choose a threshold $\tau = 10^{-3}$ to ensure comparability with prior work \cite{klokov2020discrete}.

\noindent{\textbf{Results.}} Quantitative results can be found in \cref{table:comparison_realted_work_svr}. Generally, we observe that mixtures of \acp{nf} yield at least second best performance across all metrics. AtlasNet \cite{groueix2018papier} outperforms our method for CD, which is expected since it explicitly optimizes this metric. Interestingly, mixtures of \acp{nf} demonstrate the best performance in F1-score, which is regarded as a more faithful metric for perceptual quality \cite{knapitsch2017tanks}. In \cref{fig:qualitative} we show qualitative examples of \ac{svr} using \ac{dpf} \cite{klokov2020discrete} and mixtures of \acp{nf}. We observe that mixtures of \acp{nf} yield sharper reconstructions with particular improvements on complicated geometries, \eg the lamp. Also for \ac{svr} we observe that in mixtures of \acp{nf} each flow learns to be responsible for one part of the shape.

%trim=9cm 4cm 10cm 8cm,
%9cm 4cm 10cm 7cm,
\def\svrfiga#1#2{{\includegraphics[clip, trim=8.5cm 4cm 8cm 7cm, width=0.25\linewidth]{figures/svr/#1/#2.png}}}
\def\svrfigb#1#2{{\includegraphics[clip,trim=8.5cm 4cm 10cm 7cm, width=0.25\linewidth]{figures/svr/#1/#2.png}}}
\def\svrfige#1#2{{\includegraphics[clip,trim=8.5cm 4cm 8cm 7.5cm, width=0.25\linewidth]{figures/svr/#1/#2.png}}}
\def\svrfigc#1#2{{\includegraphics[clip,trim=9cm 3cm 9cm 7cm, width=0.25\linewidth]{figures/svr/#1/#2.png}}}
\def\svrfigd#1#2{{\includegraphics[clip, trim=8.5cm 4cm 8cm 8cm, width=0.25\linewidth]{figures/svr/#1/#2.png}}}
\begin{figure}
	\centering	
	\setlength{\arrayrulewidth}{.5pt}%
	\setlength{\tabcolsep}{0.2pt}
	\renewcommand{\arraystretch}{0.5}
	\begin{tabular}{cccc}
		% ariplane
		%1.2cm 0.5cm 1.5cm 2cm,
		\includegraphics[clip, trim=1cm 1cm 1.5cm 2cm, width=0.2\linewidth]{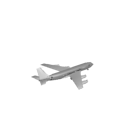} &
		\svrfiga{airplane}{0_gt} &
		\svrfiga{airplane}{0_1f} &
		\svrfiga{airplane}{0_4f} \\
		% chair
		%6cm 2cm 4cm 3cm
		\includegraphics[clip, trim=4cm 2.5cm 4cm 3cm, width=0.2\linewidth]{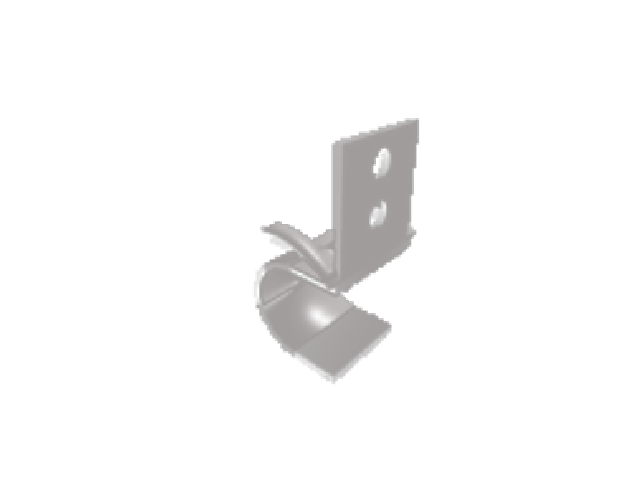} &
		\svrfiga{chair}{2_gt} &
		\svrfiga{chair}{2_1f} &
		\svrfige{chair}{2_4f} \\
		% car
		\includegraphics[clip, trim=4cm 0cm 4cm 4cm, width=0.2\linewidth]{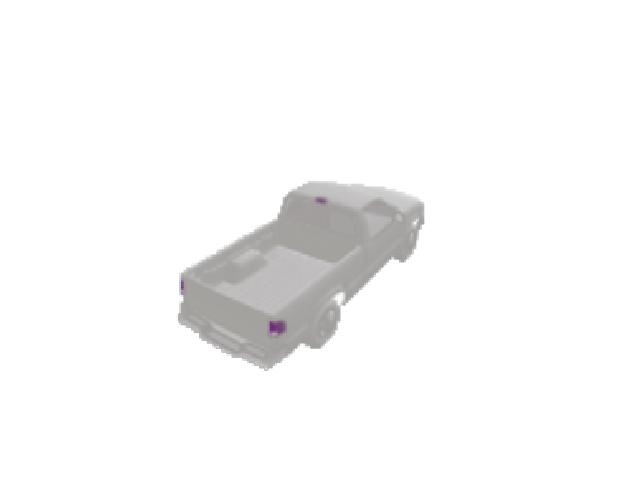} &
		\svrfigc{car}{1_gt} &
		\svrfigc{car}{1_1f} &
		\svrfigc{car}{1_4f} \\
		% lamp
		\includegraphics[clip, trim=4cm 2cm 4cm 2.5cm, width=0.2\linewidth]{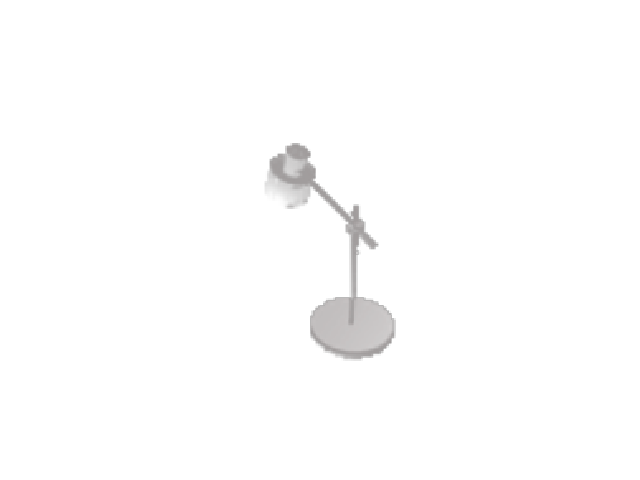} &
		\svrfigd{lamp}{3_gt} &
		\svrfigd{lamp}{3_1f} &
		\svrfigd{lamp}{3_4f} \\
		\footnotesize Input & 
		\footnotesize Ground truth & 
		\footnotesize DPF \cite{klokov2020discrete} & 
		\footnotesize Ours 4 Flows \\
	\end{tabular}
	%\vspace{-1em}
	\caption{Qualitative comparison of \acp{svr} methods on the ShapeNet \cite{chang2015shapenet} test set. On the left, we show the RGB view used as input. We also report the results for DPF \cite{klokov2020discrete}.}
	\label{fig:qualitative}
\end{figure}

% %  without EMD
% \begin{table*}[h]
% \begin{center}
%  \begin{tabular}{| c | c | c | c |} 
%  \hline
%  Method & CD & F1, $\tau=10^{-3}$  & F1, $\tau=10^{-4}$\\ [0.5ex] 
%  \hline\hline
 
%  DPF \cite{klokov2020discrete} & 5.51 & 52.4 &\\
%  Our retrained DPF & 5.8 & 51.97 & 5.88 \\ 
%  Best Flow Mixture.. & 5.67  & 52.21 & 5.93\\ 
%  \hline
%  Oracle & 1.10 & 84.0 &\\ [1ex] 
 
%  \hline
% \end{tabular}
% \caption{Single-view Reconstruction. Comparison with related work on full dataset.}
% \label{table:comparison_realted_work_svr}
% \end{center}
% \end{table*}

% with EMD
\begin{table}[h]
\begin{center}
 \begin{tabular}{| c | c | c | c | c | c |} 
 \hline
 Method & CD $\downarrow$& EMD $\downarrow$& F1 $\uparrow$, $\tau=10^{-3}$ \\ [0.5ex] 
 \hline\hline
 AtlasNet \cite{groueix2018papier} & \textbf{5.34} & 12.54 & \underline{52.2}  \\
 DCG \cite{wang2019deep} & 6.35 & 18.94 & 45.7 \\
 Pixel2Mesh \cite{wang2018pixel2mesh} & 5.91 & 13.80 & - \\
 DPF \cite{klokov2020discrete} & 5.80 & \textbf{11.17} & 52.0  \\
 Ours 4 Flows &\underline{ 5.66} & \underline{11.18} & \textbf{52.3}  \\
 \hline
 Oracle & 1.10 & 5.70 & 84.0 \\ [1ex] 
 
 \hline
\end{tabular}
\vspace{0.5em}
\caption{Single-view Reconstruction. Comparison with related work on 13 categories of ShapeNet \cite{chang2015shapenet}. CD is multiplied by $10^3$, EMD is multiplied by $10^2$.}
\vspace{-5mm}
\label{table:comparison_realted_work_svr}
\end{center}
\end{table}

\subsection{Rotation Invariant Latent Variables}
\label{section:rotation_invariance}
This section qualitatively evaluates whether our mixtures of \acp{nf} can learn to specialize to semantically meaningful regions of the object, which is an interesting property that could aid applications such as point cloud registration.
%rather than  not regions of the euclidean space 

\def\gefiga#1{{\includegraphics[clip,trim=8.5cm 4cm 8cm 7.0cm,width=0.25\linewidth]{figures/rotation_invariant_latents/#1.png}}}
\begin{figure}
	%\centering	
	\setlength{\arrayrulewidth}{.2pt}%
	\setlength{\tabcolsep}{0.20pt}
	\renewcommand{\arraystretch}{0.5}
	\begin{tabular}{cccc}
		% gt
		\gefiga{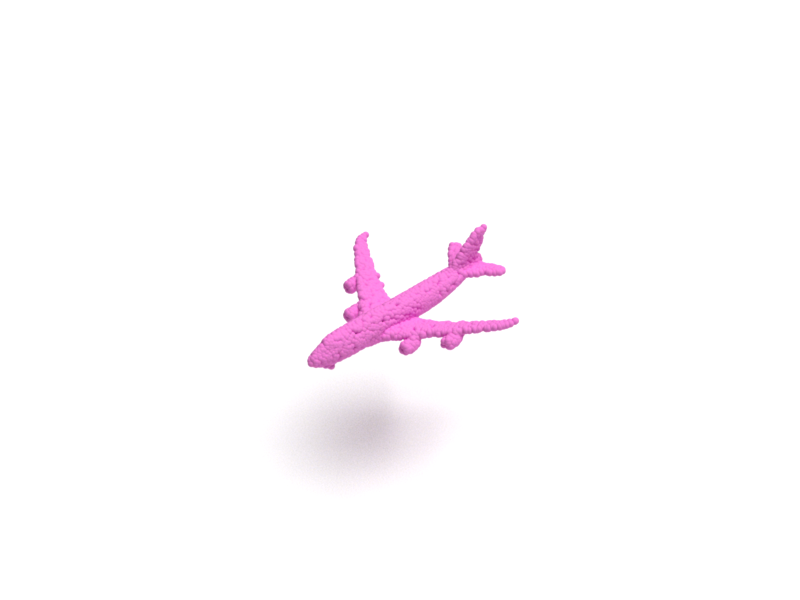} &
		\gefiga{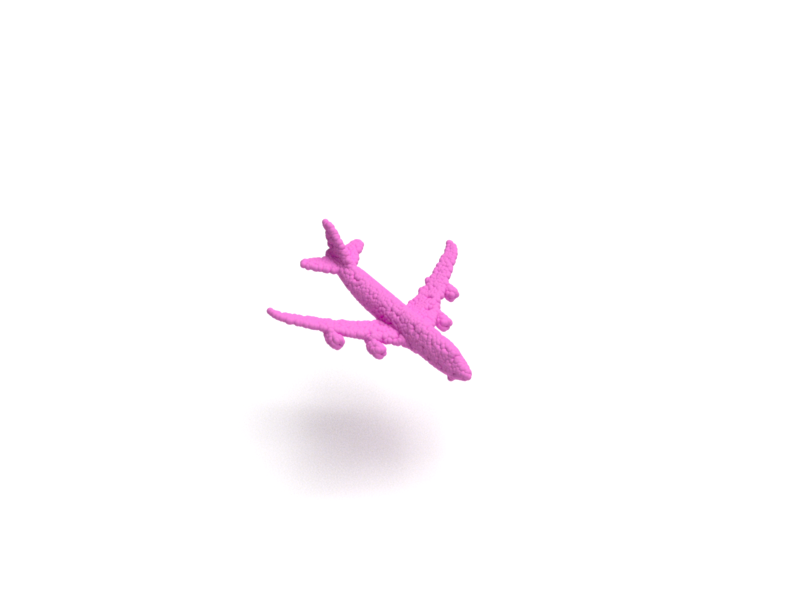} &
		\gefiga{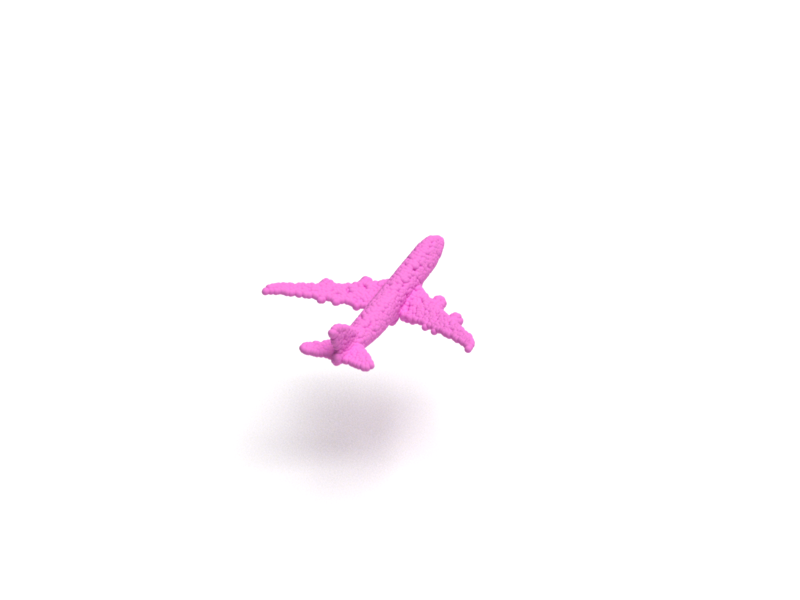} &
		\gefiga{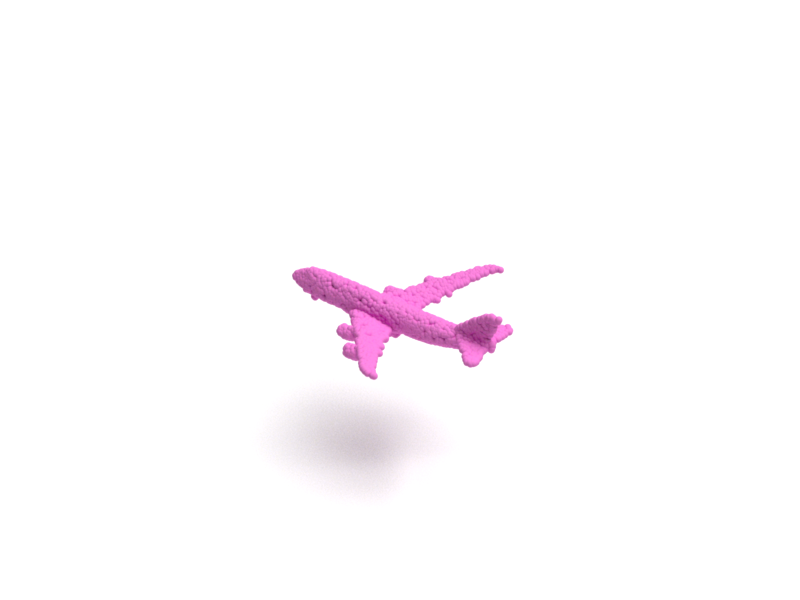} \\
		\gefiga{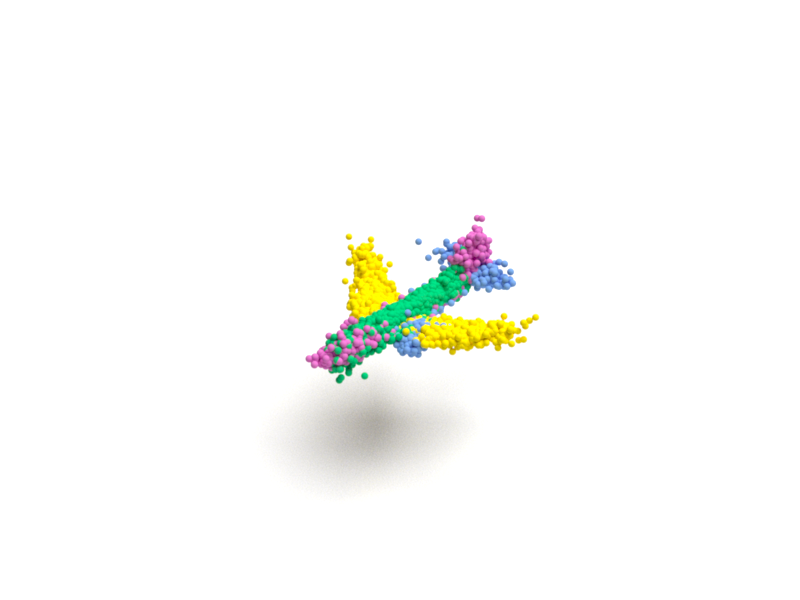} &
		\gefiga{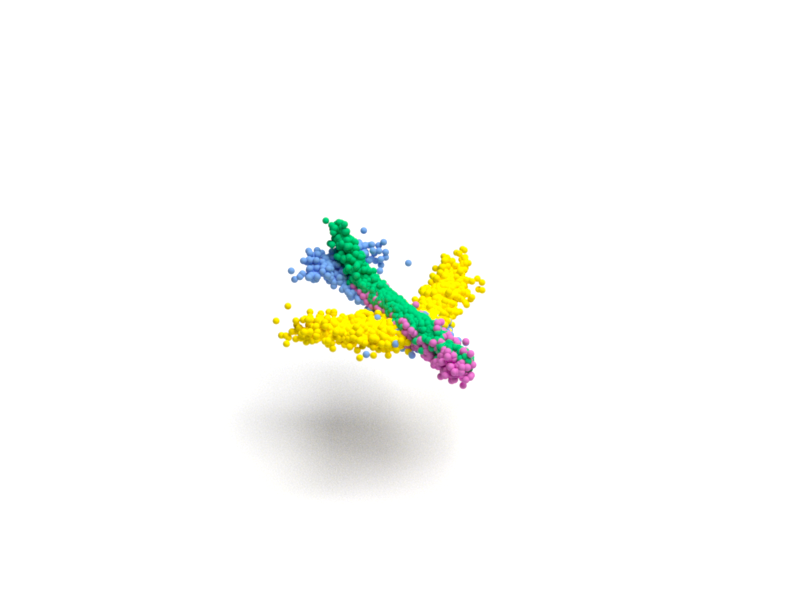} &
		\gefiga{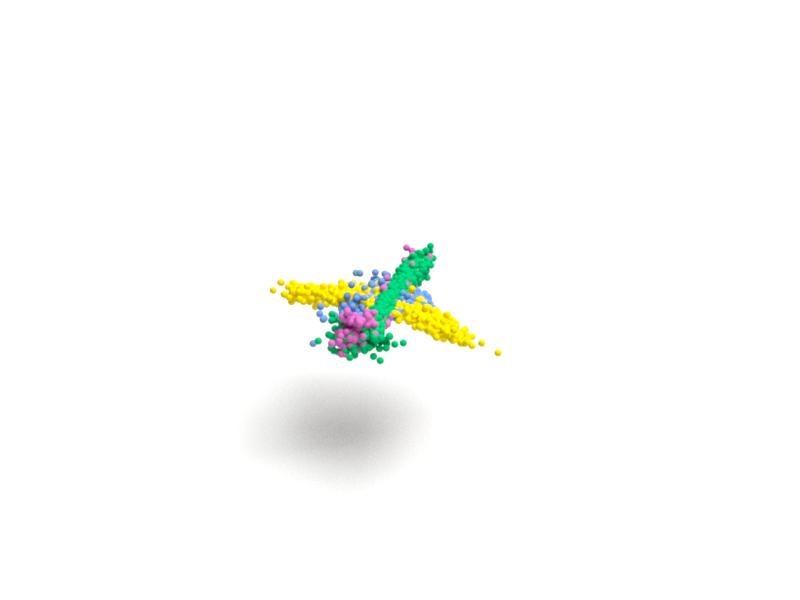} &
		\gefiga{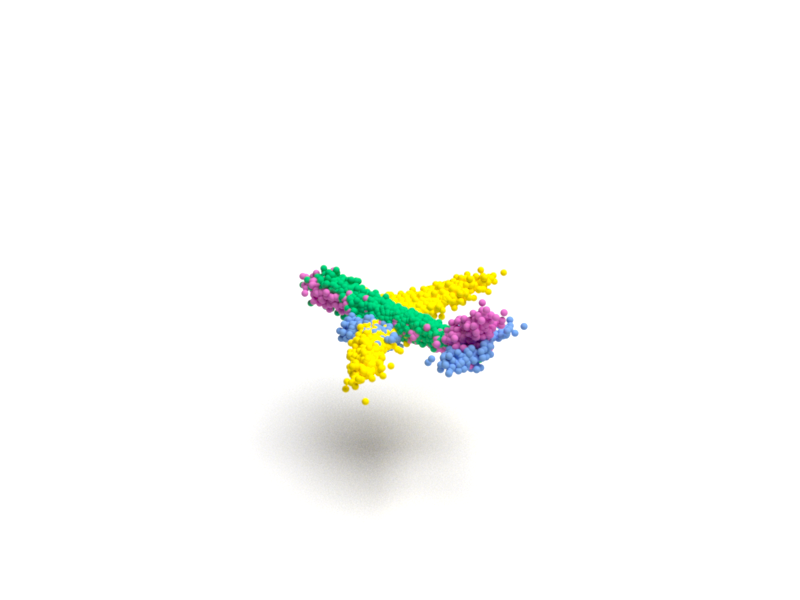} \\
		\gefiga{gt_0} &
		\gefiga{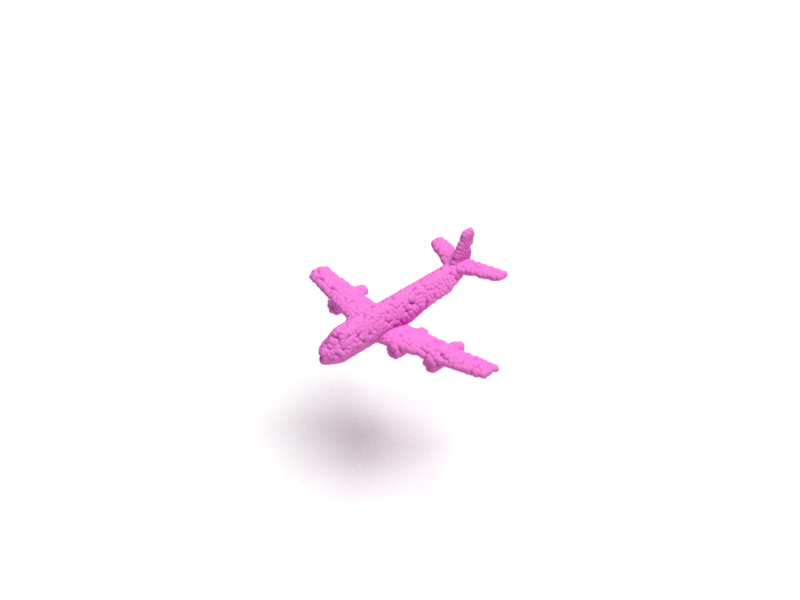} &
		\gefiga{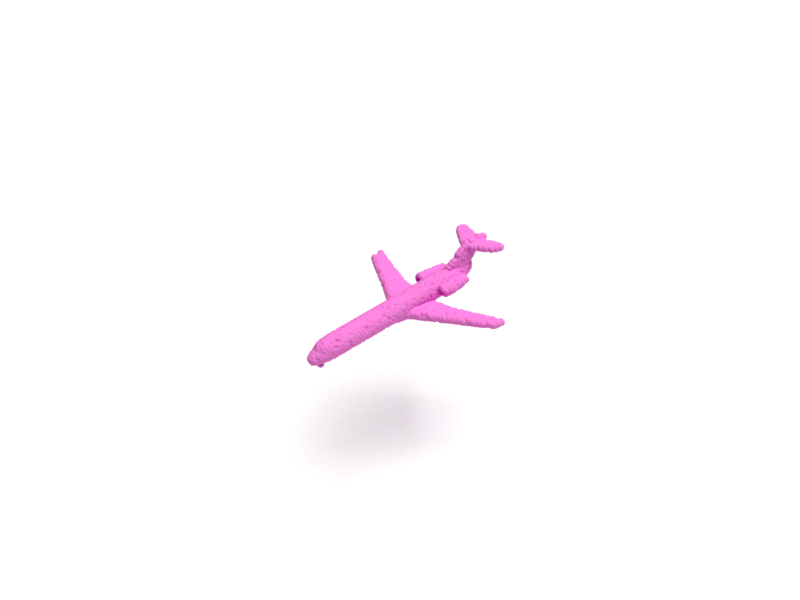} &
		\gefiga{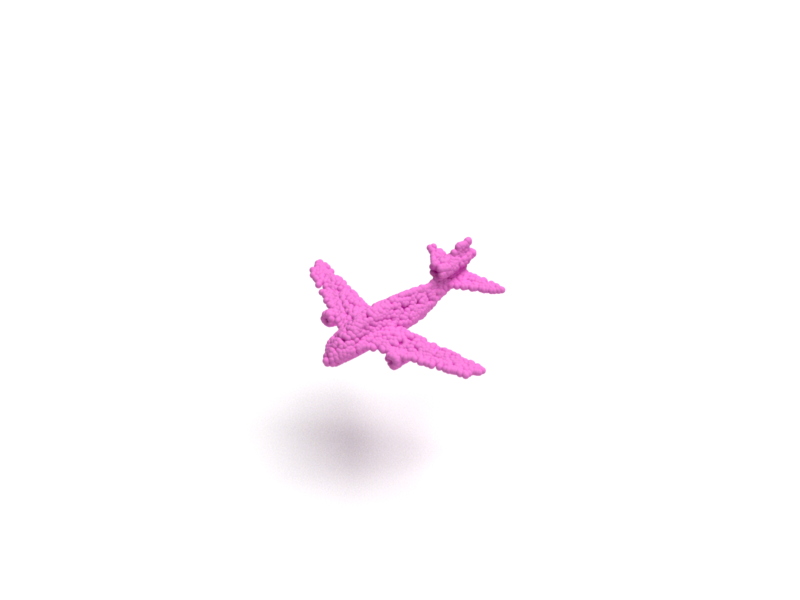} \\
		\gefiga{0} &
		\gefiga{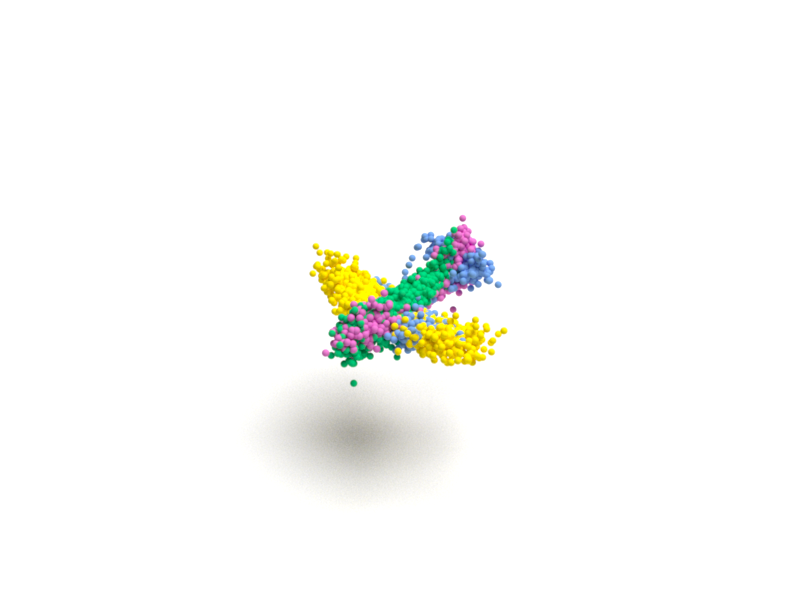} &
		\gefiga{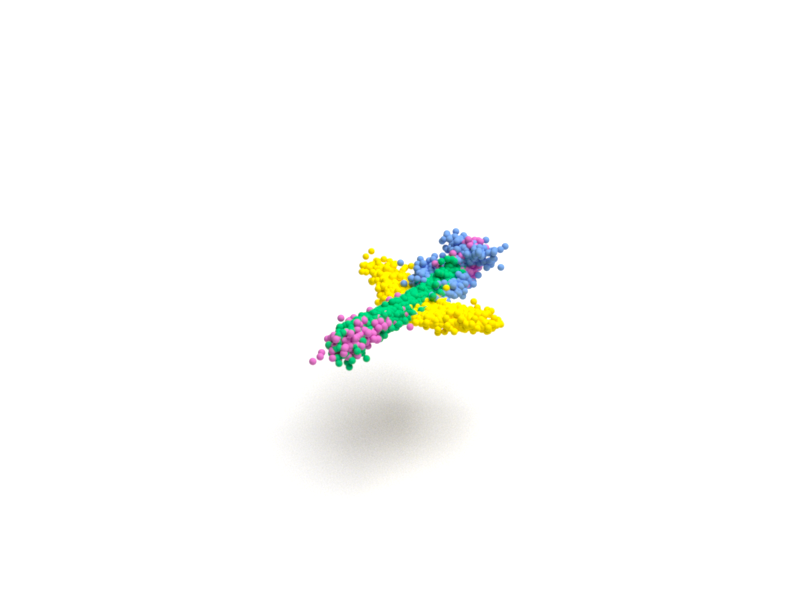} &
		\gefiga{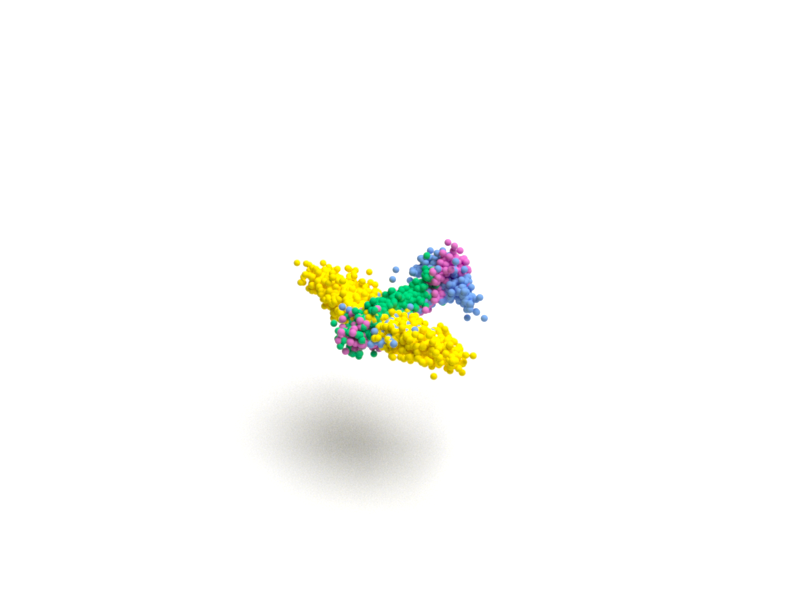} \\		
	\end{tabular}
	\caption{Qualitative examples of reconstructions on the ShapeNet \cite{chang2015shapenet} test set when training the autoencoding model with random rotations. The first two rows show the reconstructions obtained when applying random rotations to the same input point cloud. The third and fourth row show different test shapes and their reconstructions.} %While learning the distribution of randomly rotated point clouds is challenging, we observe that the mixture components learn to specialize in a semantically meaningful manner that is rotation invariant and generalizes across different shapes.}
	\vspace{-4mm}
	\label{fig:rotation_invariant_latents}
\end{figure}
\noindent{\textbf{Experimental setup.}} We train a mixture of $4$ \acp{nf} following the setup used for autoencoding on the airplane category of ShapeNet \cite{chang2015shapenet}. We also augment the training data with random 3D rotations. We qualitatively examine whether the mixture assignments are rotation invariant. If mixtures of \acp{nf} are able to specialize in a semantically meaningful way, we expect approximate rotational invariance.

\noindent{\textbf{Results.}} In \cref{fig:rotation_invariant_latents} we illustrate qualitative results for this experiment. Once the model is trained, we randomly rotate the input shapes before passing them through the model. We observe that different components of the mixture model learn to specialize in reconstructing different semantic parts of the airplanes (\eg yellow $\rightarrow$ wings, green $\rightarrow$ center of the airplane). However, learning the distribution of randomly rotated 3D point clouds is a much harder task and this is reflected in slightly less detailed models reconstructed by our method, as can be seen in  \cref{fig:rotation_invariant_latents}.

\subsection{Decreasing Number of Parameters}\label{section:small_models}
Based on prior work on \acp{nf} \cite{dinh2019rad, cornish2020relaxing} and our reasoning in \cref{seciotn:method}, mixtures of \acp{nf} yield increasing benefits in the regime of smaller decoder sizes. This experiment aims at verifying this intuition by comparing the reconstruction performance of \ac{dpf} \cite{klokov2020discrete} with mixtures of \acp{nf} for decoders with decreasing number of parameters. We are particularly interested in the reconstruction performance since it directly measures the representational strength of the underlying model. We refer to the supplement for generation metrics associated with this experiment.

\noindent{\textbf{Experimental setup.}} We train \ac{dpf} and mixtures of \acp{nf} using our generative modeling setup on the airplane category (see \cref{section:experiments}) varying the size of the decoder. For \ac{dpf} we use a decoder \ac{nf} with 63 (original size), 24, 12 and 6 coupling layers and unchanged $H=64$. We compare the reconstruction performance regarding the F1-score against a mixture of four \acp{nf}. $H$ and $N$ of the mixture are chosen such that it contains slightly fewer parameters. The detailed choice of $H$ and $N$ can be found in the supplement.

\noindent{\textbf{Results.}} In \cref{fig:small_decoder_and_runtime} (a) we observe that the relative improvement in terms of reconstruction performance increases for smaller decoder. While the mixture of four \acp{nf} achieves a relative improvement over \ac{dpf} \cite{klokov2020discrete} of 3.11\% in the original size, decreasing the number of coupling layers to 6 more than doubles the relative improvement up to 7.65\%.

% \subsection{Resampling of Sparse Regions}
% FYI look at Figure 4 in this paper https://arxiv.org/pdf/2008.06520.pdf.

% \begin{itemize}
%     \item later...
% \end{itemize}

\subsection{Inference Runtime Comparison}
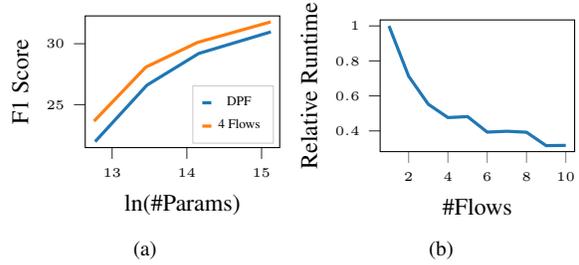
\begin{figure}
	\centering	
	\setlength{\arrayrulewidth}{.5pt}%
	\setlength{\tabcolsep}{1pt}
	\begin{tabular}{cc}
		% gt
% 		\includegraphics[width=0.45\linewidth]{figures/tiny_chair_f1.pdf} & 
% 		\includegraphics[width=0.45\linewidth]{figures/inference_runtime.pdf} \\
        \input{figures/tiny_chair_f1} & 
		\input{figures/inference_runtime} \\
		\footnotesize (a) & 
		\footnotesize (b) \\		
	\end{tabular}
	\vspace{0.5em}
	\caption{(a) Average reconstruction performance (F1-score) on the chair category of ShapeNet for \ac{dpf} (blue) and a mixture of 4 \acp{nf} (orange) vs. the parameter count. The relative improvement increases for small decoder. (b) Relative inference runtime of mixtures of \acp{nf} compared with \ac{dpf} at similar parameter count vs. the mixture size (\#Flows). More components reduce the inference runtime since each point is sampled from a smaller network.}
	\label{fig:small_decoder_and_runtime}
\end{figure}

Finally, we plot the inference runtime of mixtures of \acp{nf} against the number of components at constant parameter count. Specifically, we measure the average time per generated point during sampling. \cref{fig:small_decoder_and_runtime} (b) shows the relative inference runtime improvement depending on the size of the mixture. Since sampling of each point requires only a smaller \ac{nf} the average runtime decreases with an increasing number of mixture components. However, since the runtime of a \ac{nf} mainly depends on the number of coupling layers, we do not observe a linear behaviour.

% \begin{figure}
%     \centering
%     \vspace{0pt}
%     \includegraphics[width=0.45\linewidth]{figures/inference_runtime.pdf}
%     \label{fig:OODreconstruction}
%     \vspace{-2mm}
%     \label{fig:OODmnist_reconstruction}
%     \caption{Inference runtime comparison.}
% \end{figure}

% \begin{figure}
%     \centering
%     \vspace{0pt}
%     \includegraphics[width=0.45\linewidth]{figures/tiny_chair_f1.pdf}
%     \label{fig:OODreconstruction}
%     \vspace{-2mm}
%     \caption{F1 score in reconstruction depending on the number of parameters.}
%     \label{fig:OODmnist_reconstruction}
% \end{figure}

%% file: figures/tiny_chair_f1.tex
% This file was created by tikzplotlib v0.9.8.
\begin{tikzpicture}

\definecolor{color0}{rgb}{0.12156862745098,0.466666666666667,0.705882352941177}
\definecolor{color1}{rgb}{1,0.498039215686275,0.0549019607843137}

\begin{axis}[
legend style={
  fill opacity=1,
  draw opacity=1,
  text opacity=1,
  at={(0.97, 0.03)},
  anchor=south east,
  draw=white!80!black
},
tick align=outside,
tick pos=left,
xlabel={ln(\#Params)},
xmajorticks=true,
width=0.5\linewidth,
height=0.4\linewidth,
xmin=12.6402205822513, xmax=15.2425533015693,
xtick style={color=white!15!black},
y grid style={white!80!black},
ylabel={F1 Score},
ymajorticks=true,
ymin=21.479, ymax=32.281,
ytick style={color=white!15!black},
tick label style = {font=\tiny\tiny},
label style = {font=\small},
legend style={font=\tiny\tiny},
legend image post style={scale=0.2},
]
\addplot [very thick, color0]
table {%
12.7728910039961 21.97
13.466038184556 26.6
14.159185365116 29.2
15.1242654506912 30.97
};
\addlegendentry{DPF}
\addplot [very thick, color1]
table {%
12.7585084331294 23.65
13.4516901458647 28.08
14.1448373264246 30.11
15.1218652061065 31.79
};
\addlegendentry{4 Flows}
\end{axis}

\end{tikzpicture}

%% file: figures/inference_runtime.tex
% This file was created by tikzplotlib v0.9.8.
\begin{tikzpicture}

\definecolor{color0}{rgb}{0.12156862745098,0.466666666666667,0.705882352941177}

\begin{axis}[
tick align=outside,
xlabel={\#Flows},
xmajorticks=true,
xmin=0.55, xmax=10.45,
tick pos=left,
width=0.5\linewidth,
height=0.4\linewidth,
ylabel={Relative Runtime},
ymajorticks=true,
ymin=0.281601511067292, ymax=1.03420945185394,
ytick style={color=white!15!black},
tick label style = {font=\tiny\tiny},
label style = {font=\small},
]
\addplot [very thick, color0]
table {%
1 1
2 0.71260430298589
3 0.552463790928486
4 0.476115158206495
5 0.48166378369918
6 0.393081643698338
7 0.397981590377759
8 0.392442580217469
9 0.315810962921231
10 0.31719067408911
};
\end{axis}

\end{tikzpicture}

%% file: sections/05_discussion.tex
We proposed mixtures of \acp{nf} for modeling 3D point clouds which outperform models based on a single \ac{nf} \cite{yang2019pointflow, klokov2020discrete} on generation, autoencoding and \ac{svr} (see \cref{subsec:exp_autoencoding} \& \cref{subsec:sv_rec}). This resonates with the theoretical insight that single-flow-based models struggle on complicated geometries \cite{dinh2019rad, cornish2020relaxing}. We showed that mixtures of \acp{nf} can bypass these shortcomings by learning to compose a shape as a product of experts. While our consistent improvements are smaller in the overparameterized regime, we demonstrated that the relative gain over single-flow-based models increases for smaller models. This indicates that mixtures of \acp{nf} indeed denote a useful inductive bias for point clouds.

Furthermore, we observe that mixtures of \acp{nf} exhibit other interesting properties. The specialization of mixture components generalizes across different shapes (\eg the same flow is always responsible for the wings in \cref{fig:rotation_invariant_latents}) and can be made rotational invariant by adding random rotations at training time. This implies that mixtures of \acp{nf} gain a deeper understanding of the underlying shape. However, currently these invariant clusters lead to worse quality of the reconstructed point clouds which denotes a promising future research direction. Moreover, interpolating subregions individually leads to unrealistic shapes (see supplement). In future research we plan to explore ways to allow realistic interpolation of subregions.

%This property potentially gives rise to interesting applications in point cloud registration - even between different objects of the same class as a semantically meaningful clustering approach - and is a promising direction for future research.

%% file: sections_supplment/supplement.tex
Subsequently, we present further experimental results and details regarding mixtures of \acp{nf} for point clouds. Therefore, \cref{sec:ablation_study} investigates the impact of varying the number of components in a mixture of \acp{nf}.  In \cref{section:training_details} we give a detailed description of the training, optimization and architecture used in our experiments. Further, \cref{section:generation_small_models} demonstrate additional results on generative modeling. In \cref{section:interpolation} we show qualitative results of interpolating between latent representations of shapes, followed by the exploration of dense sampling with sparse input (see in \cref{section:sampling}. \cref{section:toy_example} presents an analytic toy example regarding the benefits of applying a mixture of \acp{nf} to point clouds as opposed to single-flow-based models. Lastly, we show more qualitative examples regarding generation, autoencoding and SVR in \cref{section:additional_qualitative_generation}, \cref{section:additional_qualitative_autoencoding} and \cref{section:additional_qualitative_SVR} respectively.

\section{Ablation Study}\label{section:ablation}

\input{sections_supplment/ablation_study}

\section{Training Details}\label{section:training_details}

\input{sections_supplment/training_details}

\section{Specialization of Individual NFs}

\input{sections_supplment/specialization_individual_flows}

\section{Toy Example on the Advantages of Mixtures of Normalizing Flows}\label{section:toy_example}

\input{sections_supplment/toy_example}

\section{Further Results on Generative Modelling}\label{section:generation_small_models}

\input{sections_supplment/generative_modelling}

\section{Interpolating Latent Representations}\label{section:interpolation}

\input{sections_supplment/latent_space_interpolation}

\section{Qualitative Results with Sparse Input}\label{section:sampling}

\input{sections_supplment/sampling}

\section{Additional Qualitative Results on Generation}\label{section:additional_qualitative_generation}

\input{sections_supplment/qualitative_results_on_generation}

\section{Additional Qualitative Results on Autoencoding}\label{section:additional_qualitative_autoencoding}

\input{sections_supplment/qualitative_results_on_ae}

\section{Additional Qualitative Results on SVR}\label{section:additional_qualitative_SVR}

\input{sections_supplment/qualitative_results_on_svr}

%% file: sections_supplment/ablation_study.tex
\label{sec:ablation_study}
We perform an ablation study regarding the number components $n$ in a mixture of \acp{nf}. Therefore, we train mixtures of \acp{nf} using a varying number components $n$ on the categories airplane ($n\in \left[ 1, 2, 4, 6, 8, 10 \right]$), chair ($n\in \left[ 1, 4, 8 \right]$), car ($n\in \left[ 1, 4, 8 \right]$) and report reconstruction performance in terms of the CD, EMD and F1-score (see \cref{table:ablation_autoencoding}). We observe that any number $>1$ leads to a clear improvement over a single-flow-based model. In our main experiments we choose $n=4$ as it performs well across all categories. However, we also note that there appears to be no strong preference regarding the number of components. We interpret this as evidence that geometries on ShapeNet are not complex enough to benefit from a very large $n$.

\begin{table}[h]
\begin{center}
 \begin{tabular}{| c | c | c | c | c |} 
 \hline
 Method & \#flows & CD $\downarrow$ & EMD $\downarrow$ & F1 $\uparrow$, $\tau=10^{-4}$\\ [0.5ex] 
 \hline\hline
 &1 & 2.90 & 3.53 & 60.68\\ 
 
 &2 & 2.89 & 3.52 & 61.23\\

 &4 & \textbf{2.88} & 3.50 & 61.07\\
 
 Airplane & 6 & 2.89 & 3.51 & 61.14\\
 
 &8 & 3.05 & \textbf{3.49} & \textbf{61.24}\\
 
 &10 & 2.90 & 3.50 & 61.08\\ [1ex] 
 
 \hline\hline
 &1 & 6.66 & 4.61 & 30.97\\

 Chair & 4 & \textbf{6.45} & \textbf{4.53} & \textbf{31.94}\\
 
 &8 & 6.60 & 4.54 & 31.77 \\ [1ex] 
 
 \hline\hline
 &1 & \textbf{7.41} & 4.44 & 21.20\\

 Car & 4 & 7.73 & \textbf{4.38} & \textbf{22.34}\\
 
 &8 & 7.75 & 4.39 & 21.22\\ [1ex]
 
 \hline
\end{tabular}
\caption{Auto-encoding. Ablation study on different number of \acp{nf}. CD is multiplied by $10^4$, EMD is multiplied by $10^2$.}
\label{table:ablation_autoencoding}
\end{center}
\end{table}

% % NEW RESULTS
% \begin{table}[h]
% \begin{center}
%  \begin{tabular}{| c | c | c | c | c |} 
%  \hline
%  Method & Nr. of flows & CD & EMD & F1, $\tau=10^{-4}$\\ [0.5ex] 
%  \hline\hline
%  &1 & \textbf{2.92} & 3.55 & 60.65\\ 
 
%  &2 & 2.90 & 3.56 & 60.96\\

%  &4 & 3.03 & \textbf{3.52} & \textbf{61.08}\\
 
%  Airplane & 6 & 3.02 & \textbf{3.52} & 61.06\\
 
%  &8 & 3.11 & 3.54 & 61.01\\
 
%  &10 &  &  & \\ [1ex] 
 
%  \hline\hline
%  &1 & 6.66 & 4.61 & 30.97\\

%  Chair & 4 & \textbf{6.44} & \textbf{4.57} & \textbf{31.88}\\
 
%  &8 & 6.60 & 4.65 & 31.56\\ [1ex] 
 
%  \hline\hline
%  &1 & \textbf{7.41} & 4.44 & 21.2\\

%  Car & 4 & 8.19 & 4.42 & \textbf{21.99}\\
 
%  &8 & 7.66 & \textbf{4.36} & 21.98\\ [1ex]
 
%  \hline
% \end{tabular}
% \caption{NEW RESUTLS: Auto-encoding. Ablation study on different number of \acp{nf}. CD is multiplied by $10^4$, EMD is multiplied by $10^2$.}
% \label{table:ablation_autoencoding}
% \end{center}
% \end{table}

%% file: sections_supplment/training_details.tex
\paragraph{Architecture.} We follow \cite{klokov2020discrete} and implement the \textit{encoder} in all our experiments as a PointNet \cite{qi2017pointnet}. Our PointNet encoder consists of 5 layers with feature sizes of the layers set to 3, 64, 128, 256, 512. Subsequently, we perform max-pooling along the dimension of the points. The resulting representation is fed through a \ac{mlp} comprised of two fully-connected layers. The first has a dimensionality of 512, the second has a dimensionality of $L=128$ (generative modeling) or $L=512$ (autoencoding and \ac{svr}). 

All our \acp{nf} use coupling layers as their fundamental building blocks which translate and scale alternatingly odd/even dimensions where translation/scaling factors are computed as functions of even/odd dimensions. Scaling and translation factors are computed using two separate models applied to the masked input: linear layer (input dimension: d; output dimension: D), 1-d Batchnorm, Swish activation function and a final linear layer (input dimension: D; output dimension: d). In our conditional coupling layers the input is initially transformed by two separate models of the form: linear layer (input dimension: k; output dimension: K), 1-d Batchnorm, Swish activation function and a final linear layer (input dimension: K; output dimension: K). We then apply FiLM conditioning \cite{perez2018film} where the condition is computed by a model of the form: linear layer (input dimension: L; output dimension: K), 1-d Batchnorm, Swish activation function and a final linear layer (input dimension: K; output dimension: K). Subsequently, scaling and translation are each fed through a ReLU activation function followed by a linear layer (input dimension: K; output dimension: k).

Our learned \textit{prior} is implemented as a normalizing flow consisting of 14 coupling layers. In these coupling layers we set $K=128$ and $k=D$ (D: dimensionality of the latent space/bottleneck) since it needs to be of the same dimensionality as the latent space for reasons of invertibility (generative modeling: $d=128$, autoencoding/\ac{svr} $d=512$). When training on \ac{svr} we implement the prior model as a conditional \ac{nf}. The condition is computed using an image encoder which is implemented as a ResNet18 \cite{he2016deep}. 

\paragraph{Setting K and number of coupling in mixtures of \acp{nf}.} In order to ensure comparability between models based on a single \ac{nf} and mixtures of \acp{nf}, we reduce the size of each \ac{nf} in mixtures of \acp{nf} such it has slightly less parameters than a given reference model using a single \ac{nf}. In particular, assume a single-flow-based model comprised of $N$ coupling layers using a hidden dimensionality K. For a mixture of $m$ \acp{nf} we compute the number of coupling layers $\hat{N}$ of each component in the mixture as $\hat{N} =\lceil \frac{N}{\sqrt{m}} \rceil$. Subsequently, we determine the hidden dimensionality $\hat{K}$ of each component of the mixture by reducing K until the total number of parameters of the mixture model is smaller than the one of the single-flow-based model.

\paragraph{Optimization.} We train all our models on Nvidia Titan RTX using ADAM \cite{kingma2015adam} for 1450 (generation), 1050 (autoencoding) or 36 (\ac{svr}) epochs using a batch size of 36. We start each training with a learning rate of $2.56 \cdot 10^{-4}$ and divide it by four at certain epochs (generation: 800, 1200, 1400; autoencoding: 400, 800, 1000; \ac{svr}: 20, 30, 35).

\paragraph{Dataset.} In order to provide a fair comparison with prior work, we conduct all the experiments using the ShapeNet dataset \cite{chang2015shapenet} provided by \cite{klokov2020discrete}. In our autoencoding experiments we use the ShapeNetCore.v2, which contains $55$k point clouds subdivided into $55$ classes. As for point cloud generation, we follow \cite{yang2019pointflow} and focus on three categories of the ShapeNet \cite{chang2015shapenet} dataset: \textit{airplanes}, \textit{cars}, and \textit{chairs}. Finally, for single-view reconstruction we adopt the dataset from \cite{choy20163d}, which contains renders of shapes from the $13$ classes of ShapeNetCore.v1. For each shape 24 images at a resolution of $137 \times 137$ are rendered from random viewpoints. The ground truth point clouds are obtained sampling from the original meshes. We randomly split per class in dataset into 70/10/20 proportion distributing to train/validation/test set for generation and autoencoding, for single-view reconstruction, we use the same train/test split from \cite{choy20163d}. All experiments regarding generation and autoencoding are conducted on the normalized dataset provided by \cite{klokov2020discrete}. Similarly, for single-view reconstruction the models are trained on normalized data. However, we scale the data into a unit radius sphere during evaluation to ensure comparability with related work.

\paragraph{Visualization of F1-score} We highlight the advantage of the F1-score as a convincing metric for perceptual quality by visualizing heatmaps of its components,  precision and recall. \cref{fig:f1-score} provides two examples from which we can see that F1-score clearly focuses on high-frequency regions of the objects. We observe that for \ac{dpf} the contributions to the F1-score primarily come from the high-frequency regions of an object, while in the case of mixtures of \acp{nf} the contributions are more evenly spread over the object.

\def\gefiga#1#2{{\includegraphics[clip,trim=8.5cm 4cm 8cm 7.0cm,width=0.2\linewidth]{figures/f1-score/#1/#2.png}}}
\begin{figure*}
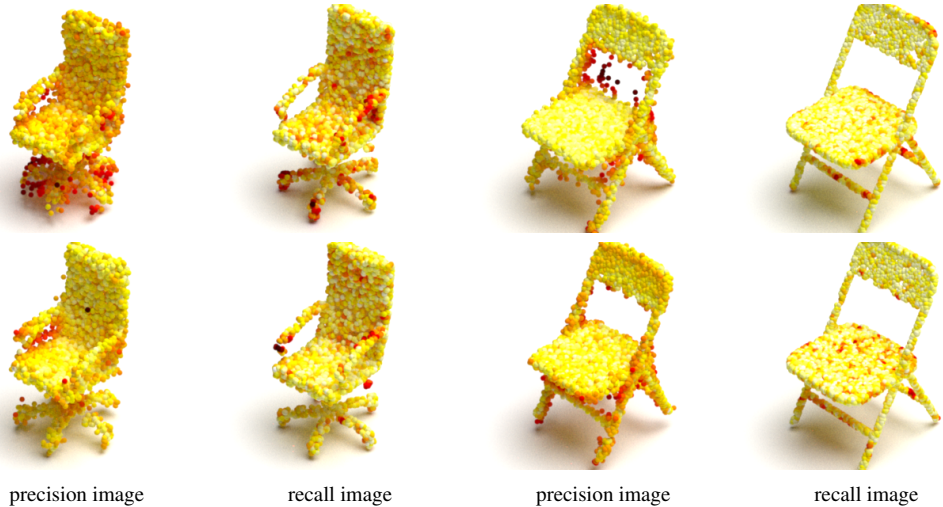

	\centering	
	\setlength{\arrayrulewidth}{.2pt}%
	\setlength{\tabcolsep}{0.1pt}
	\renewcommand{\arraystretch}{1}
	\begin{tabular}{ccccccc}
        \gefiga{dpf}{0_pr} &
	    \gefiga{dpf}{0_rec} &
		\gefiga{dpf}{1_pr} &
		\gefiga{dpf}{1_rec} \\
		\gefiga{ours}{0_pr} &
		\gefiga{ours}{0_rec} &
		\gefiga{ours}{1_pr} &
		\gefiga{ours}{1_rec} \\
		\footnotesize precision image & 
		\footnotesize recall image &
		\footnotesize precision image &
		\footnotesize recall image \\
	\end{tabular}
	\caption{Visualization on precision and recall of the F1-score (TOP: \ac{dpf}, BOTTOM: mixtures of 4 \acp{nf}). Points are considered to be more precise when the color is lighter in precision heatmap, meanwhile, points in the recall heatmap are lighter when the ground truth is well reconstructed by the model. }
	\label{fig:f1-score}
\end{figure*}

%% file: sections_supplment/specialization_individual_flows.tex
This is a consequence of Jensen's inequality. For $w_i \geq 0$ with $\sum_i w_i = 1$ and $p_i \geq 0$, we know that  $\log(\sum_i w_i p_i) \geq \sum_i w_i \log(p_i) \implies -\log(\sum_i w_i p_i) \leq -\sum_i w_i \log(p_i)$, equal for $p_i = p_j \forall i,j$. Thus a solution using unequal probabilities $p_i$ is preferred over an equal counterpart.

%% file: sections_supplment/toy_example.tex
% \begin{figure}[h!]
% 	\centering	
% 	\setlength{\arrayrulewidth}{.5pt}%
% 	\setlength{\tabcolsep}{1pt}
% 	\renewcommand{\arraystretch}{0.5}
% 	\begin{tabular}{cc}
% 		% gt
% 		\includegraphics[width=0.45\linewidth]{figures/toy_example/px.png} & 
% 		\includegraphics[width=0.45\linewidth]{figures/toy_example/py.png} \\	
% 	\end{tabular}
% 	\vspace{0.5em}
% 	\caption{Probability distribution with uniform probability density in a given interval and zero probability density otherwise.}
% 	\label{fig:toy_example}
% \end{figure}
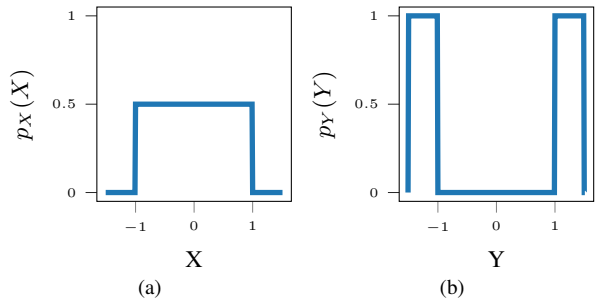
\begin{figure}
	\centering	
	\setlength{\arrayrulewidth}{.5pt}%
	\setlength{\tabcolsep}{1pt}
	\renewcommand{\arraystretch}{0.5}
	\begin{tabular}{cc}
		% gt
% 		\includegraphics[width=0.45\linewidth]{figures/tiny_chair_f1.pdf} & 
% 		\includegraphics[width=0.45\linewidth]{figures/inference_runtime.pdf} \\
        \input{figures/toy_example/px} & 
		\input{figures/toy_example/py} \\
		\footnotesize (a) & 
		\footnotesize (b) \\		
	\end{tabular}
	\vspace{0.5em}
	\caption{Probability distribution with uniform probability density in a given interval and zero probability density otherwise.}
	\label{fig:toy_example}
\end{figure}

This section presents an analytic toy example demonstrating the advantages of using mixtures of \acp{nf}. Therefore, consider the one-dimensional distributions in \cref{fig:toy_example}:
\begin{align} \label{eq:basic_nf_toy_example}
    p_{Y}(Y) &=
    \begin{cases}
      0.5 & \text{if $x<1$ and $x\geq-1$}\\
      0 & \text{else}
    \end{cases} \\
    p_{X}(X) &=
    \begin{cases}
      1 & \text{if ($x\geq1$ and $x<2$)}\\
      & \text{or ($x\geq-2$ and $x<-1$)}\\
      0 & \text{else}
    \end{cases}
\end{align}
We wish to find an invertible transformation $f: Y\rightarrow X$ such that the change of variable formula 
\begin{equation}
    p_{Y}(Y) = p_{X}(f(Y))\frac{df(Y)}{dY}
\end{equation}
is satisfied. In this simple example we can directly write down the solution, namely
\begin{equation}
    f(Y) = 
    \begin{cases}
      2\cdot (Y - 1) & \text{if $x\leq1$}\\
      2\cdot (Y + 1) & \text{if $x>1$}
    \end{cases}. \\
\end{equation}
Interestingly this function contains a discontinuity at 0. This discontinuity also implies an infinite bi-Lipschitz constant of the optimal solution as can be seen from the definition of the bi-Lipschitz constant K of a function g:
\begin{equation}
    \frac{1}{K} \left| x_2 - x_1 \right| \leq \left| g(x_2) - g(x_1) \right| \leq K \left| x_2 - x_1 \right| \forall x_1, x_2
\end{equation}
In the vicinity of the origin $K$ has to approach infinity in order to fulfill above inequality. Attempting to learn such a discontinuous function using a neural network, which is only a universal function approximator for continuous functions, is difficult. However, we can bypass the discontinuity in this solution by utilizing two invertible maps, $f_1 = 4\cdot (Y - 1.5)$ and $f_2 = 4\cdot (Y + 1.5)$, and composing them as a mixture. Thus, we are introducing an additional continuous random variable $w$ that identifies the invertible map responsible for a particular point $y\in Y$. This describes the underlying idea of applying mixtures of \acp{nf} to 3D data. By introducing additional latent variables we can empower our continuous model to avoid appoximating discontinuous behaviour.

%% file: figures/toy_example/px.tex
% This file was created by tikzplotlib v0.9.8.
\begin{tikzpicture}

\definecolor{color0}{rgb}{0.12156862745098,0.466666666666667,0.705882352941177}

\begin{axis}[
tick align=outside,
xlabel={X},
xmajorticks=true,
xmin=-1.661, xmax=1.661,
tick pos=left,
width=0.5\linewidth,
height=0.5\linewidth,
ylabel={$p_X (X)$},
ymajorticks=true,
ymin=-0.05, ymax=1.05,
ytick style={color=white!15!black},
tick label style = {font=\tiny\tiny},
label style = {font=\small},
]
\addplot [line width=2pt, color0]
table {%
-1.51 0
-1.5 0
-1.49 0
-1.48 0
-1.47 0
-1.46 0
-1.45 0
-1.44 0
-1.43 0
-1.42 0
-1.41 0
-1.4 0
-1.39 0
-1.38 0
-1.37 0
-1.36 0
-1.35 0
-1.34 0
-1.33 0
-1.32 0
-1.31 0
-1.3 0
-1.29 0
-1.28 0
-1.27 0
-1.26 0
-1.25 0
-1.24 0
-1.23 0
-1.22 0
-1.21 0
-1.2 0
-1.19 0
-1.18 0
-1.17 0
-1.16 0
-1.15 0
-1.14 0
-1.13 0
-1.12 0
-1.11 0
-1.1 0
-1.09 0
-1.08 0
-1.07 0
-1.06 0
-1.05 0
-1.04 0
-1.03 0
-1.02 0
-1.01 0
-1 0.5
-0.99 0.5
-0.98 0.5
-0.97 0.5
-0.96 0.5
-0.95 0.5
-0.94 0.5
-0.929999999999999 0.5
-0.919999999999999 0.5
-0.909999999999999 0.5
-0.899999999999999 0.5
-0.889999999999999 0.5
-0.879999999999999 0.5
-0.869999999999999 0.5
-0.859999999999999 0.5
-0.849999999999999 0.5
-0.839999999999999 0.5
-0.829999999999999 0.5
-0.819999999999999 0.5
-0.809999999999999 0.5
-0.799999999999999 0.5
-0.789999999999999 0.5
-0.779999999999999 0.5
-0.769999999999999 0.5
-0.759999999999999 0.5
-0.749999999999999 0.5
-0.739999999999999 0.5
-0.729999999999999 0.5
-0.719999999999999 0.5
-0.709999999999999 0.5
-0.699999999999999 0.5
-0.689999999999999 0.5
-0.679999999999999 0.5
-0.669999999999999 0.5
-0.659999999999999 0.5
-0.649999999999999 0.5
-0.639999999999999 0.5
-0.629999999999999 0.5
-0.619999999999999 0.5
-0.609999999999999 0.5
-0.599999999999999 0.5
-0.589999999999999 0.5
-0.579999999999999 0.5
-0.569999999999999 0.5
-0.559999999999999 0.5
-0.549999999999999 0.5
-0.539999999999999 0.5
-0.529999999999999 0.5
-0.519999999999999 0.5
-0.509999999999999 0.5
-0.499999999999999 0.5
-0.489999999999999 0.5
-0.479999999999999 0.5
-0.469999999999999 0.5
-0.459999999999999 0.5
-0.449999999999999 0.5
-0.439999999999999 0.5
-0.429999999999999 0.5
-0.419999999999999 0.5
-0.409999999999999 0.5
-0.399999999999999 0.5
-0.389999999999999 0.5
-0.379999999999999 0.5
-0.369999999999999 0.5
-0.359999999999999 0.5
-0.349999999999999 0.5
-0.339999999999999 0.5
-0.329999999999999 0.5
-0.319999999999999 0.5
-0.309999999999999 0.5
-0.299999999999999 0.5
-0.289999999999999 0.5
-0.279999999999999 0.5
-0.269999999999999 0.5
-0.259999999999999 0.5
-0.249999999999999 0.5
-0.239999999999999 0.5
-0.229999999999999 0.5
-0.219999999999999 0.5
-0.209999999999999 0.5
-0.199999999999999 0.5
-0.189999999999999 0.5
-0.179999999999999 0.5
-0.169999999999999 0.5
-0.159999999999999 0.5
-0.149999999999999 0.5
-0.139999999999999 0.5
-0.129999999999999 0.5
-0.119999999999999 0.5
-0.109999999999999 0.5
-0.0999999999999988 0.5
-0.0899999999999987 0.5
-0.0799999999999987 0.5
-0.0699999999999987 0.5
-0.0599999999999987 0.5
-0.0499999999999987 0.5
-0.0399999999999987 0.5
-0.0299999999999987 0.5
-0.0199999999999987 0.5
-0.00999999999999868 0.5
1.33226762955019e-15 0.5
0.0100000000000013 0.5
0.0200000000000014 0.5
0.0300000000000014 0.5
0.0400000000000014 0.5
0.0500000000000014 0.5
0.0600000000000014 0.5
0.0700000000000014 0.5
0.0800000000000014 0.5
0.0900000000000014 0.5
0.100000000000001 0.5
0.110000000000001 0.5
0.120000000000001 0.5
0.130000000000001 0.5
0.140000000000001 0.5
0.150000000000001 0.5
0.160000000000001 0.5
0.170000000000001 0.5
0.180000000000001 0.5
0.190000000000002 0.5
0.200000000000002 0.5
0.210000000000002 0.5
0.220000000000002 0.5
0.230000000000002 0.5
0.240000000000002 0.5
0.250000000000002 0.5
0.260000000000002 0.5
0.270000000000002 0.5
0.280000000000002 0.5
0.290000000000002 0.5
0.300000000000002 0.5
0.310000000000002 0.5
0.320000000000002 0.5
0.330000000000002 0.5
0.340000000000002 0.5
0.350000000000002 0.5
0.360000000000002 0.5
0.370000000000002 0.5
0.380000000000002 0.5
0.390000000000002 0.5
0.400000000000002 0.5
0.410000000000002 0.5
0.420000000000002 0.5
0.430000000000002 0.5
0.440000000000002 0.5
0.450000000000002 0.5
0.460000000000002 0.5
0.470000000000002 0.5
0.480000000000002 0.5
0.490000000000002 0.5
0.500000000000002 0.5
0.510000000000002 0.5
0.520000000000002 0.5
0.530000000000002 0.5
0.540000000000002 0.5
0.550000000000002 0.5
0.560000000000002 0.5
0.570000000000002 0.5
0.580000000000002 0.5
0.590000000000002 0.5
0.600000000000002 0.5
0.610000000000002 0.5
0.620000000000002 0.5
0.630000000000002 0.5
0.640000000000002 0.5
0.650000000000002 0.5
0.660000000000002 0.5
0.670000000000002 0.5
0.680000000000002 0.5
0.690000000000002 0.5
0.700000000000002 0.5
0.710000000000002 0.5
0.720000000000002 0.5
0.730000000000002 0.5
0.740000000000002 0.5
0.750000000000002 0.5
0.760000000000002 0.5
0.770000000000002 0.5
0.780000000000002 0.5
0.790000000000002 0.5
0.800000000000002 0.5
0.810000000000002 0.5
0.820000000000002 0.5
0.830000000000002 0.5
0.840000000000002 0.5
0.850000000000002 0.5
0.860000000000002 0.5
0.870000000000002 0.5
0.880000000000002 0.5
0.890000000000002 0.5
0.900000000000002 0.5
0.910000000000002 0.5
0.920000000000002 0.5
0.930000000000002 0.5
0.940000000000002 0.5
0.950000000000002 0.5
0.960000000000002 0.5
0.970000000000002 0.5
0.980000000000002 0.5
0.990000000000002 0.5
1 0
1.01 0
1.02 0
1.03 0
1.04 0
1.05 0
1.06 0
1.07 0
1.08 0
1.09 0
1.1 0
1.11 0
1.12 0
1.13 0
1.14 0
1.15 0
1.16 0
1.17 0
1.18 0
1.19 0
1.2 0
1.21 0
1.22 0
1.23 0
1.24 0
1.25 0
1.26 0
1.27 0
1.28 0
1.29 0
1.3 0
1.31 0
1.32 0
1.33 0
1.34 0
1.35 0
1.36 0
1.37 0
1.38 0
1.39 0
1.4 0
1.41 0
1.42 0
1.43 0
1.44 0
1.45 0
1.46 0
1.47 0
1.48 0
1.49 0
1.5 0
1.51 0
};
\end{axis}

\end{tikzpicture}

%% file: figures/toy_example/py.tex
% This file was created by tikzplotlib v0.9.8.
\begin{tikzpicture}

\definecolor{color0}{rgb}{0.12156862745098,0.466666666666667,0.705882352941177}

\begin{axis}[
tick align=outside,
xlabel={Y},
xmajorticks=true,
xmin=-1.661, xmax=1.661,
tick pos=left,
width=0.5\linewidth,
height=0.5\linewidth,
ylabel={$p_Y (Y)$},
ymajorticks=true,
ymin=-0.05, ymax=1.05,
ytick style={color=white!15!black},
tick label style = {font=\tiny\tiny},
label style = {font=\small},
]
\addplot [line width=2pt, color0]
table {%
-1.51 0
-1.5 1
-1.49 1
-1.48 1
-1.47 1
-1.46 1
-1.45 1
-1.44 1
-1.43 1
-1.42 1
-1.41 1
-1.4 1
-1.39 1
-1.38 1
-1.37 1
-1.36 1
-1.35 1
-1.34 1
-1.33 1
-1.32 1
-1.31 1
-1.3 1
-1.29 1
-1.28 1
-1.27 1
-1.26 1
-1.25 1
-1.24 1
-1.23 1
-1.22 1
-1.21 1
-1.2 1
-1.19 1
-1.18 1
-1.17 1
-1.16 1
-1.15 1
-1.14 1
-1.13 1
-1.12 1
-1.11 1
-1.1 1
-1.09 1
-1.08 1
-1.07 1
-1.06 1
-1.05 1
-1.04 1
-1.03 1
-1.02 1
-1.01 1
-1 0
-0.99 0
-0.98 0
-0.97 0
-0.96 0
-0.95 0
-0.94 0
-0.929999999999999 0
-0.919999999999999 0
-0.909999999999999 0
-0.899999999999999 0
-0.889999999999999 0
-0.879999999999999 0
-0.869999999999999 0
-0.859999999999999 0
-0.849999999999999 0
-0.839999999999999 0
-0.829999999999999 0
-0.819999999999999 0
-0.809999999999999 0
-0.799999999999999 0
-0.789999999999999 0
-0.779999999999999 0
-0.769999999999999 0
-0.759999999999999 0
-0.749999999999999 0
-0.739999999999999 0
-0.729999999999999 0
-0.719999999999999 0
-0.709999999999999 0
-0.699999999999999 0
-0.689999999999999 0
-0.679999999999999 0
-0.669999999999999 0
-0.659999999999999 0
-0.649999999999999 0
-0.639999999999999 0
-0.629999999999999 0
-0.619999999999999 0
-0.609999999999999 0
-0.599999999999999 0
-0.589999999999999 0
-0.579999999999999 0
-0.569999999999999 0
-0.559999999999999 0
-0.549999999999999 0
-0.539999999999999 0
-0.529999999999999 0
-0.519999999999999 0
-0.509999999999999 0
-0.499999999999999 0
-0.489999999999999 0
-0.479999999999999 0
-0.469999999999999 0
-0.459999999999999 0
-0.449999999999999 0
-0.439999999999999 0
-0.429999999999999 0
-0.419999999999999 0
-0.409999999999999 0
-0.399999999999999 0
-0.389999999999999 0
-0.379999999999999 0
-0.369999999999999 0
-0.359999999999999 0
-0.349999999999999 0
-0.339999999999999 0
-0.329999999999999 0
-0.319999999999999 0
-0.309999999999999 0
-0.299999999999999 0
-0.289999999999999 0
-0.279999999999999 0
-0.269999999999999 0
-0.259999999999999 0
-0.249999999999999 0
-0.239999999999999 0
-0.229999999999999 0
-0.219999999999999 0
-0.209999999999999 0
-0.199999999999999 0
-0.189999999999999 0
-0.179999999999999 0
-0.169999999999999 0
-0.159999999999999 0
-0.149999999999999 0
-0.139999999999999 0
-0.129999999999999 0
-0.119999999999999 0
-0.109999999999999 0
-0.0999999999999988 0
-0.0899999999999987 0
-0.0799999999999987 0
-0.0699999999999987 0
-0.0599999999999987 0
-0.0499999999999987 0
-0.0399999999999987 0
-0.0299999999999987 0
-0.0199999999999987 0
-0.00999999999999868 0
1.33226762955019e-15 0
0.0100000000000013 0
0.0200000000000014 0
0.0300000000000014 0
0.0400000000000014 0
0.0500000000000014 0
0.0600000000000014 0
0.0700000000000014 0
0.0800000000000014 0
0.0900000000000014 0
0.100000000000001 0
0.110000000000001 0
0.120000000000001 0
0.130000000000001 0
0.140000000000001 0
0.150000000000001 0
0.160000000000001 0
0.170000000000001 0
0.180000000000001 0
0.190000000000002 0
0.200000000000002 0
0.210000000000002 0
0.220000000000002 0
0.230000000000002 0
0.240000000000002 0
0.250000000000002 0
0.260000000000002 0
0.270000000000002 0
0.280000000000002 0
0.290000000000002 0
0.300000000000002 0
0.310000000000002 0
0.320000000000002 0
0.330000000000002 0
0.340000000000002 0
0.350000000000002 0
0.360000000000002 0
0.370000000000002 0
0.380000000000002 0
0.390000000000002 0
0.400000000000002 0
0.410000000000002 0
0.420000000000002 0
0.430000000000002 0
0.440000000000002 0
0.450000000000002 0
0.460000000000002 0
0.470000000000002 0
0.480000000000002 0
0.490000000000002 0
0.500000000000002 0
0.510000000000002 0
0.520000000000002 0
0.530000000000002 0
0.540000000000002 0
0.550000000000002 0
0.560000000000002 0
0.570000000000002 0
0.580000000000002 0
0.590000000000002 0
0.600000000000002 0
0.610000000000002 0
0.620000000000002 0
0.630000000000002 0
0.640000000000002 0
0.650000000000002 0
0.660000000000002 0
0.670000000000002 0
0.680000000000002 0
0.690000000000002 0
0.700000000000002 0
0.710000000000002 0
0.720000000000002 0
0.730000000000002 0
0.740000000000002 0
0.750000000000002 0
0.760000000000002 0
0.770000000000002 0
0.780000000000002 0
0.790000000000002 0
0.800000000000002 0
0.810000000000002 0
0.820000000000002 0
0.830000000000002 0
0.840000000000002 0
0.850000000000002 0
0.860000000000002 0
0.870000000000002 0
0.880000000000002 0
0.890000000000002 0
0.900000000000002 0
0.910000000000002 0
0.920000000000002 0
0.930000000000002 0
0.940000000000002 0
0.950000000000002 0
0.960000000000002 0
0.970000000000002 0
0.980000000000002 0
0.990000000000002 0
1 1
1.01 1
1.02 1
1.03 1
1.04 1
1.05 1
1.06 1
1.07 1
1.08 1
1.09 1
1.1 1
1.11 1
1.12 1
1.13 1
1.14 1
1.15 1
1.16 1
1.17 1
1.18 1
1.19 1
1.2 1
1.21 1
1.22 1
1.23 1
1.24 1
1.25 1
1.26 1
1.27 1
1.28 1
1.29 1
1.3 1
1.31 1
1.32 1
1.33 1
1.34 1
1.35 1
1.36 1
1.37 1
1.38 1
1.39 1
1.4 1
1.41 1
1.42 1
1.43 1
1.44 1
1.45 1
1.46 1
1.47 1
1.48 1
1.49 1
1.5 0
1.51 0
};
\end{axis}

\end{tikzpicture}

%% file: sections_supplment/generative_modelling.tex
\subsection*{Point Cloud Matches Using MMD-CD and MMd-EMD}

Fig. \ref{fig: failure_matches} depicts matches used by the MMD metric based on CD and EMD. We observe that matched point clouds are highly dissimilar.

%9.5cm 3.5cm 10cm 6cm
\def\gefig#1{{\includegraphics[clip,trim=8cm 4cm 8cm 6.5cm, width=0.25\linewidth]{figures/bad case in generation/#1.png}}}
\def\gefige#1{{\includegraphics[clip,trim=6cm 1cm 6cm 6cm,width=0.25\linewidth]{figures/bad case in generation/#1.png}}}
\def\gefigf#1{{\includegraphics[clip,trim=9cm 4cm 9cm 7cm,width=0.25\linewidth]{figures/bad case in generation/#1.png}}}
\def\gefigk#1{{\includegraphics[clip,trim=8cm 3cm 8cm 6.5cm,width=0.25\linewidth]{figures/bad case in generation/#1.png}}}
\def\gefigh#1{{\includegraphics[clip,trim=10cm 4cm 10cm 6.5cm,width=0.25\linewidth]{figures/bad case in generation/#1.png}}}
\begin{figure}
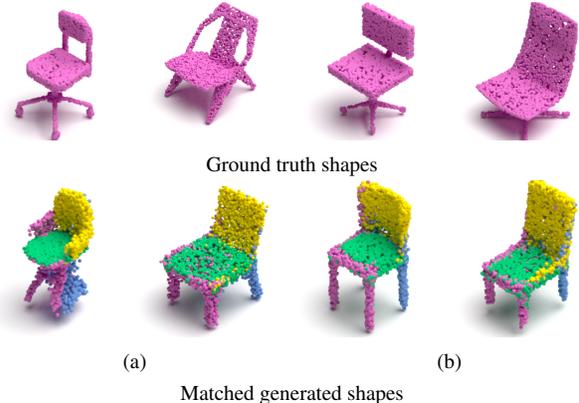

	\centering	
	\setlength{\arrayrulewidth}{.5pt}%
	\setlength{\tabcolsep}{0.1pt}
	\begin{tabular}{cccc}
		% second two cases
		\gefig{0_gt} &
		\gefige{1_gt} &
		\gefig{2_gt} &
		\gefig{3_gt} \\

		\multicolumn{4}{c}{\footnotesize Ground truth shapes} \\
		% first two cases
        \gefigf{0_fail} &
		\gefigk{1_fail} &
		\gefigf{2_fail} &
		\gefigf{3_fail} \\
		\multicolumn{2}{c}{\footnotesize(a)} & \multicolumn{2}{c}{\footnotesize(b)} \\
		& \multicolumn{2}{c}{\footnotesize Matched generated shapes} &  \\
	\end{tabular}
	\caption{Failure cases of matches for ground truth shapes when computing MMD based on CD (a) and EMD (b). For each ground truth shape within the test set depicted (top row), searching for the most similar generated shape according to the CD and EMD distance yields clearly different matched shapes (bottom row). This supports the claim that MMD-CD/EMD do not clearly reflect the perceptual quality of generated point clouds.}
	\label{fig:failure_matches}
\end{figure}

\subsection*{Quantitative Results on Generative including Standard Deviation}

We report quantitative metrics on generative modeling including their standard deviation in \cref{table:comparison_related_work_generation_supplement}. 

% With EMD
\begin{sidewaystable*}
% \begin{table*}[h]
\begin{center}
 \begin{tabular}{| c | c | c | c| c | c | c | c | c | c | c | c |} 
 \hline
 & & JSD $\downarrow$ & \multicolumn{3}{|c|}{MMD} & \multicolumn{3}{|c|}{COV $\uparrow$} & \multicolumn{3}{|c|}{1-NNA $\downarrow$} \\
 Cates & Method & & CD $\downarrow$ & EMD $\downarrow$ & F1 $\uparrow$ & CD & EMD & F1 & CD & EMD & F1\\ [0.5ex] 
 \hline\hline
 & \cite{achlioptas2018learning}-CD & 2.76 $\pm$ 0.2 & \textbf{5.69} $\pm$ 0.0 & 5.16 $\pm$ 0.0 & - & 39.5 $\pm$ 0.8 & 17.1 $\pm$ 0.6 & - & 72.9 $\pm$ 0.8 & 92.1 $\pm$ 0.6 & -\\ 
 
 & \cite{achlioptas2018learning}-EMD & 1.77 $\pm$ 0.1 & 6.05 $\pm$ 0.0 & \textbf{4.15} $\pm$ 0.0 & - & 39.7 $\pm$ 1.4 & 40.4 $\pm$ 1.2 & - & 75.7 $\pm$ 0.6 & 73.0 $\pm$ 1.2 & -\\

 Plane & \cite{yang2019pointflow} & 1.42 $\pm$ 0.1 & 6.05 $\pm$ 0.1 & 4.32 $\pm$ 0.0 & - & 44.7 $\pm$ 1.2 & \textbf{48.4} $\pm$ 1.0 & - & 70.9 $\pm$ 1.0 & 68.4 $\pm$ 1.0 & -\\
 
 & \cite{klokov2020discrete} & \underline{1.14} $\pm$ 0.1 & \underline{6.03} $\pm$ 0.1 & 4.27 $\pm$ 0.0 & \textbf{50.84} $\pm$ 0.5 & \underline{46.4} $\pm$ 1.3 & 48.2 $\pm$ 1.2 & \textbf{42.7} $\pm$ 1.0 & \underline{70.3} $\pm$ 1.2 & \underline{67.5} $\pm$ 1.1 & 72.7 $\pm$ 0.8\\
 
 & Ours& \textbf{1.03} $\pm$ 0.1 & 6.06 $\pm$ 0.1 & \underline{4.26} $\pm$ 0.0 & 50.11 $\pm$ 0.5 & \textbf{46.5} $\pm$ 0.85 & \textbf{48.4} $\pm$ 1.5 & 42.3 $\pm$ 0.9 & \textbf{70.1} $\pm$ 1.1 & \textbf{66.9} $\pm$ 1.5 & \textbf{71.7} $\pm$ 1.3\\ 
 \hline
 & Oracle & 0.50 $\pm$ 0.0 & 5.97 $\pm$ 0.1 & 3.98 $\pm$ 0.0 & 75.48 $\pm$ 0.3 & 51.4 $\pm$ 1.0 & 52.7 $\pm$ 1.3 & 94.3 $\pm$ 0.5 & 49.8 $\pm$ 1.3 & 48.2 $\pm$ 1.1  & 50.2 $\pm$ 1.0\\ [1ex] 
 
 \hline\hline
 & \cite{achlioptas2018learning}-CD & 3.65 $\pm$ 0.1 & \textbf{16.66} $\pm$ 0.1 & 7.91 $\pm$ 0.0 & - & 42.3 $\pm$ 0.5 & 17.1 $\pm$ 0.5 & - & 68.5 $\pm$ 0.5 & 96.5 $\pm$ 0.1 & -\\ 
 
 & \cite{achlioptas2018learning}-EMD & \textbf{1.27} $\pm$ 0.1 & \underline{16.78} $\pm$ 0.1 & \textbf{5.75} $\pm$ 0.0 & - & 44.3 $\pm$ 0.9 & 43.8 $\pm$ 1.0 & - & 66.6 $\pm$ 0.6 & \textbf{67.8} $\pm$ 0.7 & -\\

 Chair & \cite{yang2019pointflow} & 1.51 $\pm$ 0.1 & 17.15 $\pm$ 0.1 & 6.20 $\pm$ 0.0 & - &  43.3 $\pm$ 0.8 & \textbf{46.5} $\pm$ 1.0 & - & 67.0 $\pm$ 0.3 & 70.4 $\pm$ 0.6 & -\\
 
 & \cite{klokov2020discrete} & \underline{1.37} $\pm$ 0.1 & 17.24 $\pm$ 0.2 & 6.13 $\pm$ 0.0 &  19.63 $\pm$ 0.2 & \underline{45.1} $\pm$ 1.0 & 46.0 $\pm$ 0.7 & 34.7 $\pm$ 0.9 & \textbf{64.8} $\pm$ 0.7 & 68.2 $\pm$ 0.8 & 67.7 $\pm$ 0.8\\
 
 & Ours & 1.45 $\pm$ 0.1 & 17.30 $\pm$ 0.1 & \underline{6.11} $\pm$ 0.0 & \textbf{21.08} $\pm$ 0.3 & \textbf{45.2} $\pm$ 1.5 & \textbf{46.5} $\pm$ 0.6 & \textbf{39.2} $\pm$ 1.3 & \underline{65.3} $\pm$ 1.1 & \textbf{65.6} $\pm$ 0.8 & \textbf{62.2} $\pm$ 0.9\\ 
 \hline
 & Oracle & 0.49 $\pm$ 0.1 & 16.39 $\pm$ 0.1 & 5.71 $\pm$ 0.0 & 49.14 $\pm$ 0.2 & 52.8 $\pm$ 0.8 & 53.4 $\pm$ 1.1 & 99.8 $\pm$ 0.3 & 49.7 $\pm$ 0.7 & 49.6 $\pm$ 0.9 & 49.5 $\pm$ 0.7\\ [1ex] 
 
 \hline\hline
 & \cite{achlioptas2018learning}-CD & 2.65 $\pm$ 0.1 & \textbf{8.83} $\pm$ 0.06 & 5.36 $\pm$ 0.0 & - & 41.3 $\pm$ 0.8 & 15.9 $\pm$ 1.3 & - & \textbf{62.6} $\pm$ 0.6 & 92.7 $\pm$ 0.4 & -\\ 
 
 & \cite{achlioptas2018learning}-EMD & 1.31 $\pm$ 0.1 & \underline{9.00} $\pm$ 0.1 & \textbf{4.40} $\pm$ 0.0 & - & 38.3 $\pm$ 1.2 & 32.9 $\pm$ 0.7 & - & \underline{65.2} $\pm$ 0.4 & \textbf{63.2} $\pm$ 1.0 & -\\

 Car & \cite{yang2019pointflow} & 0.59 $\pm$ 0.0 & 9.53 $\pm$ 0.1 & 4.71 $\pm$ 0.0 & - & \textbf{42.3} $\pm$ 1.0 & 35.8 $\pm$ 1.3 & - & 70.1 $\pm$ 0.9 & 74.2 $\pm$ 0.6 & -\\
 
 & \cite{klokov2020discrete} & \underline{0.57} $\pm$ 0.0 & 9.67 $\pm$ 0.1 & \underline{4.60} $\pm$ 0.0 & \textbf{18.11} $\pm$ 0.1 & 40.8 $\pm$ 1.4 & \underline{43.7} $\pm$ 1.0 & 35.84 $\pm$ 1.7 & 71.3 $\pm$ 1.1 & 66.0 $\pm$ 1.5 & 66.7 $\pm$ 1.2 \\
 
 & Ours & \textbf{0.55} $\pm$ 0.0 & 9.50 $\pm$ 0.1 & 4.62 $\pm$ 0.01 &\textbf{18.20} $\pm$ 0.2 &   \underline{41.4} $\pm$ 1.2 & \textbf{43.8} $\pm$ 1.1 &  \textbf{37.83} $\pm$ 1.3 & 69.0 $\pm$ 0.9 & \underline{64.8} $\pm$ 1.2  & \textbf{66.1} $\pm$ 0.2\\ 
 \hline
 & Oracle & 0.37 $\pm$ 0.0 & 9.24 $\pm$ 0.1 & 4.56 $\pm$ 0.1 & 35.03 $\pm$ 0.1 & 52.8 $\pm$ 1.1 & 52.7 $\pm$ 0.9 & 99.5 $\pm$ 1.2 & 50.9 $\pm$ 1.1 & 50.5 $\pm$ 1.2 & 49.1 $\pm$ 0.4\\ [1ex] 
 
 \hline
\end{tabular}
\caption{Generative modeling. Comparison with related work. JSD and MMD-EMD, MMD-F1 ($\tau=10^{-4}$) are multiplied by $10^2$, MMD-CD is multiplied by $10^4$}
\label{table:comparison_related_work_generation_supplement}
\end{center}
% \end{table*}
\end{sidewaystable*}

\subsection*{Results on Generative Modelling with Reduced Model Size}

We report quantitative (see \cref{table:generative_modeling_small_models}) and qualitative (see \cref{fig:generative_modeling_qualitative_small_models}) results on generative modeling using decoder models of reduced parameter counts for a single-flow-based model and a mixture of 4 \acp{nf}. We reduce the parameter count by decreasing the number of coupling layers used by the normalizing flow (24, 12, 6). Note that the original model used in our main experiments contains 63 coupling layers. We observe that for 24 \& 12 coupling layer the quantitative metrics on generative modeling remain largely unchanged while the quality of the generated samples (see \cref{fig:generative_modeling_qualitative_small_models}) and reconstruction performance (see main paper) clearly degrade. We only see a clear quantitative degradation in generative modeling performance when limiting the model to 6 coupling layers. We conclude that commonly used metrics for generative modeling struggle to represent perceptual quality of generated point clouds.

% Without EMD
% \begin{table*}[h]
% \begin{center}
%  \begin{tabular}{| c | c | c | c | c | c |} 
%  \hline
%  && JSD $\downarrow$ & MMD $\downarrow$ & COV $\uparrow$ & 1-NNA $\downarrow$ \\
%  \#Coupling Layers & \#Flows & & CD & CD & CD\\ [0.5ex] 
%  \hline\hline
%  24 & 1 & 1.11 & 17.24 & 45.66 & 66.58\\ 
 
%  24 & 4 & 1.09 & 17.49 & 45.82 & 67.15\\ 

%  12 & 1 & 1.16 & 17.25 & 44.81 & 66.36\\ 
 
%  12 & 4 & 1.10 & 17.48 & 44.97 & 70.03\\[1ex]  
 
%  \hline
% \end{tabular}
% \caption{Quantitative evaluation of generative modeling using a smaller decoder (8 \& 4 coupling layers).}
% \label{table:comparison_on_smaller_size}
% \end{center}
% \end{table*}

% With EMD
\begin{table*}[h]
\begin{center}
 \begin{tabular}{| c | c | c | c | c | c | c | c | c |} 
 \hline
 && JSD $\downarrow$ & \multicolumn{2}{|c|}{MMD $\downarrow$} & \multicolumn{2}{|c|}{COV $\uparrow$} & \multicolumn{2}{|c|}{1-NNA $\downarrow$} \\
 Nr. coupling layers & Nr. of Flows & & CD & EMD & CD & EMD & CD & EMD \\ [0.5ex] 
 \hline\hline
 24 & 1 & 1.11 & \textbf{17.24} & \textbf{6.09} & 45.7 & \textbf{47.7} & 66.6 & \textbf{66.4} \\ 
 
 24 & 4 & \textbf{1.09} & 17.49 & 6.28 & \textbf{45.8} & 43.5 & 67.2 & 73.2 \\ 

 12 & 1 & 1.16 & 17.25 & 6.14 & 44.8 & 47.0 & \textbf{66.4} & 71.7 \\ 
 
 12 & 4 & 1.10 & 17.48 & 6.16 & 45.0 & 45.5 & 70.0 & 72.6 \\
 
 6 & 1 & 1.32 & 19.47 & 6.26 & 39.8 & 46.1 & 82.1 & 79.5 \\ 
 
 6 & 4 & 1.32 & 19.47 & 6.47 & 39.8 & 42.0 & 82.1 & 81.2 \\[1ex] 
 
 \hline
\end{tabular}
\caption{Quantitative evaluation of generative modeling using a smaller decoder with  (24, 12 \& 6 coupling layers).}
\label{table:generative_modeling_small_models}
\end{center}
\end{table*}

\def\gefiga#1{{\includegraphics[clip,trim=8.5cm 3.3cm 8cm 6.5cm,width=0.15\linewidth]{figures/qualitative_8_coupling_layer/#1.png}}}
\begin{figure*}
	\centering	
	\setlength{\arrayrulewidth}{.2pt}%
	\setlength{\tabcolsep}{0.1pt}
	\renewcommand{\arraystretch}{3}
	\begin{tabular}{ccccccc}
	    \gefiga{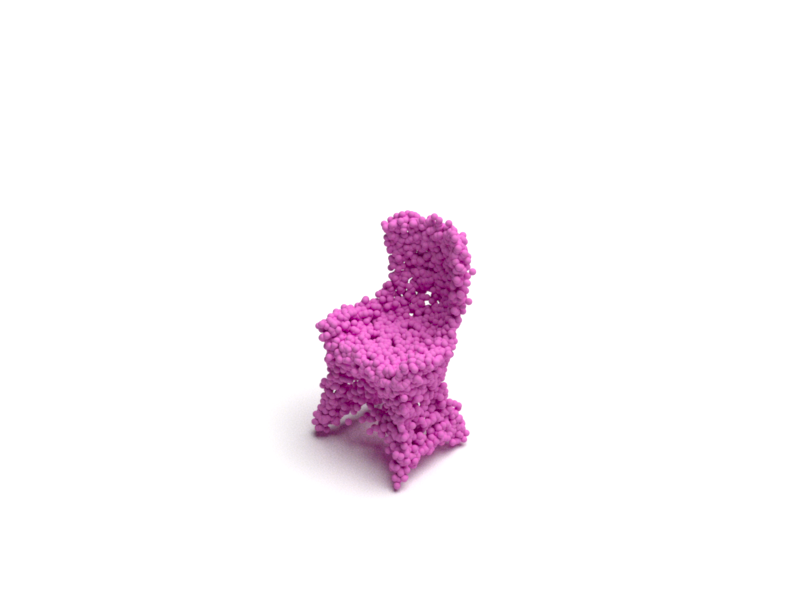} &
		\gefiga{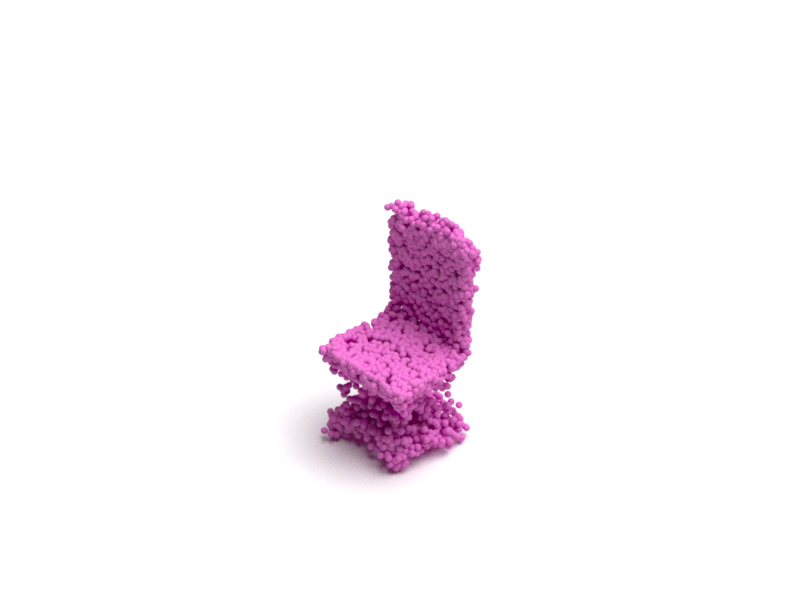} &
		\gefiga{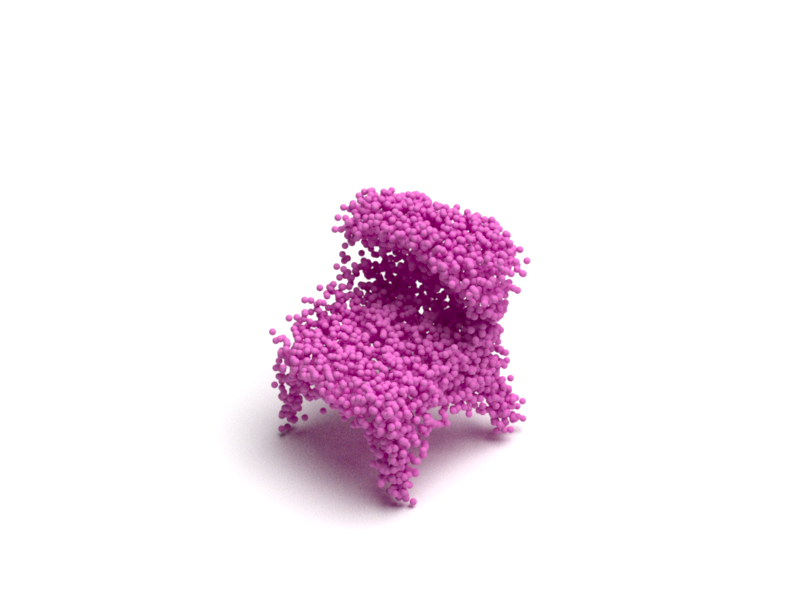} &
		\gefiga{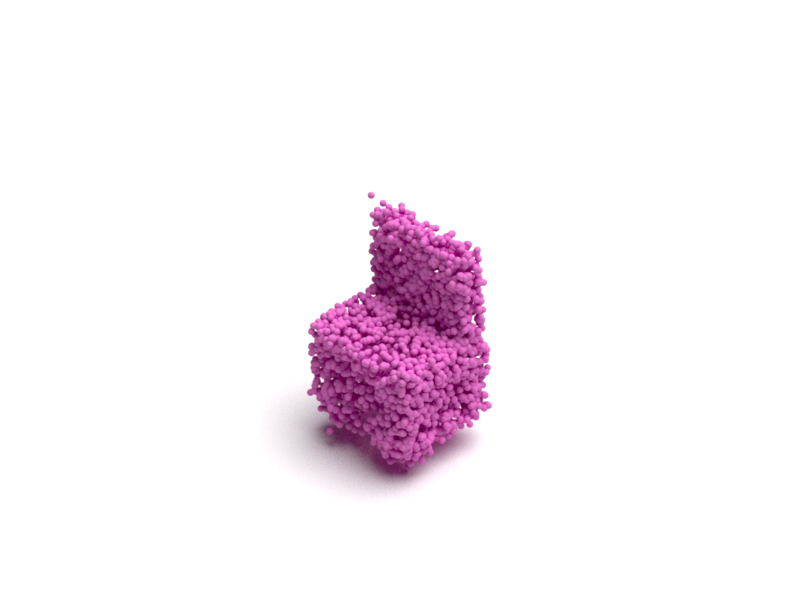} &
		\gefiga{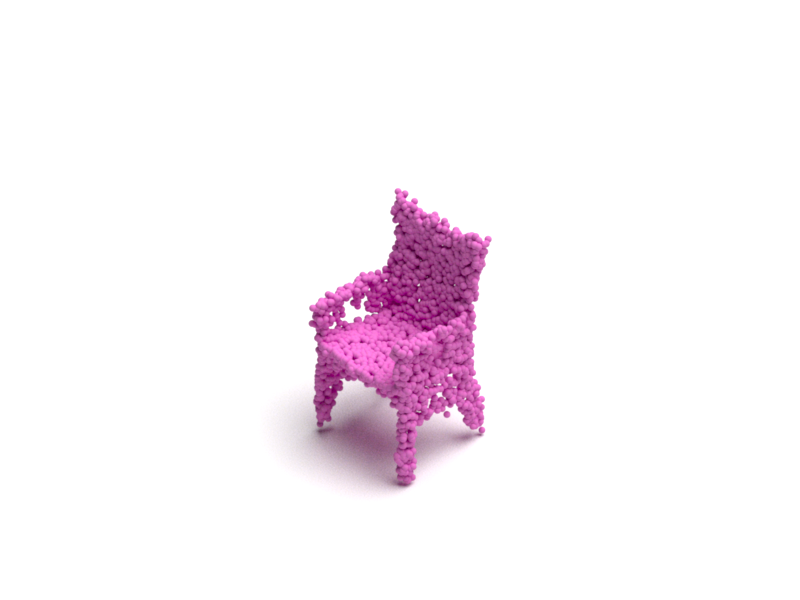} &
		\gefiga{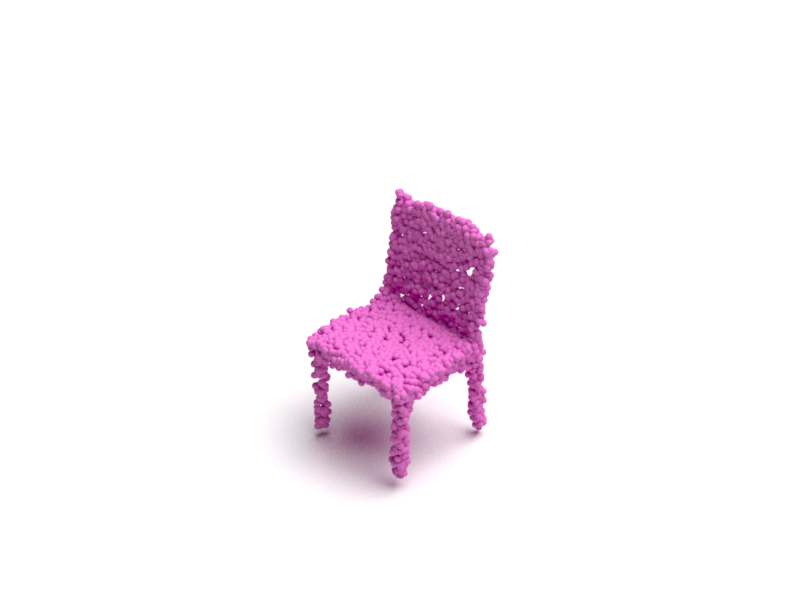} \\
	    \gefiga{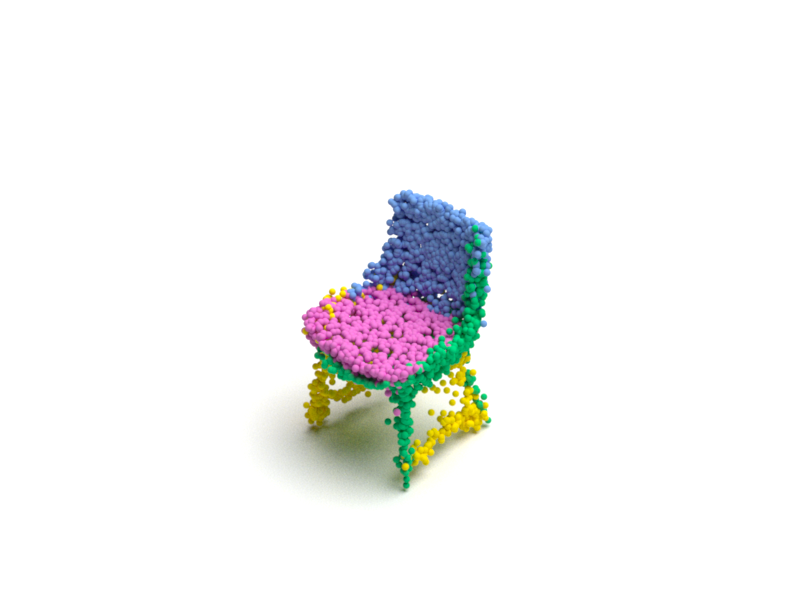} &
		\gefiga{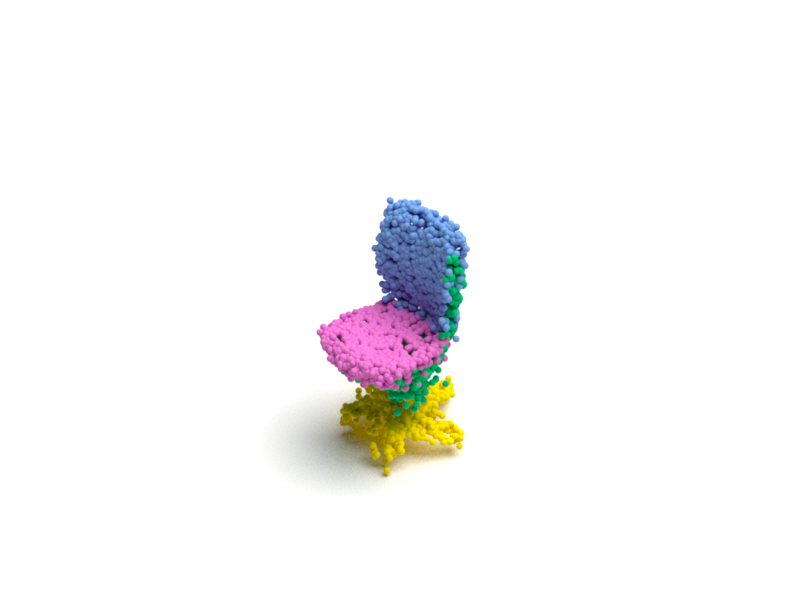} &
		\gefiga{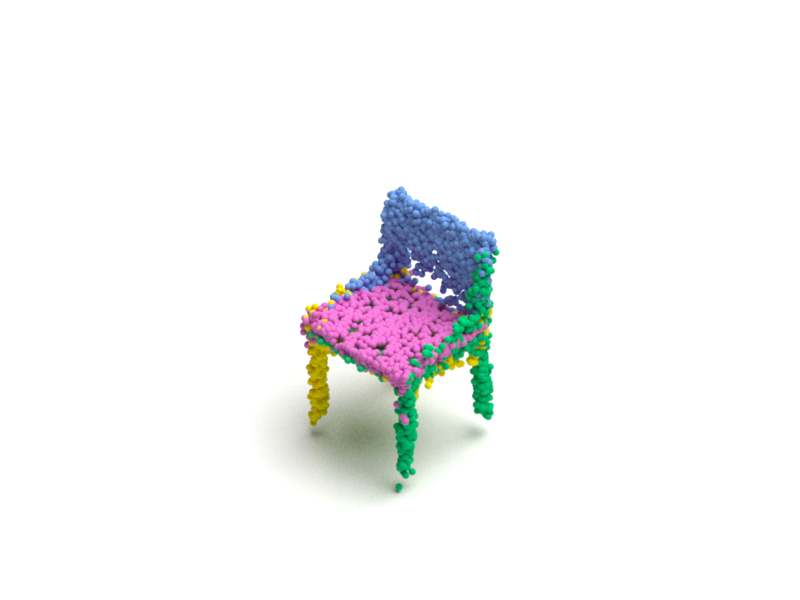} &
		\gefiga{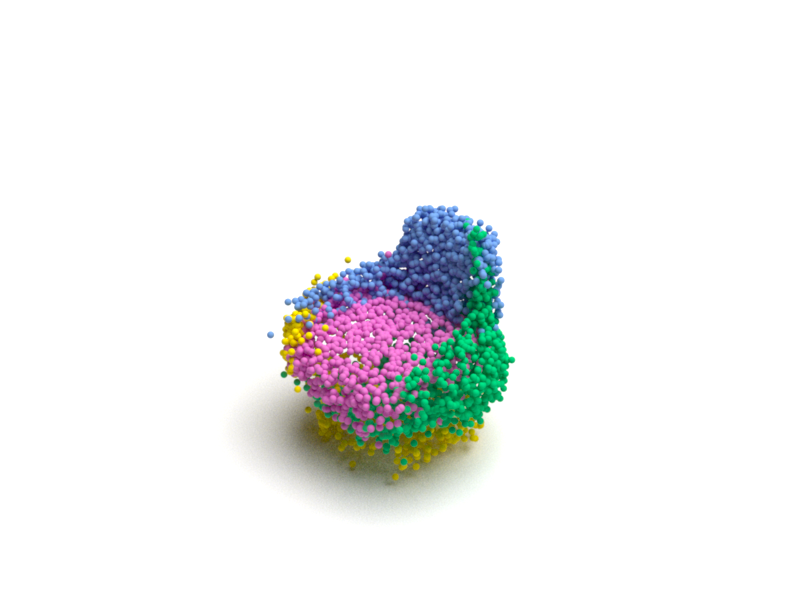} &
		\gefiga{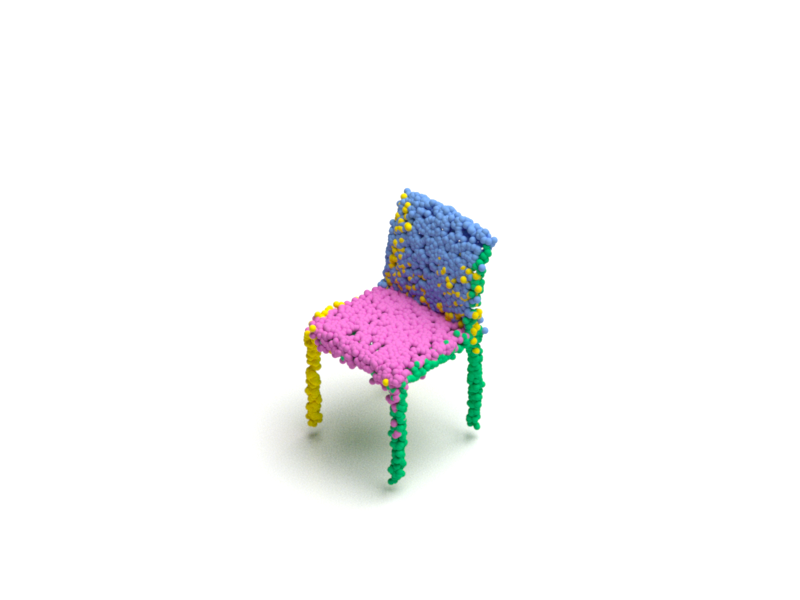} &
		\gefiga{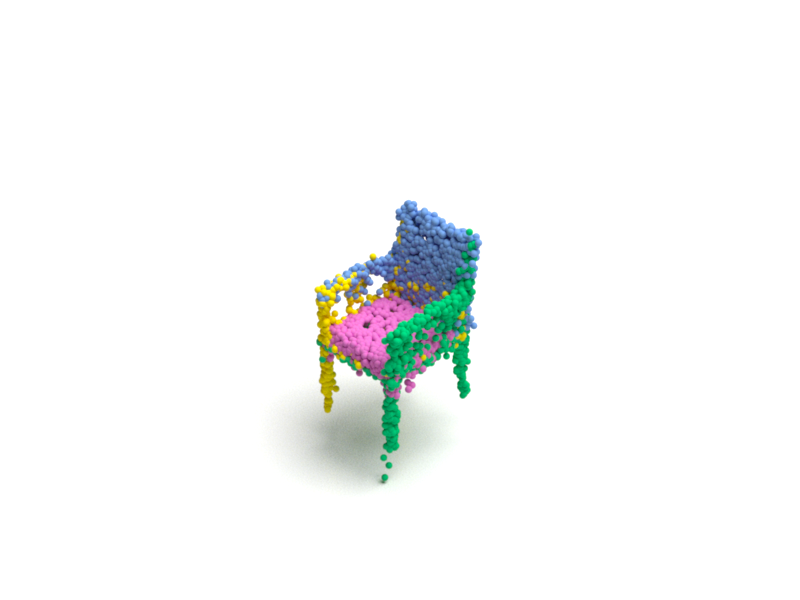} \\
	\end{tabular}
	\caption{Qualitative results of generating point clouds using a decoder with 1 \ac{nf} with 24 coupling layers (TOP) and 4 \acp{nf} with equivalent parameter count (BOTTOM).}
	%\vspace{-4mm}
	\label{fig:generative_modeling_qualitative_small_models}
\end{figure*}

%% file: sections_supplment/latent_space_interpolation.tex
In \cref{fig:interpolation} and \cref{fig:interpolation_one_flow}, we show qualitative examples of interpolating between latent representations learned by our models trained on airplane, car and chair. We sample two point clouds (the left-most and the right-most in the \cref{fig:interpolation}) and map them on their latent representations using our encoder. Subsequently, we linearly interpolate between these latent representation and reconstruct the result using our decoder model which is based on a mixture of 4 \acp{nf} (see in \cref{fig:interpolation}). Interestingly, our model also allows interpolating individual parts of one shape (see \cref{fig:interpolation_one_flow}). 

\def\gefiga#1#2{{\includegraphics[clip,trim=8.5cm 4cm 8cm 7.0cm,width=0.15\linewidth]{figures/interpolates/#1/#2.png}}}
\def\gefigb#1#2{{\includegraphics[clip,trim=8.5cm 3cm 8cm 8.0cm,width=0.15\linewidth]{figures/interpolates/#1/#2.png}}}
\def\gefigc#1#2{{\includegraphics[clip,trim=8.5cm 2cm 8cm 10.0cm,width=0.15\linewidth]{figures/interpolates/#1/#2.png}}}
\def\gefigd#1#2{{\includegraphics[clip,trim=8cm 3cm 7cm 7.0cm,width=0.15\linewidth]{figures/interpolates/#1/#2.png}}}
\def\gefige#1#2{{\includegraphics[clip,trim=10cm 7.5cm 9cm 7cm,width=0.15\linewidth]{figures/interpolates/#1/#2.png}}}
\begin{figure*}
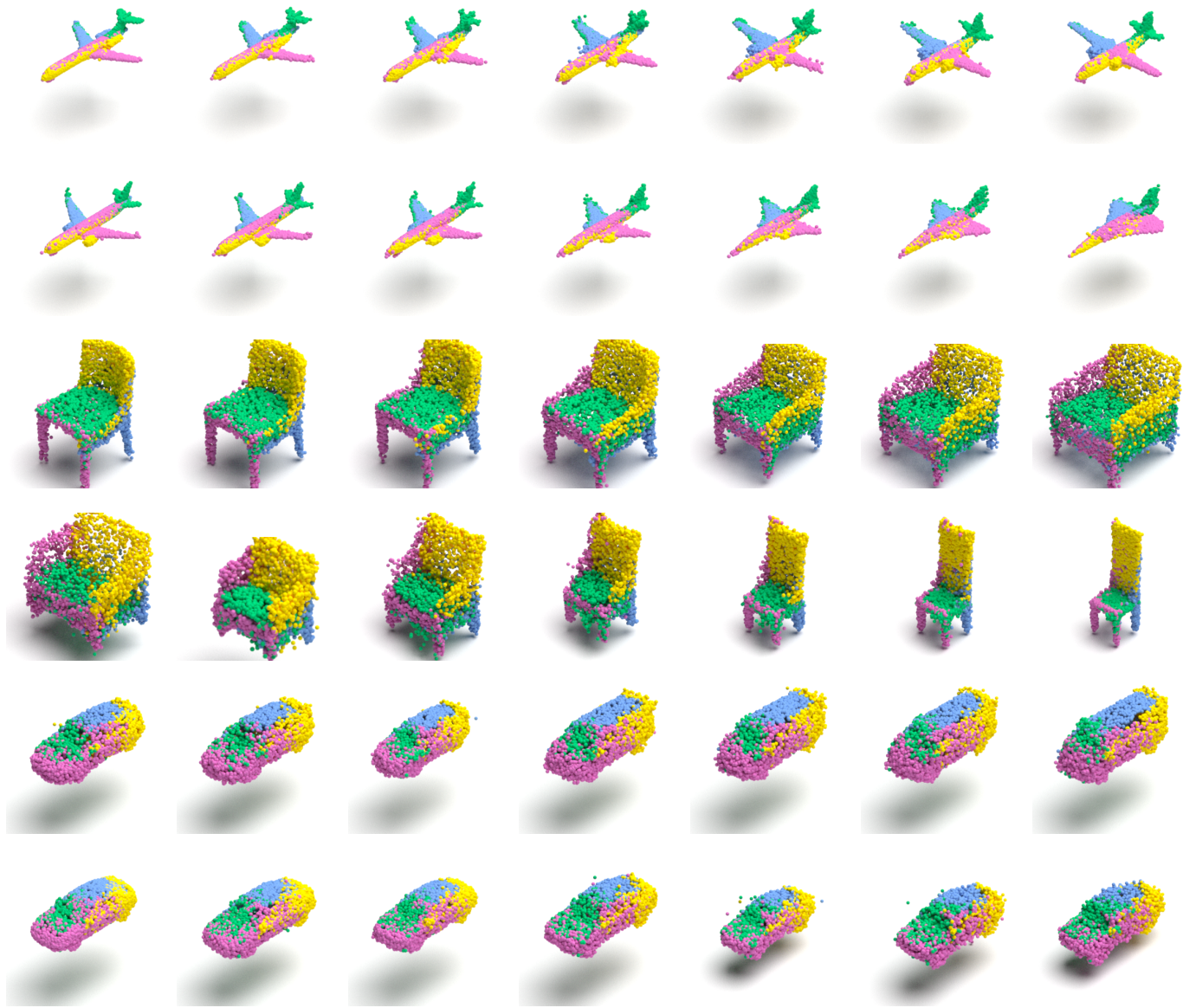

	%\centering	
	\setlength{\arrayrulewidth}{.2pt}%
	\setlength{\tabcolsep}{0.1pt}
	\renewcommand{\arraystretch}{3}
	\begin{tabular}{ccccccc}
	    \gefiga{airplane}{1_0} &
		\gefiga{airplane}{1_1} &
		\gefiga{airplane}{1_2} &
		\gefiga{airplane}{1_3} &
		\gefiga{airplane}{1_4} &
		\gefiga{airplane}{1_5} &
		\gefiga{airplane}{1_6} \\
		\gefiga{airplane}{2_0} &
		\gefiga{airplane}{2_1} &
		\gefiga{airplane}{2_2} &
		\gefiga{airplane}{2_3} &
		\gefiga{airplane}{2_4} &
		\gefiga{airplane}{2_5} &
		\gefiga{airplane}{2_6} \\
	    \gefiga{chair}{1_0} &
		\gefiga{chair}{1_1} &
		\gefiga{chair}{1_2} &
		\gefiga{chair}{1_4} &
		\gefigd{chair}{1_5} &
		\gefigd{chair}{1_6} &
		\gefigd{chair}{1_7} \\
		\gefiga{chair}{2_0} &
		\gefige{chair}{2_1} &
		\gefiga{chair}{2_2} &
		\gefiga{chair}{2_3} &
		\gefiga{chair}{2_4} &
		\gefiga{chair}{2_5} &
		\gefiga{chair}{2_6} \\
        %car
		\gefiga{car}{1_0} &
		\gefiga{car}{1_1} &
		\gefiga{car}{1_2} &
		\gefiga{car}{1_3} &
		\gefiga{car}{1_4} &
		\gefiga{car}{1_5} &
		\gefiga{car}{1_7} \\
		\gefiga{car}{2_0} &
		\gefiga{car}{2_1} &
		\gefiga{car}{2_2} &
		\gefiga{car}{2_3} &
		\gefigc{car}{2_4} &
		\gefigb{car}{2_6} &
		\gefigc{car}{2_7} \\
	\end{tabular}
	\caption{Qualitative examples of interpolation between two point clouds.}
	\label{fig:interpolation}
\end{figure*}

\def\inta#1#2{{\includegraphics[clip,trim=8.5cm 4cm 8cm 7.0cm,width=0.15\linewidth]{figures/interpolate_one_flow/#1/#2.png}}}
\def\intb#1#2{{\includegraphics[clip,trim=8cm 3cm 7cm 7.0cm,width=0.15\linewidth]{figures/interpolate_one_flow/#1/#2.png}}}
\def\intc#1#2{{\includegraphics[clip,trim=8.5cm 6cm 8cm 7.0cm,width=0.15\linewidth]{figures/interpolate_one_flow/#1/#2.png}}}
\def\intd#1#2{{\includegraphics[clip,trim=10cm 7cm 10cm 7.0cm,width=0.15\linewidth]{figures/interpolate_one_flow/#1/#2.png}}}
\def\inte#1#2{{\includegraphics[clip,trim=12cm 7.5cm 8cm 7.0cm,width=0.15\linewidth]{figures/interpolate_one_flow/#1/#2.png}}}
\def\intf#1#2{{\includegraphics[clip,trim=8cm 4cm 8cm 6.0cm,width=0.15\linewidth]{figures/interpolate_one_flow/#1/#2.png}}}
\def\intg#1#2{{\includegraphics[clip,trim=8cm 2cm 8cm 8.0cm,width=0.15\linewidth]{figures/interpolate_one_flow/#1/#2.png}}}
\def\inte#1#2{{\includegraphics[clip,trim=6cm 4.5cm 12cm 7.5cm,width=0.15\linewidth]{figures/interpolate_one_flow/#1/#2.png}}}
\def\inth#1#2{{\includegraphics[clip,trim=8.5cm 3cm 8cm 8.0cm,width=0.15\linewidth]{figures/interpolate_one_flow/#1/#2.png}}}
\begin{figure*}
	\centering	
	\setlength{\arrayrulewidth}{.2pt}%
	\setlength{\tabcolsep}{0.1pt}
	\renewcommand{\arraystretch}{1}
	\begin{tabular}{ccccccc}
		\inta{airplane}{1_0} &
		\inta{airplane}{1_1} &
		\inta{airplane}{1_2} &
		\inta{airplane}{1_3} &
		\intb{airplane}{1_4} &
		\inta{airplane}{1_5} &
		\inta{airplane}{1_6} \\
		\multicolumn{7}{c}{Interpolate the pink component}\\
		\inta{airplane}{6_0_inter} &
		\inta{airplane}{6_2_inter} &
		\inta{airplane}{6_3_inter} &
		\inta{airplane}{6_4_inter} &
		\inta{airplane}{6_5_inter} &
		\inta{airplane}{6_6_inter} &
		\inta{airplane}{6_8_inter} \\
		\multicolumn{7}{c} {Interpolate the pink and blue components} \\
	    %chair
	    \intb{chair}{8_0_inter} &
		\intb{chair}{8_2_inter} &
		\intf{chair}{8_3_inter} &
		\intf{chair}{8_4_inter} &
		\intf{chair}{8_5_inter} &
		\intf{chair}{8_6_inter} &
		\inta{chair}{8_8_inter} \\
		\multicolumn{7}{c} {Interpolate the yellow component}   \\
		\inta{chair}{2_0} &
		\inta{chair}{2_1} &
		\inta{chair}{2_2} &
		\inta{chair}{2_3} &
		\inta{chair}{2_4} &
		\inta{chair}{2_5} &
		\inta{chair}{2_6} \\
		\multicolumn{7}{c}{Interpolate the yellow component} \\
		\intg{car}{2_0_inter} &
		\inta{car}{2_2_inter} &
		\inta{car}{2_3_inter} &
		\inta{car}{2_4_inter} &
		\inta{car}{2_5_inter} &
		\inta{car}{2_6_inter} &
		\inta{car}{2_8_inter} \\
		\multicolumn{7}{c} {Interpolate the yellow and pink components} \\
		\inta{car}{3_0_inter} &
		\inta{car}{3_2_inter} &
		\inta{car}{3_3_inter} &
		\inte{car}{3_4_inter} &
		\inth{car}{3_5_inter} &
		\inta{car}{3_6_inter} &
		\inta{car}{3_8_inter} \\
		\multicolumn{7}{c}{Interpolate the yellow and pink components}
	\end{tabular}
	\caption{Qualitative examples of interpolating individual components of the mixture of \acp{nf}.}
	\label{fig:interpolation_one_flow}
\end{figure*}
\vspace{-4.3mm}

%% file: sections_supplment/sampling.tex
Here, we investigate whether mixtures of \acp{nf} can upsample sparse point clouds. In \cref{fig:sparse_sampling} we show qualitative examples. Following DPF \cite{klokov2020discrete}, a sparse point cloud of 512 is upsampled to 32768 points. In line with previous results, mixtures of \acp{nf} yield better results in high-frequency regions.

\def\gefiga#1#2{{\includegraphics[clip,trim=10cm 4cm 10cm 7.0cm,width=0.12\linewidth]{figures/sampling/#1/#2.png}}}
\def\gefigb#1#2{{\includegraphics[clip,trim=10cm 4cm 10cm 7.0cm, width=0.12\linewidth]{figures/sampling/#1/#2.png}}}
\def\gefigc#1#2{{\includegraphics[clip,trim=9cm 3cm 9cm 6.5cm,width=0.12\linewidth]{figures/sampling/#1/#2.png}}}
\begin{figure*}
	\centering	
	\setlength{\arrayrulewidth}{.2pt}%
	\setlength{\tabcolsep}{0.2pt}
	\renewcommand{\arraystretch}{1}
	\begin{tabular}{cccccccc}
	    \gefiga{airplane}{1_gt} &
	    \gefiga{airplane}{1_dpf} &
		\gefiga{airplane}{1_resample} &
		\gefigb{airplane}{1_gtd} &
	    \gefiga{airplane}{2_gt} &
	    \gefiga{airplane}{2_dpf} &
		\gefiga{airplane}{2_resample} &
		\gefiga{airplane}{2_gtd} \\
		\gefiga{airplane}{5_gt} &
		\gefiga{airplane}{5_dpf} &
        \gefiga{airplane}{5_resample} &
        \gefiga{airplane}{5_gtd} &
		\gefiga{airplane}{9_gt} &
		\gefiga{airplane}{9_dpf} &
		\gefiga{airplane}{9_resample} &
		\gefiga{airplane}{9_gtd}\\
	    \gefigc{chair}{2_gt} &
	    \gefigc{chair}{2_dpf} &
		\gefigc{chair}{2_resample} &
		\gefigc{chair}{2_gtd} &
		\gefiga{chair}{3_gt} &
		\gefiga{chair}{3_dpf} &
		\gefiga{chair}{3_resample} &
		\gefiga{chair}{3_gtd}\\
		\gefigc{chair}{4_gt} &
		\gefigc{chair}{4_dpf} &
        \gefigc{chair}{4_resample} &
        \gefigc{chair}{4_gtd} &
		\gefiga{chair}{9_gt} &
		\gefiga{chair}{9_dpf} &
		\gefiga{chair}{9_resample} &
		\gefiga{chair}{9_gtd}\\
	    \gefiga{car}{1_gt} &
	    \gefiga{car}{1_dpf} &
		\gefiga{car}{1_resample} &
		\gefiga{car}{1_gtd} &
		\gefiga{car}{2_gt} &
		\gefiga{car}{2_dpf} &
		\gefiga{car}{2_resample} &
		\gefiga{car}{2_gtd}\\
		\gefiga{car}{5_gt} &
		\gefiga{car}{5_dpf} &
        \gefiga{car}{5_resample} &
        \gefiga{car}{5_gtd} &
		\gefiga{car}{8_gt} &
		\gefiga{car}{8_dpf} &
		\gefiga{car}{8_resample} &
		\gefiga{car}{8_gtd}\\
		\footnotesize Sparse Input & 
		\footnotesize DPF \cite{klokov2020discrete}& 
		\footnotesize Ours &
		\footnotesize Ground truth &
		\footnotesize Sparse Input & 
		\footnotesize DPF \cite{klokov2020discrete}& 
		\footnotesize Ours &
		\footnotesize Ground truth \\
	\end{tabular}
	\caption{Qualitative examples of upsampling sparse point clouds.}
	\label{fig:sparse_sampling}
\end{figure*}

\vspace{-4.3mm}

%% file: sections_supplment/qualitative_results_on_generation.tex
We show additional qualitative examples of generated point clouds using mixtures of \acp{nf} in \cref{fig:generation}.

\def\gefiga#1#2{{\includegraphics[clip,trim=8.5cm 4cm 8cm 7.0cm,width=0.15\linewidth]{figures/generation/#1/#2.png}}}
\def\gefigd#1#2{{\includegraphics[clip,trim=8cm 2cm 7cm 9.0cm,width=0.15\linewidth]{figures/generation/#1/#2.png}}}
\def\gefige#1#2{{\includegraphics[clip,trim=8cm 4cm 8cm 6.8cm,width=0.15\linewidth]{figures/generation/#1/#2.png}}}
\def\gefigf#1#2{{\includegraphics[clip,trim=7cm 2cm 7cm 7cm,width=0.15\linewidth]{figures/generation/#1/#2.png}}}
\def\gefigg#1#2{{\includegraphics[clip,trim=6cm 2cm 6cm 6.5cm,width=0.15\linewidth]{figures/generation/#1/#2.png}}}
\begin{figure*}
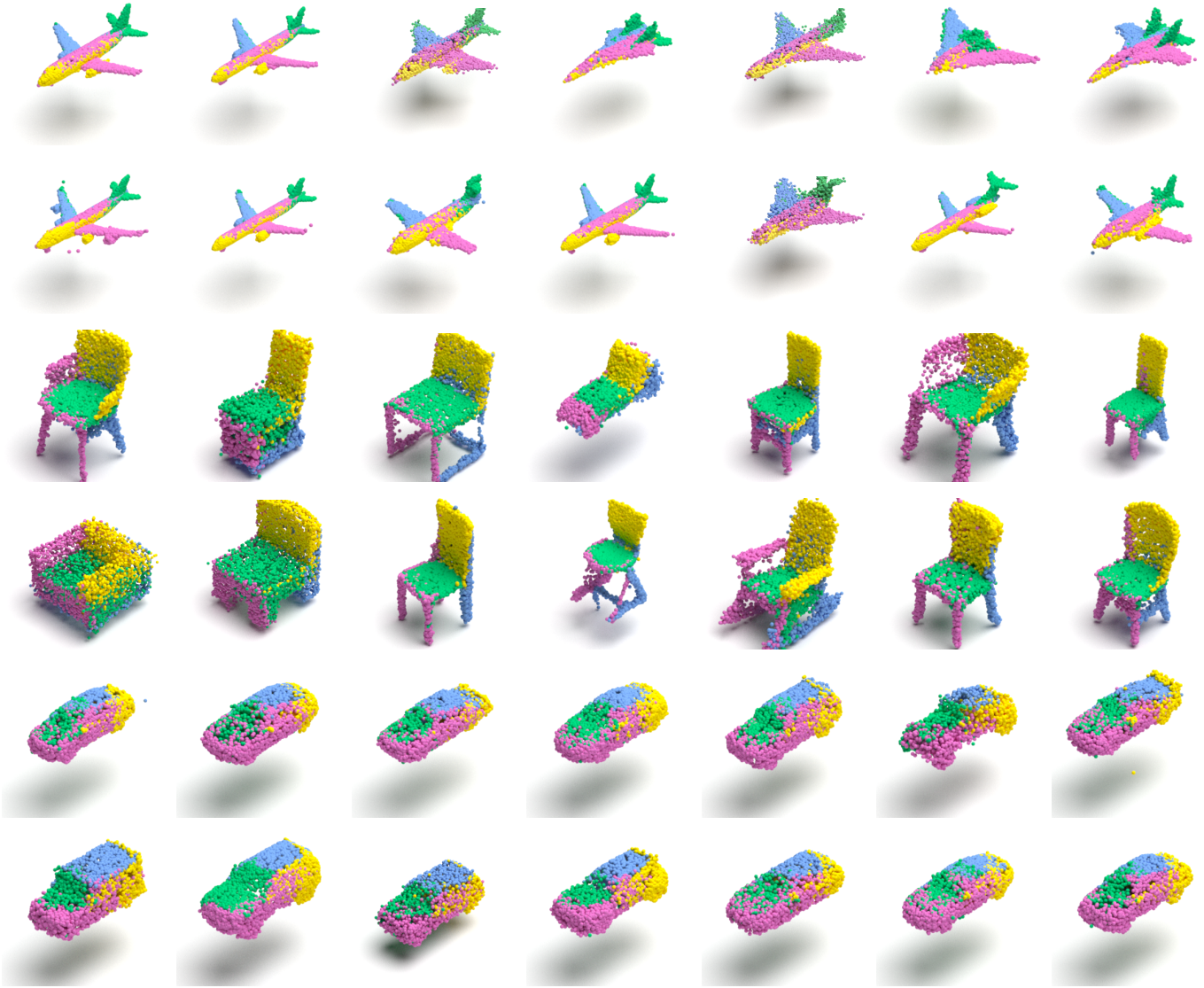

	%\centering	
	\setlength{\arrayrulewidth}{.2pt}%
	\setlength{\tabcolsep}{0.1pt}
	\renewcommand{\arraystretch}{2}
	\begin{tabular}{ccccccc}
	    \gefiga{airplane}{0} &
		\gefiga{airplane}{1} &
		\gefigg{airplane}{2} &
		\gefiga{airplane}{3} &
		\gefigg{airplane}{4} &
		\gefiga{airplane}{5} &
		\gefiga{airplane}{6} \\
        \gefiga{airplane}{7} &
		\gefiga{airplane}{8} &
		\gefiga{airplane}{9} &
		\gefiga{airplane}{10} &
		\gefigg{airplane}{11} &
		\gefiga{airplane}{12} &
		\gefiga{airplane}{13} \\
		\gefiga{chair}{0} &
		\gefiga{chair}{1} &
		\gefige{chair}{2} &
		\gefiga{chair}{3} &
		\gefiga{chair}{4} &
		\gefige{chair}{5} &
		\gefiga{chair}{6} \\
        \gefigf{chair}{7} &
		\gefigf{chair}{8} &
		\gefiga{chair}{9} &
		\gefiga{chair}{10} &
		\gefiga{chair}{11} &
		\gefiga{chair}{12} &
		\gefiga{chair}{13} \\
	    \gefiga{car}{0} &
		\gefiga{car}{1} &
		\gefiga{car}{2} &
		\gefiga{car}{3} &
		\gefiga{car}{4} &
		\gefiga{car}{5} &
		\gefiga{car}{6} \\
        \gefiga{car}{7} &
		\gefiga{car}{8} &
		\gefigd{car}{9} &
		\gefiga{car}{10} &
		\gefiga{car}{11} &
		\gefiga{car}{12} &
		\gefiga{car}{13} \\
	\end{tabular}
	\caption{Additional qualitative examples of generated point clouds using mixtures of \acp{nf}.}
	\label{fig:generation}
\end{figure*}
\vspace{-4.3mm}

%% file: sections_supplment/qualitative_results_on_ae.tex
We show additional qualitative examples on autoencoding of point clouds using mixtures of \acp{nf} in \cref{fig:autoencoding}.

\def\gefiga#1#2{{\includegraphics[clip,trim=8.5cm 4cm 8cm 7.0cm,width=0.18\linewidth]{figures/ae/#1/#2.png}}}
\begin{figure*}
	%\centering	
	\setlength{\arrayrulewidth}{.2pt}%
	\setlength{\tabcolsep}{0.1pt}
	\renewcommand{\arraystretch}{1}
	\begin{tabular}{ccccccc}
	    \gefiga{airplane}{0_gt} &
	    \gefiga{airplane}{0_dpf} &
		\gefiga{airplane}{0_gen} &
		\gefiga{airplane}{1_gt} &
		\gefiga{airplane}{1_dpf} &
		\gefiga{airplane}{1_gen} \\
		\gefiga{airplane}{5_gt} &
		\gefiga{airplane}{5_dpf} &
        \gefiga{airplane}{5_gen} &
		\gefiga{airplane}{7_gt} &
		\gefiga{airplane}{7_dpf} &
		\gefiga{airplane}{7_gen} \\
	    \gefiga{chair}{0_gt} &
	    \gefiga{chair}{0_dpf} &
		\gefiga{chair}{0_gen} &
		\gefiga{chair}{3_gt} &
		\gefiga{chair}{3_dpf} &
		\gefiga{chair}{3_gen} \\
		\gefiga{chair}{5_gt} &
		\gefiga{chair}{5_dpf} &
        \gefiga{chair}{5_gen} &
		\gefiga{chair}{6_gt} &
		\gefiga{chair}{6_dpf} &
		\gefiga{chair}{6_gen} \\
	    \gefiga{car}{0_gt} &
	    \gefiga{car}{0_dpf} &
		\gefiga{car}{0_gen} &
		\gefiga{car}{3_gt} &
		\gefiga{car}{3_dpf} &
		\gefiga{car}{3_gen} \\
		\gefiga{car}{6_gt} &
		\gefiga{car}{6_dpf} &
        \gefiga{car}{6_gen} &
		\gefiga{car}{7_gt} &
		\gefiga{car}{7_dpf} &
		\gefiga{car}{7_gen} \\
		\footnotesize Ground Truth & 
		\footnotesize DPF \cite{klokov2020discrete} &
		\footnotesize Ours &
		\footnotesize Ground Truth & 
		\footnotesize DPF \cite{klokov2020discrete}& 
		\footnotesize Ours \\
	\end{tabular}
	\caption{Additional qualitative examples of autoencoding using mixtures of \acp{nf}.}
	\label{fig:autoencoding}
\end{figure*}
\vspace{-4.3mm}

%% file: sections_supplment/qualitative_results_on_svr.tex
We show additional qualitative examples on \ac{svr} of point clouds using mixtures of \acp{nf} in \cref{fig:SVR}.

\def\gefiga#1#2{{\includegraphics[clip,trim=8.5cm 4cm 8cm 7.0cm,width=0.12\linewidth]{figures/svr_more/#1/#2.png}}}
\def\gefigb#1#2{{\includegraphics[clip,trim=8.5cm 4cm 8cm 6.0cm,width=0.12\linewidth]{figures/svr_more/#1/#2.png}}}
\begin{figure*}
	%\centering	
	\setlength{\arrayrulewidth}{.2pt}%
	\setlength{\tabcolsep}{0.1pt}
	\renewcommand{\arraystretch}{1}
	\begin{tabular}{cccccccc}
	    \includegraphics[clip, trim=4cm 2cm 4cm 2cm, width=0.12\linewidth]{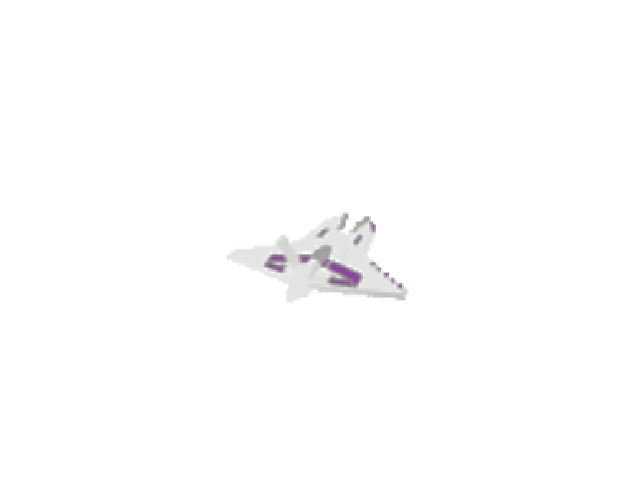} &
	    \gefiga{airplane}{0_gt} &
	    \gefiga{airplane}{0_dpf} &
		\gefiga{airplane}{0_gen} &
		\includegraphics[clip, trim=4cm 2cm 4cm 2cm, width=0.12\linewidth]{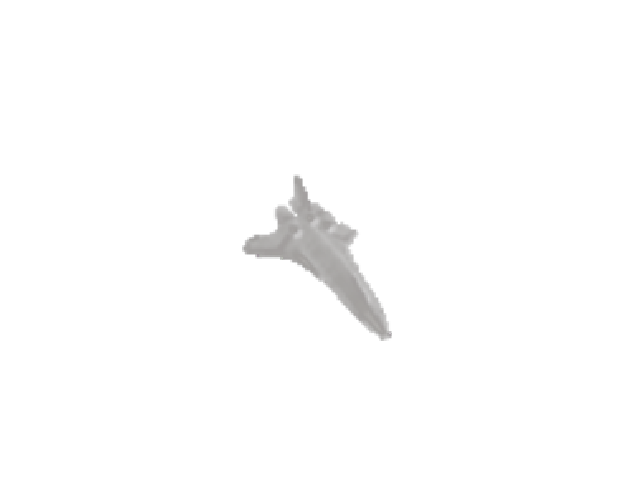} &
		\gefiga{airplane}{1_gt} &
		\gefiga{airplane}{1_dpf} &
		\gefiga{airplane}{1_gen} \\
	    \includegraphics[clip, trim=4cm 2cm 4cm 2cm, width=0.12\linewidth]{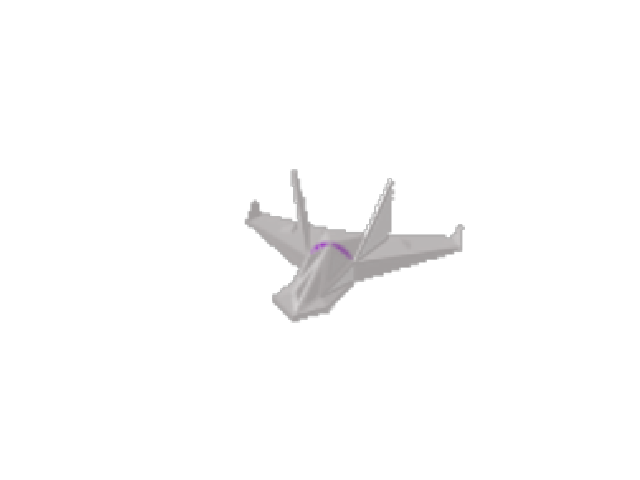} &
		\gefiga{airplane}{2_gt} &
		\gefiga{airplane}{2_dpf} &
        \gefiga{airplane}{2_gen} &
        \includegraphics[clip, trim=4cm 2cm 4cm 2cm, width=0.12\linewidth]{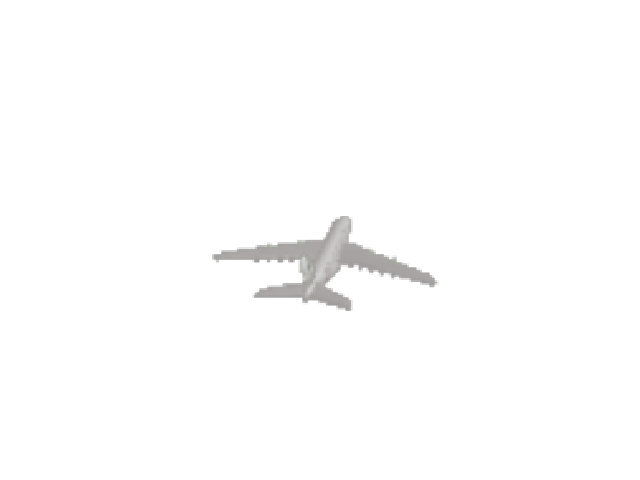} &
		\gefiga{airplane}{3_gt} &
		\gefiga{airplane}{3_dpf} &
		\gefiga{airplane}{3_gen} \\
		\includegraphics[clip, trim=4cm 2cm 4cm 2cm, width=0.12\linewidth]{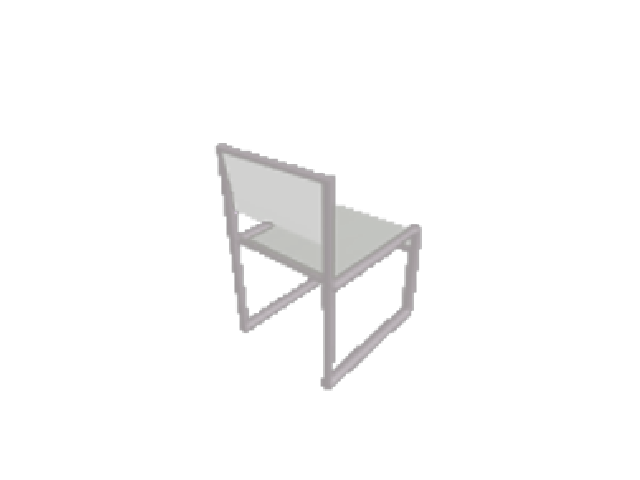} &
	    \gefiga{chair}{0_gt} &
	    \gefiga{chair}{0_dpf} &
		\gefiga{chair}{0_gen} &
		\includegraphics[clip, trim=4cm 2cm 4cm 2cm, width=0.12\linewidth]{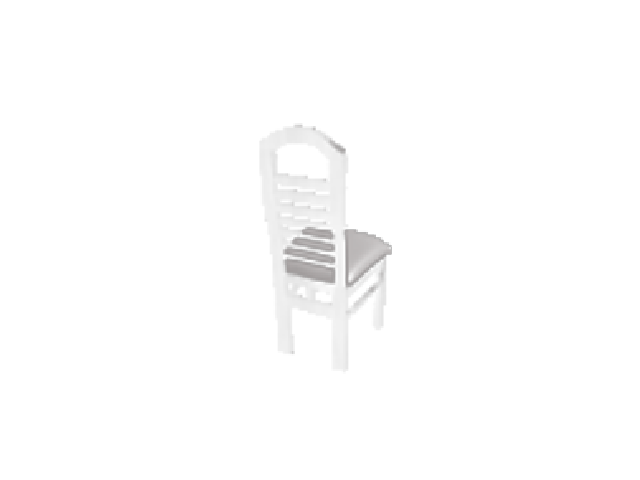} &
		\gefiga{chair}{1_gt} &
		\gefiga{chair}{1_dpf} &
		\gefiga{chair}{1_gen} \\
		\includegraphics[clip, trim=4cm 2cm 4cm 2cm, width=0.12\linewidth]{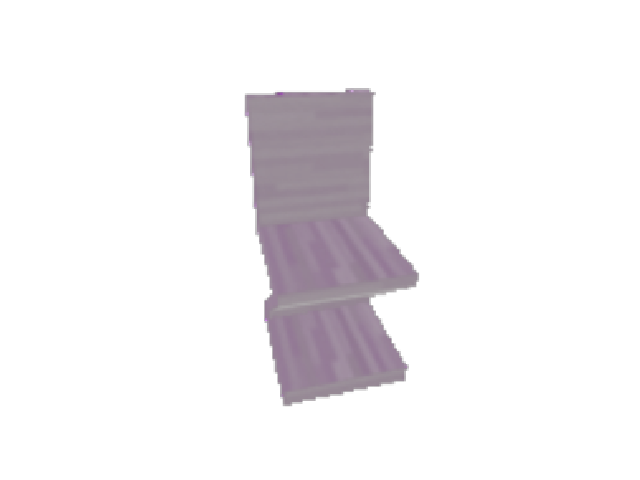} &
		\gefiga{chair}{2_gt} &
		\gefiga{chair}{2_dpf} &
        \gefiga{chair}{2_gen} &
        \includegraphics[clip, trim=4cm 2cm 4cm 2cm, width=0.12\linewidth]{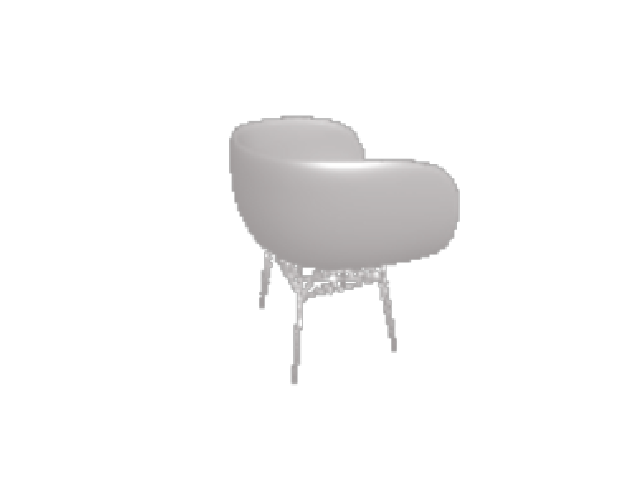} &
		\gefigb{chair}{3_gt} &
		\gefigb{chair}{3_dpf} &
		\gefigb{chair}{3_gen} \\
		\includegraphics[clip, trim=4cm 2cm 4cm 2cm, width=0.12\linewidth]{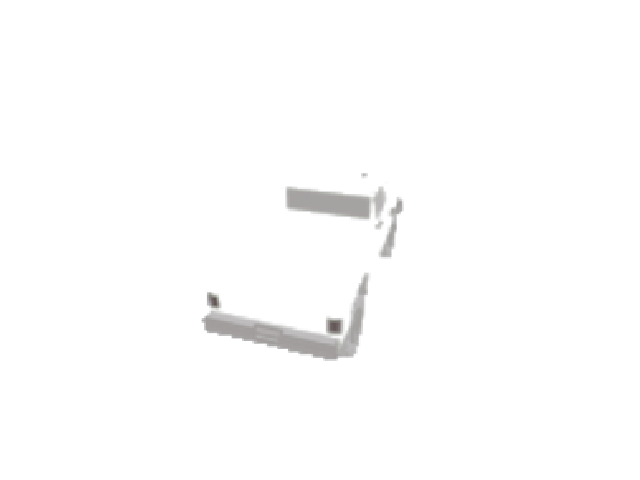} &
	    \gefiga{car}{0_gt} &
	    \gefiga{car}{0_dpf} &
		\gefiga{car}{0_gen} &
		\includegraphics[clip, trim=4cm 2cm 3cm 2cm, width=0.12\linewidth]{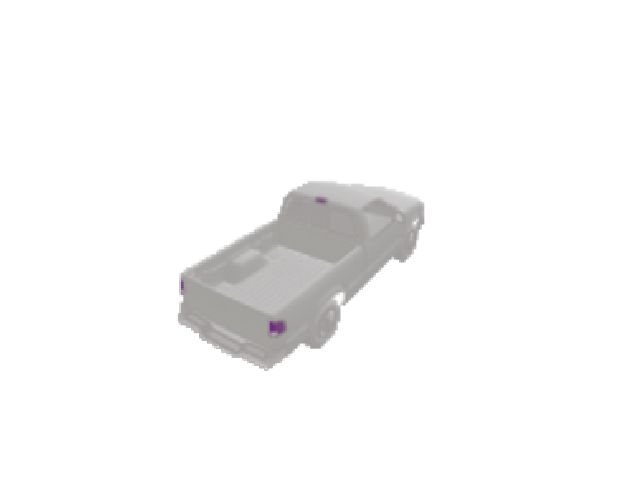} &
		\gefiga{car}{1_gt} &
		\gefiga{car}{1_dpf} &
		\gefiga{car}{1_gen} \\
		\includegraphics[clip, trim=4cm 2cm 4cm 2cm, width=0.12\linewidth]{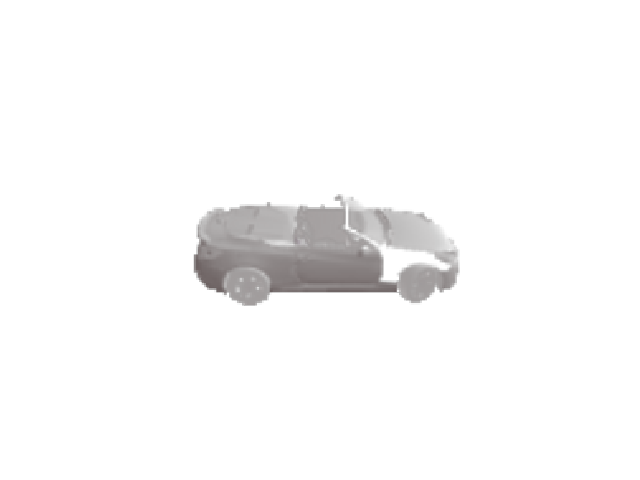} &
		\gefiga{car}{2_gt} &
		\gefiga{car}{2_dpf} &
        \gefiga{car}{2_gen} &
        \includegraphics[clip, trim=4cm 2cm 4cm 2cm, width=0.12\linewidth]{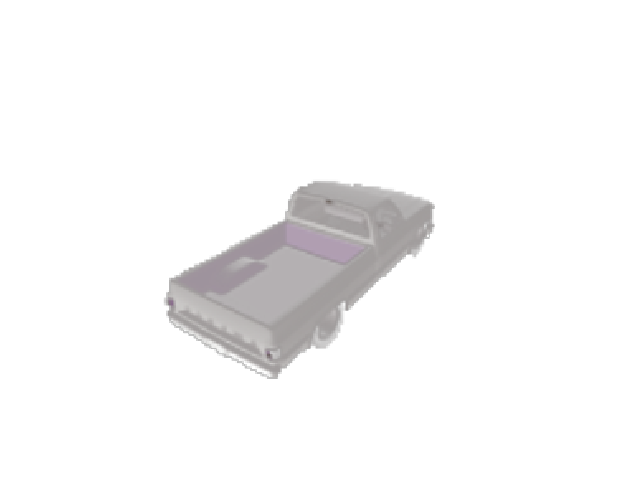} &
		\gefiga{car}{3_gt} &
		\gefiga{car}{3_dpf} &
		\gefiga{car}{3_gen} \\
		\includegraphics[clip, trim=4cm 2cm 4cm 2cm, width=0.12\linewidth]{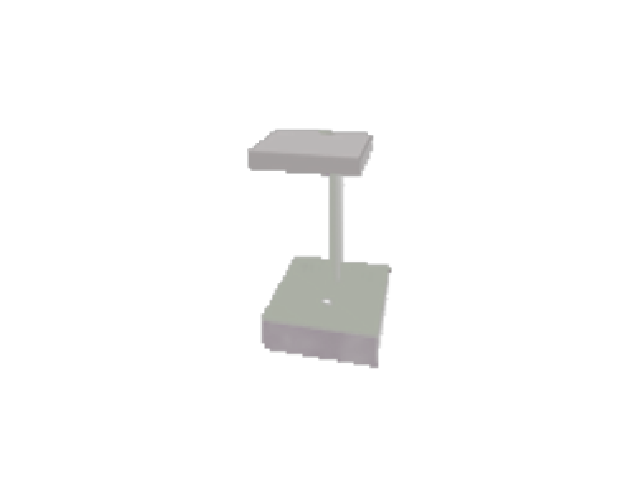} &
	    \gefiga{lamp}{0_gt} &
	    \gefiga{lamp}{0_dpf} &
		\gefiga{lamp}{0_gen} &
		\includegraphics[clip, trim=4cm 2cm 4cm 2cm, width=0.12\linewidth]{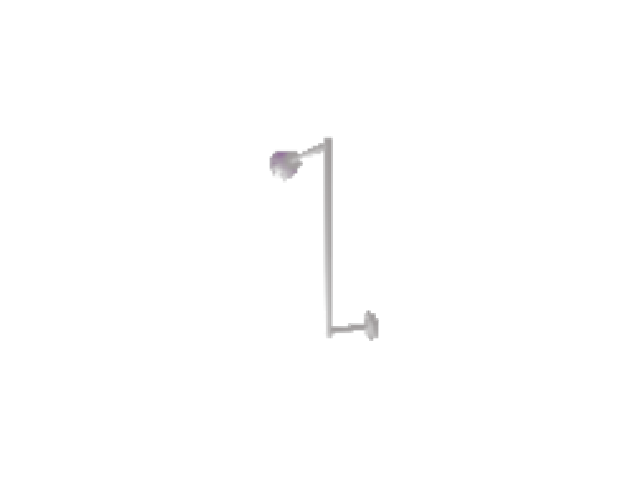} &
		\gefiga{lamp}{1_gt} &
		\gefiga{lamp}{1_dpf} &
		\gefiga{lamp}{1_gen} \\
		\includegraphics[clip, trim=4cm 2cm 4cm 2cm, width=0.12\linewidth]{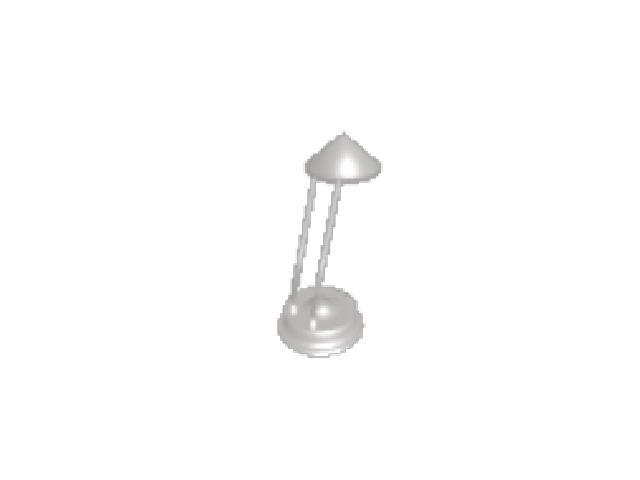} &
		\gefiga{lamp}{2_gt} &
		\gefiga{lamp}{2_dpf} &
        \gefiga{lamp}{2_gen} &
        \includegraphics[clip, trim=4cm 2cm 4cm 2cm, width=0.12\linewidth]{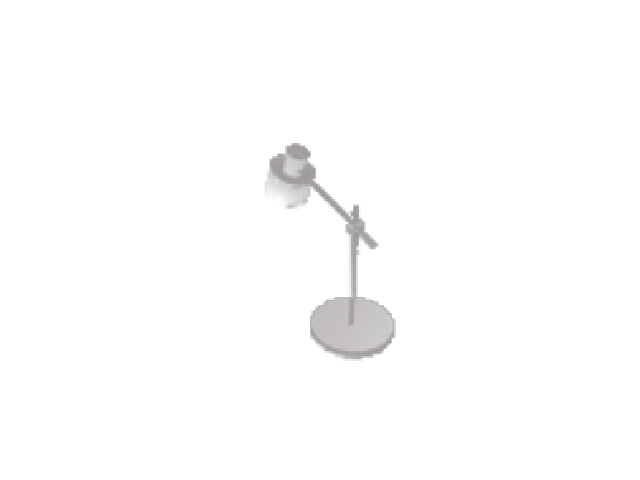} &
		\gefigb{lamp}{3_gt} &
		\gefigb{lamp}{3_dpf} &
		\gefigb{lamp}{3_gen} \\
		\footnotesize Image & 
		\footnotesize Ground Truth & 
		\footnotesize DPF \cite{klokov2020discrete} &
		\footnotesize Ours &
		\footnotesize Image & 
		\footnotesize Ground Truth & 
		\footnotesize DPF \cite{klokov2020discrete}& 
		\footnotesize Ours \\
	\end{tabular}
	\caption{Additional qualitative examples of SVR using mixtures of \acp{nf}.}
	\label{fig:SVR}
\end{figure*}
\vspace{-4.3mm}

%% file: main.bbl
\begin{thebibliography}{10}\itemsep=-1pt

\bibitem{achlioptas2018learning}
Panos Achlioptas, Olga Diamanti, Ioannis Mitliagkas, and Leonidas Guibas.
\newblock Learning representations and generative models for 3d point clouds.
\newblock In {\em International conference on machine learning}, pages 40--49.
  PMLR, 2018.

\bibitem{ardizzone2019guided}
Lynton Ardizzone, Carsten L{\"u}th, Jakob Kruse, Carsten Rother, and Ullrich
  K{\"o}the.
\newblock Guided image generation with conditional invertible neural networks.
\newblock {\em arXiv preprint arXiv:1907.02392}, 2019.

\bibitem{arroyo2021variational}
Diego~Martin Arroyo, Janis Postels, and Federico Tombari.
\newblock Variational transformer networks for layout generation.
\newblock {\em Proceedings of the IEEE/CVF Conference on Computer Vision and
  Pattern Recognition}, 2021.

\bibitem{ben2018multi}
Heli Ben-Hamu, Haggai Maron, Itay Kezurer, Gal Avineri, and Yaron Lipman.
\newblock Multi-chart generative surface modeling.
\newblock {\em ACM Transactions on Graphics (TOG)}, 37(6):1--15, 2018.

\bibitem{NEURIPS2020_05192834}
Johann Brehmer and Kyle Cranmer.
\newblock Flows for simultaneous manifold learning and density estimation.
\newblock In H. Larochelle, M. Ranzato, R. Hadsell, M.~F. Balcan, and H. Lin,
  editors, {\em Advances in Neural Information Processing Systems}, volume~33,
  pages 442--453, 2020.

\bibitem{cai2020learning}
Ruojin Cai, Guandao Yang, Hadar Averbuch-Elor, Zekun Hao, Serge Belongie, Noah
  Snavely, and Bharath Hariharan.
\newblock Learning gradient fields for shape generation.
\newblock {\em arXiv preprint arXiv:2008.06520}, 2020.

\bibitem{chang2015shapenet}
Angel~X Chang, Thomas Funkhouser, Leonidas Guibas, Pat Hanrahan, Qixing Huang,
  Zimo Li, Silvio Savarese, Manolis Savva, Shuran Song, Hao Su, et~al.
\newblock Shapenet: An information-rich 3d model repository.
\newblock {\em arXiv preprint arXiv:1512.03012}, 2015.

\bibitem{chen2018neural}
Ricky T.~Q. Chen, Yulia Rubanova, Jesse Bettencourt, and David~K Duvenaud.
\newblock Neural ordinary differential equations.
\newblock In {\em Advances in Neural Information Processing Systems},
  volume~31, 2018.

\bibitem{chen2016variational}
Xi Chen, Diederik~P Kingma, Tim Salimans, Yan Duan, Prafulla Dhariwal, John
  Schulman, Ilya Sutskever, and Pieter Abbeel.
\newblock Variational lossy autoencoder.
\newblock {\em International Conference on Learning Representations}, 2017.

\bibitem{choy20163d}
Christopher~B Choy, Danfei Xu, JunYoung Gwak, Kevin Chen, and Silvio Savarese.
\newblock 3d-r2n2: A unified approach for single and multi-view 3d object
  reconstruction.
\newblock In {\em European conference on computer vision}, pages 628--644.
  Springer, 2016.

\bibitem{cornish2020relaxing}
Rob Cornish, Anthony Caterini, George Deligiannidis, and Arnaud Doucet.
\newblock Relaxing bijectivity constraints with continuously indexed
  normalising flows.
\newblock In {\em International Conference on Machine Learning}, pages
  2133--2143. PMLR, 2020.

\bibitem{dinh2014nice}
Laurent Dinh, David Krueger, and Yoshua Bengio.
\newblock Nice: Non-linear independent components estimation.
\newblock {\em arXiv preprint arXiv:1410.8516, 2014}, 2014.

\bibitem{dinh2016density}
Laurent Dinh, Jascha Sohl-Dickstein, and Samy Bengio.
\newblock Density estimation using real nvp.
\newblock {\em International Conference on Learning Representations}, 2017.

\bibitem{dinh2019rad}
Laurent Dinh, Jascha Sohl-Dickstein, Razvan Pascanu, and Hugo Larochelle.
\newblock A rad approach to deep mixture models.
\newblock {\em arXiv preprint arXiv:1903.07714}, 2019.

\bibitem{fan2017point}
Haoqiang Fan, Hao Su, and Leonidas~J Guibas.
\newblock A point set generation network for 3d object reconstruction from a
  single image.
\newblock In {\em Proceedings of the IEEE conference on computer vision and
  pattern recognition}, pages 605--613, 2017.

\bibitem{gadelha2018multiresolution}
Matheus Gadelha, Rui Wang, and Subhransu Maji.
\newblock Multiresolution tree networks for 3d point cloud processing.
\newblock In {\em Proceedings of the European Conference on Computer Vision
  (ECCV)}, pages 103--118, 2018.

\bibitem{giaquinto2020gradient}
Robert Giaquinto and Arindam Banerjee.
\newblock Gradient boosted normalizing flows.
\newblock {\em Advances in Neural Information Processing Systems}, 33, 2020.

\bibitem{goodfellow2014generative}
Ian~J Goodfellow, Jean Pouget-Abadie, Mehdi Mirza, Bing Xu, David Warde-Farley,
  Sherjil Ozair, Aaron Courville, and Yoshua Bengio.
\newblock Generative adversarial networks.
\newblock {\em arXiv preprint arXiv:1406.2661}, 2014.

\bibitem{grathwohl2018ffjord}
Will Grathwohl, Ricky~TQ Chen, Jesse Bettencourt, Ilya Sutskever, and David
  Duvenaud.
\newblock Ffjord: Free-form continuous dynamics for scalable reversible
  generative models.
\newblock {\em International Conference on Learning Representations}, 2019.

\bibitem{groueix2018papier}
Thibault Groueix, Matthew Fisher, Vladimir~G Kim, Bryan~C Russell, and Mathieu
  Aubry.
\newblock A papier-m{\^a}ch{\'e} approach to learning 3d surface generation.
\newblock In {\em Proceedings of the IEEE conference on computer vision and
  pattern recognition}, pages 216--224, 2018.

\bibitem{gulrajani2017improved}
Ishaan Gulrajani, Faruk Ahmed, Martin Arjovsky, Vincent Dumoulin, and Aaron
  Courville.
\newblock Improved training of wasserstein gans.
\newblock In {\em Proceedings of the 31st International Conference on Neural
  Information Processing Systems}, pages 5769--5779, 2017.

\bibitem{He_2016_CVPR}
Kaiming He, Xiangyu Zhang, Shaoqing Ren, and Jian Sun.
\newblock Deep residual learning for image recognition.
\newblock In {\em Proceedings of the IEEE Conference on Computer Vision and
  Pattern Recognition (CVPR)}, June 2016.

\bibitem{he2016deep}
Kaiming He, Xiangyu Zhang, Shaoqing Ren, and Jian Sun.
\newblock Deep residual learning for image recognition.
\newblock In {\em Proceedings of the IEEE conference on computer vision and
  pattern recognition}, pages 770--778, 2016.

\bibitem{heusel2017gans}
Martin Heusel, Hubert Ramsauer, Thomas Unterthiner, Bernhard Nessler, and Sepp
  Hochreiter.
\newblock Gans trained by a two time-scale update rule converge to a local nash
  equilibrium.
\newblock In {\em Proceedings of the 31st International Conference on Neural
  Information Processing Systems}, pages 6629--6640, 2017.

\bibitem{isola2017image}
Phillip Isola, Jun-Yan Zhu, Tinghui Zhou, and Alexei~A Efros.
\newblock Image-to-image translation with conditional adversarial networks.
\newblock In {\em Proceedings of the IEEE conference on computer vision and
  pattern recognition}, pages 1125--1134, 2017.

\bibitem{izmailov2020semi}
Pavel Izmailov, Polina Kirichenko, Marc Finzi, and Andrew~Gordon Wilson.
\newblock Semi-supervised learning with normalizing flows.
\newblock In {\em International Conference on Machine Learning}, pages
  4615--4630. PMLR, 2020.

\bibitem{kim2020softflow}
Hyeongju Kim, Hyeonseung Lee, Woo~Hyun Kang, Joun~Yeop Lee, and Nam~Soo Kim.
\newblock Softflow: Probabilistic framework for normalizing flow on manifolds.
\newblock {\em International Conference on Neural Information Processing
  Systems}, 2020.

\bibitem{kimura2020chartpointflow}
Takumi Kimura, Takashi Matsubara, and Kuniaki Uehara.
\newblock Chartpointflow for topology-aware 3d point cloud generation.
\newblock {\em arXiv preprint arXiv:2012.02346}, 2020.

\bibitem{kingma2015adam}
Diederik~P Kingma and Jimmy~Lei Ba.
\newblock Adam: A method for stochastic gradient descent.
\newblock In {\em ICLR: International Conference on Learning Representations},
  pages 1--15, 2015.

\bibitem{kingma2016improved}
Durk~P Kingma, Tim Salimans, Rafal Jozefowicz, Xi Chen, Ilya Sutskever, and Max
  Welling.
\newblock Improved variational inference with inverse autoregressive flow.
\newblock {\em Advances in Neural Information Processing Systems},
  29:4743--4751, 2016.

\bibitem{kingma2013auto}
Diederik~P Kingma and Max Welling.
\newblock Auto-encoding variational bayes.
\newblock {\em arXiv preprint arXiv:1312.6114}, 2013.

\bibitem{klokov2020discrete}
Roman Klokov, Edmond Boyer, and Jakob Verbeek.
\newblock Discrete point flow networks for efficient point cloud generation.
\newblock In {\em 16th European Conference on Computer Vision}, 2020.

\bibitem{knapitsch2017tanks}
Arno Knapitsch, Jaesik Park, Qian-Yi Zhou, and Vladlen Koltun.
\newblock Tanks and temples: Benchmarking large-scale scene reconstruction.
\newblock {\em ACM Transactions on Graphics (ToG)}, 36(4):1--13, 2017.

\bibitem{lugmayr2020srflow}
Andreas Lugmayr, Martin Danelljan, Luc Van~Gool, and Radu Timofte.
\newblock Srflow: Learning the super-resolution space with normalizing flow.
\newblock In {\em European Conference on Computer Vision}, pages 715--732.
  Springer, 2020.

\bibitem{luo2021diffusion}
Shitong Luo and Wei Hu.
\newblock Diffusion probabilistic models for 3d point cloud generation.
\newblock In {\em Proceedings of the IEEE/CVF Conference on Computer Vision and
  Pattern Recognition}, pages 2837--2845, 2021.

\bibitem{naeem2020reliable}
Muhammad~Ferjad Naeem, Seong~Joon Oh, Youngjung Uh, Yunjey Choi, and Jaejun
  Yoo.
\newblock Reliable fidelity and diversity metrics for generative models.
\newblock In {\em International Conference on Machine Learning}, pages
  7176--7185. PMLR, 2020.

\bibitem{paschalidou2021neural}
Despoina Paschalidou, Angelos Katharopoulos, Andreas Geiger, and Sanja Fidler.
\newblock Neural parts: Learning expressive 3d shape abstractions with
  invertible neural networks.
\newblock {\em Proceedings of the IEEE/CVF International Conference on Computer
  Vision}, 2021.

\bibitem{perez2018film}
Ethan Perez, Florian Strub, Harm De~Vries, Vincent Dumoulin, and Aaron
  Courville.
\newblock Film: Visual reasoning with a general conditioning layer.
\newblock In {\em Proceedings of the AAAI Conference on Artificial
  Intelligence}, volume~32, 2018.

\bibitem{pires2020variational}
Guilherme~GP Pires and M{\'a}rio~AT Figueiredo.
\newblock Variational mixture of normalizing flows.
\newblock {\em arXiv preprint arXiv:2009.00585}, 2020.

\bibitem{pumarola2020c}
Albert Pumarola, Stefan Popov, Francesc Moreno-Noguer, and Vittorio Ferrari.
\newblock C-flow: Conditional generative flow models for images and 3d point
  clouds.
\newblock In {\em Proceedings of the IEEE/CVF Conference on Computer Vision and
  Pattern Recognition}, pages 7949--7958, 2020.

\bibitem{qi2017pointnet}
Charles~R Qi, Hao Su, Kaichun Mo, and Leonidas~J Guibas.
\newblock Pointnet: Deep learning on point sets for 3d classification and
  segmentation.
\newblock In {\em Proceedings of the IEEE conference on computer vision and
  pattern recognition}, pages 652--660, 2017.

\bibitem{rezende2015variational}
Danilo Rezende and Shakir Mohamed.
\newblock Variational inference with normalizing flows.
\newblock In {\em International Conference on Machine Learning}, pages
  1530--1538. PMLR, 2015.

\bibitem{shu20193d}
Dong~Wook Shu, Sung~Woo Park, and Junseok Kwon.
\newblock 3d point cloud generative adversarial network based on tree
  structured graph convolutions.
\newblock In {\em Proceedings of the IEEE/CVF International Conference on
  Computer Vision}, pages 3859--3868, 2019.

\bibitem{spurek2020hyperflow}
Przemys{\l}aw Spurek, Maciej Zi{\k{e}}ba, Jacek Tabor, and Tomasz
  Trzci{\'n}ski.
\newblock Hyperflow: Representing 3d objects as surfaces.
\newblock {\em arXiv preprint arXiv:2006.08710}, 2020.

\bibitem{sun2020pointgrow}
Yongbin Sun, Yue Wang, Ziwei Liu, Joshua Siegel, and Sanjay Sarma.
\newblock Pointgrow: Autoregressively learned point cloud generation with
  self-attention.
\newblock In {\em Proceedings of the IEEE/CVF Winter Conference on Applications
  of Computer Vision}, pages 61--70, 2020.

\bibitem{tao2020}
An Tao.
\newblock Unsupervised point cloud reconstruction for classific feature
  learning.
\newblock {\em
  \url{https://github.com/AnTao97/UnsupervisedPointCloudReconstruction}}, 2020.

\bibitem{tatarchenko2019single}
Maxim Tatarchenko, Stephan~R Richter, Ren{\'e} Ranftl, Zhuwen Li, Vladlen
  Koltun, and Thomas Brox.
\newblock What do single-view 3d reconstruction networks learn?
\newblock In {\em Proceedings of the IEEE/CVF Conference on Computer Vision and
  Pattern Recognition}, pages 3405--3414, 2019.

\bibitem{valsesia2018learning}
Diego Valsesia, Giulia Fracastoro, and Enrico Magli.
\newblock Learning localized generative models for 3d point clouds via graph
  convolution.
\newblock In {\em International conference on learning representations}, 2018.

\bibitem{wang2019deep}
Kaiqi Wang, Ke Chen, and Kui Jia.
\newblock Deep cascade generation on point sets.
\newblock In {\em IJCAI}, volume~2, page~4, 2019.

\bibitem{wang2018pixel2mesh}
Nanyang Wang, Yinda Zhang, Zhuwen Li, Yanwei Fu, Wei Liu, and Yu-Gang Jiang.
\newblock Pixel2mesh: Generating 3d mesh models from single rgb images.
\newblock In {\em Proceedings of the European Conference on Computer Vision
  (ECCV)}, pages 52--67, 2018.

\bibitem{dgcnn}
Yue Wang, Yongbin Sun, Ziwei Liu, Sanjay~E. Sarma, Michael~M. Bronstein, and
  Justin~M. Solomon.
\newblock Dynamic graph cnn for learning on point clouds.
\newblock {\em ACM Transactions on Graphics (TOG)}, 2019.

\bibitem{wolf2021deflow}
Valentin Wolf, Andreas Lugmayr, Martin Danelljan, Luc Van~Gool, and Radu
  Timofte.
\newblock Deflow: Learning complex image degradations from unpaired data with
  conditional flows.
\newblock {\em arXiv preprint arXiv:2101.05796}, 2021.

\bibitem{yang2019pointflow}
Guandao Yang, Xun Huang, Zekun Hao, Ming-Yu Liu, Serge Belongie, and Bharath
  Hariharan.
\newblock Pointflow: 3d point cloud generation with continuous normalizing
  flows.
\newblock In {\em Proceedings of the IEEE/CVF International Conference on
  Computer Vision}, pages 4541--4550, 2019.

\bibitem{yang2018foldingnet}
Yaoqing Yang, Chen Feng, Yiru Shen, and Dong Tian.
\newblock Foldingnet: Point cloud auto-encoder via deep grid deformation.
\newblock In {\em Proceedings of the IEEE Conference on Computer Vision and
  Pattern Recognition}, pages 206--215, 2018.

\bibitem{zamorski2020adversarial}
Maciej Zamorski, Maciej Zi{\k{e}}ba, Piotr Klukowski, Rafa{\l} Nowak, Karol
  Kurach, Wojciech Stokowiec, and Tomasz Trzci{\'n}ski.
\newblock Adversarial autoencoders for compact representations of 3d point
  clouds.
\newblock {\em Computer Vision and Image Understanding}, 193:102921, 2020.

\end{thebibliography}
